\pdfoutput=1

\documentclass{article}
\PassOptionsToPackage{numbers}{natbib}

\usepackage[final,main]{neurips_2026}

\usepackage[utf8]{inputenc} 
\usepackage[T1]{fontenc} 
\usepackage[table,usenames,dvipsnames]{xcolor}
\usepackage{amsmath, amssymb, amsthm}
\usepackage{graphicx}
\usepackage{enumitem}
\usepackage{appendix}
\usepackage{booktabs}
\usepackage{microtype}  
\usepackage{nicefrac}
\usepackage[scr=boondox]{mathalfa}
\usepackage{wrapfig}
\graphicspath{{figs/}}

\usepackage{hyperref}
\hypersetup{
 colorlinks=true,
 linkcolor=blue,
 citecolor=green,
 filecolor=magenta,   
 urlcolor=cyan
}

\definecolor{colorOdom}{HTML}{A4A4A4}      %
\definecolor{colorGT}{HTML}{BCCFB1}        %
\definecolor{colorCorr}{HTML}{2875B1}      %

\definecolor{softpurple}{HTML}{EDE7F6}
\definecolor{softblue}{HTML}{E6F0FA}
\definecolor{softgreen}{HTML}{E8F5E9}
\definecolor{softgrayblue}{HTML}{EEF2F7}
\definecolor{lightgray}{HTML}{EFEFEF}
\definecolor{midgray}{HTML}{C0C0C0}
\definecolor{darkgray}{HTML}{9B9B9B}

\DeclareMathAlphabet{\mathcursive}{U}{esstixcal}{m}{n}
\DeclareMathAlphabet{\EuScript}{U}{eus}{m}{n}
\SetMathAlphabet{\EuScript}{bold}{U}{eus}{b}{n}

\usepackage{tikz}
\usetikzlibrary{arrows.meta, positioning, chains, shapes, shadows.blur}
\tikzset{
 flow_box/.style={rectangle, draw, thick, fill=black!5, minimum height=1.5cm, text centered, font=\bfseries, align=center},
 io/.style={rectangle, rounded corners, draw, thick, fill=blue!10, minimum size=0.6cm, align=center},
 output/.style={rectangle, rounded corners, draw, thick, fill=green!10, minimum size=0.6cm, align=center},
 arrow/.style={-Latex, thick, shorten >=1pt, shorten <=1pt}
}

\newcommand{\RR}{\mathbb{R}}
\newcommand{\NN}{\mathbb{N}}

\newcommand{\PhysicalState}{x} %
\newcommand{\PhysicalSpace}{\RR^n}   %
\newcommand{\X}{\EuScript{X}}  %
\newcommand{\Xa}{\Breve{\X}}  %

\newcommand{\LatentState}{z}   %
\newcommand{\LatentSpace}{\RR^m}  %
\newcommand{\Zcal}{\mathcal{Z}}   %

 \newcommand{\Observation}{o}   %
\newcommand{\ObsSpace}{\RR^p}  %
\newcommand{\Ycal}{\mathcal{O}}   %

\def\ytob{{\mathchoice%
{\setbox0=\hbox{$\EuScript{Y}$}\raise -2.5pt\box0}%
{\setbox0=\hbox{\small$\EuScript{Y}$}\raise -2.1pt\box0}%
{\setbox0=\hbox{\scriptsize$\EuScript{Y}$}\raise -2pt\box0}%
{\setbox0=\hbox{\tiny$\EuScript{Y}$}\raise -1.5pt\box0}%
}}

 \newcommand{\obsPhys}{\Observation}
 \newcommand{\obsLatent}{\mathscr{o}}
\def\bfobsLatent{{\mathchoice%
{\boldsymbol{\mathscr{o}}\mkern-16mu\boldsymbol{\mathscr{o}}}%
{\boldsymbol{\mathscr{o}}\mkern-16mu\boldsymbol{\mathscr{o}}}%
{\boldsymbol{\mathscr{o}}\mkern-13mu\boldsymbol{\mathscr{o}}}%
{\boldsymbol{\mathscr{o}}}%
}}

\newcommand{\TargetVar}{s}  %
\newcommand{\Estimate}{\hat{\PhysicalState}} %

\newcommand{\GenerativeDrift}{f}  %
\newcommand{\GenerativeObs}{h} %
\newcommand{\LatentDynamics}{\varphi}%
\newcommand{\BaseDynamics}{g}  %
\newcommand{\Decoder}{\gamma}  %
\newcommand{\Encoder}{\boldsymbol{T}}%

\newcommand{\TimeStep}{\delta} %
\newcommand{\Gain}{\kappa{}}  %
\newcommand{\Params}{\Omega}   %

\newcommand{\flowf}{{\Phi_f}}

\def\edot#1{\renewcommand{\arraystretch}{0.2}%
\begin{array}[b]{@{}c@{}}\textrm{\scriptsize $\circ$}\\\renewcommand{\arraystretch}{1}#1\end{array}%
\renewcommand{\arraystretch}{1}}

\def\flowf{{\Phi_{\GenerativeDrift}}}
\def\Lie{\mathcursive{L}_{\GenerativeDrift}}

\def\dum{\eta}
\def\bfdum{\boldsymbol{\dum}}

\def\dumr{\xi}
\def\bfdumr{\boldsymbol{\dumr}}
\def\bfcaldum{\boldsymbol{\mathfrak{Y}}}

\def\tob{\tau}                           %
\def\ytob{\obsLatent}                    %

\newcommand{\myparagraph}[1]{\noindent \textbf{#1}}

\newif\ifshowmodif
\showmodiftrue
\newcommand{\startmodifVA}{\ifshowmodif\begingroup\color{black}\fi}
\usepackage{xspace}
\newcommand{\stopmodif}{\ifshowmodif\endgroup\fi\xspace}

\newcommand{\PB}[1]{\textcolor{black}{#1}}
\newcommand{\lb}[1]{#1}

\newtheorem{theorem}{Theorem}[section]
\newtheorem{proposition}[theorem]{Proposition}
\newtheorem{lemma}[theorem]{Lemma}
\newtheorem{corollary}[theorem]{Corollary}
\newtheorem{definition}[theorem]{Definition}
\newtheorem{remark}[theorem]{Remark}
\newtheorem{assumption}[theorem]{Assumption}

\title{Provable State Estimation with Recurrent Models}

\author{%
  Vincent Andrieu\textsuperscript{1}, Pauline Bernard\textsuperscript{2}, Lucas Brivadis\textsuperscript{3}, Laurent Praly\textsuperscript{2} and Christian Wolf\textsuperscript{4}\\[0.5em]
  \small \textsuperscript{1}LAGEPP, CNRS, UCBL 1 \quad \textsuperscript{2}CAS, Mines Paris - PSL \\
  \small \textsuperscript{3}L2S, CNRS \quad \textsuperscript{4}Naver Labs Europe
}
\date{}

\begin{document}

\setlength{\abovedisplayskip}{3pt}
\setlength{\belowdisplayskip}{3pt}

\maketitle

\begin{abstract}
Estimating the state of a dynamical system from observations is a crucial task in robotics and control. Despite their empirical success, recurrent models often lack formal guarantees on convergence and robustness. We bridge this gap by analyzing these architectures from the perspective of nonlinear observer theory. By framing RNNs as a time-discretization of continuous latent dynamical systems (Neural ODEs), we demonstrate that enforcing a contraction condition on the latent flow guarantees the existence of a smooth, injective \PB{map} of the true physical state into the latent space. We prove that this architecture acts as a \textit{tunable} observer: by increasing a single scalar gain that scales the latent vector field, the estimation error can be driven to an arbitrarily small neighborhood of zero at an exponentially fast rate. We further establish quantitative bounds on the MLP complexity required to decode this latent state, and extend our formal guarantees to the discrete-time residual RNN implementation. We validate the theory on a real mobile robotic task, where a provably contracting model outperforms a vanilla GRU with substantially more stable training. We also report that unconstrained GRUs converge to contracting dynamics on this task, suggesting that contraction is an emergent property of well-trained sequence models.
\end{abstract}


\section{Introduction}
\label{sec:intro}

We consider the classical problem of estimating 
the state of a dynamical system, 
given some online measurements of the observations $\obsPhys(t)$. This is motivated by applications in robotics control, system monitoring, fault diagnosis, and data assimilation and has been addressed independently in different fields, related to \textit{Observer Design} \cite{BerAndAst} in control theory and to \textit{State Representation Learning} \cite{StateRepFiliat2018} in ML and embodied AI. In many methods, the quality of the latent state representation is determined by the downstream application building on it, and in this case, it is often learned together (``\textit{end-to-end}'') with the downstream task with RL \cite{janny2025} or imitation learning \cite{DBLP:conf/nips/DingFAP19}; alternatively through pre-training, i.e. through next token prediction \cite{humanoidlocomotiontokenprediction2024}, with contrastive learning \cite{timeconstrastive2018,curl2020} or recently with joint embedding predictive architectures \cite{jepa2}.

In this work, we suppose the existence of a target variable $\TargetVar(t)$ of interest linked to the state of the system (e.g. the pose of a robot or a hidden quantity of a biological reactor) and are interested in its estimation. A data-driven model for this task is required to learn, both, a latent state representation and a mapping from the latent state to the variable of interest.
Natural questions then arise: \textit{Under which conditions does this neural architecture guarantee that the estimate $\hat{\TargetVar}(t)$ asymptotically converges to the true target $\TargetVar(t)$?  Is it possible to establish formal guarantees regarding the performance and tunability of a trained model?}

To answer these questions, we suppose that the relation between the observations and the target variable is generated by a deterministic dynamical system. More precisely, we assume the existence of a physical\footnote{We call this state \textit{physical} as it is usually obtained through the application of physics laws (e.g. motion laws giving $x$ as positions and velocities, electromagnetic laws giving $x$ as fluxes and currents, chemical laws giving $x$ as concentrations, etc.), as opposed to the \textit{latent} state (later introduced in this paper) that is purely abstract. Note, though, that the knowledge of $f,h,\sigma$ is not necessary in this paper, so those maps could also have an abstract nature. What is important is that $s$ and $o$ are linked via a state that has a kind of \textit{markovian} property, namely, its instantaneous evolution only depends on its current value.} state $\PhysicalState(t) \in  \PhysicalSpace$, governed by an unknown ODE, that determines both $\obsPhys(t)$ and $\TargetVar(t)$:
\begin{equation}
\frac{d\PhysicalState}{dt}(t) = \GenerativeDrift(\PhysicalState(t)), \quad \obsPhys(t) = \GenerativeObs(\PhysicalState(t)), \quad \TargetVar(t) = \varsigma(\PhysicalState(t)),
\label{eq:generative_model}
\end{equation}
where $\GenerativeDrift$, $\GenerativeObs$, and $\varsigma$ are continuous functions. Moreover, we assume this state is somehow minimal, in the sense that it can be dynamically reconstructed from the knowledge of $o$ -- it is \textit{observable} in the control theory sense \cite{BerAndAst}. Our method thus relies on reconstructing the full physical state itself, looking for an estimate $\Estimate(t) \approx \PhysicalState(t)$, from which an estimate $\hat{s}(t)$ of $s(t)$ can then be recovered. Without loss of generality, we thus set $\TargetVar(t) = \PhysicalState(t)$ in this paper\footnote{\startmodifVA Reconstructing only $\varsigma(x)$ when $x$ itself is not observable is the object of \textit{functional observability} \cite{darouach2000functional,fernando2010functional,montanari2022functional}, also addressed with KKL observers \cite{spirito2022functional}; we leave it for future work (see Section \ref{sec:conclusion}).\stopmodif}.
When the dependence on the initial condition is required, we denote the unique solution at time $t$ starting from $\PhysicalState_0$ as $\PhysicalState(t) = \flowf(\PhysicalState_0, t)$.
This point of view formulates the machine learning problem as an observer design problem.%

\myparagraph{From RNNs to Neural ODEs} ---
\textit{Recurrent Neural Networks} (RNNs) and their variants (LSTM \cite{DBLP:journals/neco/HochreiterS97}, GRU \cite{cho-etal-2014-learning}) have been shown to be powerful architectures for solving machine learning problems on sequential data. They process sequential observations $\obsPhys_k \in \ObsSpace$ according to the following recurrence relation:
\begin{equation}\label{eq:rnn_standard}
 \LatentState_{k+1} = \LatentDynamics_\Params(\LatentState_k, \Observation_{k+1}),
\end{equation}
where $\LatentState_k \in \LatentSpace$ denotes the hidden (or latent) state of the network, $\LatentDynamics_\Params$ denotes the non-linear update map, and $\Params$ is a vector of trainable weights. The output of the network, which gives an estimate of a target variable $\TargetVar_k$, is usually computed by a readout map, often a Multi-Layer Perceptron (MLP):
\begin{equation}\label{eq:rnn_output}
 \hat{\TargetVar}_k = \Decoder_\Params(\LatentState_k).
\end{equation}
\begin{figure} \centering 
\definecolor{physblue}{RGB}{45,100,180}
\definecolor{rnnorng}{RGB}{210,110,20}
\definecolor{mlpgrn} {RGB}{40,150,80}
\definecolor{traj1}  {RGB}{190,50,50}
\definecolor{traj2}  {RGB}{50,90,200}
\definecolor{traj3}  {RGB}{130,50,180}
\definecolor{traj4}  {RGB}{20,150,130}
\definecolor{reftrj} {RGB}{0,0,0}

\noindent\begin{tikzpicture}[
  every node/.style={font=\small},
  abox/.style={rectangle, draw=#1!70, line width=1pt, rounded corners=5pt,
               fill=#1!9, minimum width=5.7cm, minimum height=1.1cm,
               align=center, text width=5.5cm},
  arr/.style ={-{Latex[length=4.5pt,width=4pt]}, line width=1pt},
  darr/.style={-{Latex[length=4pt,width=3.5pt]}, line width=0.9pt, dashed},
  rbox/.style={rounded corners=5pt, draw=#1!45, line width=1pt, fill=#1!5},
]

\node[abox=physblue] (sys) at (1.55,4.90)
  {Physical System\\[3pt]$\dot{x}=f(x),\;o=h(x),\;s=\sigma(x)$};
\node[abox=rnnorng]  (rnn) at (1.55,3.30)
  {Contracting RNN\\[3pt]$\dot{z}=\kappa\,g_\Omega(z,o)$};
\node[abox=mlpgrn]   (mlp) at (1.55,1.65)
  {MLP Decoder\\[3pt]$\hat{s}=\gamma_\Omega(z)$};

\draw[arr, physblue!70!black]
  (sys.south) -- node[right=3pt]{$o(t)$} (rnn.north);
\draw[arr, rnnorng!80!black]
  (rnn.south) -- node[right=3pt]{$z(t)$} (mlp.north);
\draw[arr, mlpgrn!70!black]
  (mlp.south) -- ++(0,-0.15)
  node[below=-3pt, align=center]{$\hat{s}(t)\;\xrightarrow{\;t\to\infty\;}\;s(t)$};

\draw[darr, physblue!80]
  ([xshift=-4pt]sys.south west)
  .. controls (-1.80,4.25) and (-1.80,3.95) ..
  ([xshift=-4pt]rnn.north west)
  node[pos=0.50, right=6pt, align=center, font=\footnotesize]
    {$\mathbf{T}_\kappa$ \scriptsize embedding};

\begin{scope}[cm={0.420,0,0,0.4889,(3.359,1.0755)}]

\fill[rounded corners=5pt, draw=rnnorng!70, line width=1pt, fill=rnnorng!9]
  (3.55,3.05) rectangle (19.7,6.05);
\node[rnnorng!80, font=\footnotesize\bfseries] at (3.98,5.82) {$\mathcal{Z}$};

\node[black, font=\small\bfseries] at (11.625,6.95) {Contracting property};

\draw[reftrj, line width=0.4pt, dashed, line cap=round]
  (4.8,4.57) -- (18.8,4.57);
\node[reftrj, font=\scriptsize, below=3pt] at (10.5,4.57)
  {$\mathbf{T}_\kappa(x(t))-z(t)$};

\draw[traj1, line width=1.4pt]
  (4.8,5.80)
  .. controls (6.3,4.95) and (8.5,4.63) .. (11.5,4.58)
  .. controls (14.5,4.57) and (16.5,4.57) .. (18.8,4.57);
\draw[traj3, line width=1.1pt, dashed]
  (4.8,5.22)
  .. controls (6.3,4.78) and (8.5,4.60) .. (11.5,4.57)
  .. controls (14.5,4.57) and (16.5,4.57) .. (18.8,4.57);

\draw[traj4, line width=1.1pt, dashed]
  (4.8,3.80)
  .. controls (6.3,4.32) and (8.5,4.52) .. (11.5,4.56)
  .. controls (14.5,4.57) and (16.5,4.57) .. (18.8,4.57);

\draw[traj2, line width=1.4pt]
  (4.8,3.22)
  .. controls (6.3,4.16) and (8.5,4.49) .. (11.5,4.55)
  .. controls (14.5,4.57) and (16.5,4.57) .. (18.8,4.57);

\node[black!70, font=\small, align=center,
      fill=rnnorng!8, draw=rnnorng!50, rounded corners=3pt, inner sep=3pt]
  at (13.2,5.90)
  {$|z_a(t)-z_b(t)|\;\leq\;e^{-\kappa t}\,|z_a(0)-z_b(0)|$};

\fill[rbox=mlpgrn] (3.55,-0.32) rectangle (19.7,2.68);
\node[mlpgrn!65, font=\footnotesize\bfseries] at (3.98,2.52) {$\mathcal{S}$};

\draw[reftrj, line width=0.4pt, dashed, line cap=round]
  (4.8,1.18) -- (18.8,1.18);
\node[reftrj, font=\scriptsize, below=3pt] at (10.5,1.18) {};%

\draw[traj1, line width=1.4pt]
  (4.8,2.40)
  .. controls (7.5,1.68) and (11.0,1.36) .. (14.0,1.32)
  .. controls (16.0,1.32) and (17.5,1.32) .. (18.8,1.32);
\draw[traj3, line width=1.1pt, dashed]
  (4.8,2.05)
  .. controls (5.5,2.00) and (7.0,0.65) .. (8.0,0.72)
  .. controls (8.5,0.75) and (10.5,1.42) .. (11.5,1.35)
  .. controls (13.0,1.10) and (16.0,1.25) .. (18.8,1.24);

\draw[traj4, line width=1.1pt, dashed]
  (4.8,0.35)
  .. controls (5.2,0.28) and (6.5,1.75) .. (7.5,1.70)
  .. controls (8.0,1.65) and (9.5,0.85) .. (10.5,0.95)
  .. controls (12.0,1.05) and (15.0,1.18) .. (18.8,1.11);

\draw[traj2, line width=1.4pt]
  (4.8,-0.05)
  .. controls (7.5,0.65) and (11.0,1.00) .. (14.0,1.04)
  .. controls (16.0,1.04) and (17.5,1.04) .. (18.8,1.04);

\node[black!65, font=\footnotesize,
      fill=mlpgrn!15, draw=mlpgrn!60, rounded corners=2pt, inner sep=2.5pt]
  at (13.2,-0.18)
  {$|\hat{s}(t)-s(t)|<\epsilon\;$ for $\;t\ge t_o$};

\end{scope}

\end{tikzpicture}
\vspace*{-2mm}
\caption{\label{fig:teaser}\textbf{Provable state estimation:} a contracting Neural ODE with latent dynamics  $\dot{z} = \kappa g_\Omega(z, o)$ processes observations $o(t)$ from the physical system with unobserved $x(t)$, of which elements $s(t)$ are of interest. A decoder $\gamma_\Omega$ maps the latent state to an estimate $\hat{s}(t)$. The existence of an injective state-to-latent map $\mathbf{T}_\kappa: \mathcal{X} \to \mathcal{Z}$ is guaranteed (Theorems~\ref{thm:embedding}-\ref{thm:injectivity}). 
 A contraction property ensures that  two latent trajectories $z_a, z_b$ initialized anywhere in $\mathcal{Z}$
 exponentially forgets initial conditions.  The estimate $\hat{s}(t)$ reaches an $\epsilon$-neighborhood of the true $s(t)$ after a finite observation time, and increasing $\kappa$ accelerates convergence at the cost of larger network capacity (Theorem~\ref{thm:neural_complexity}).}
  \vspace{-1em}
\end{figure}
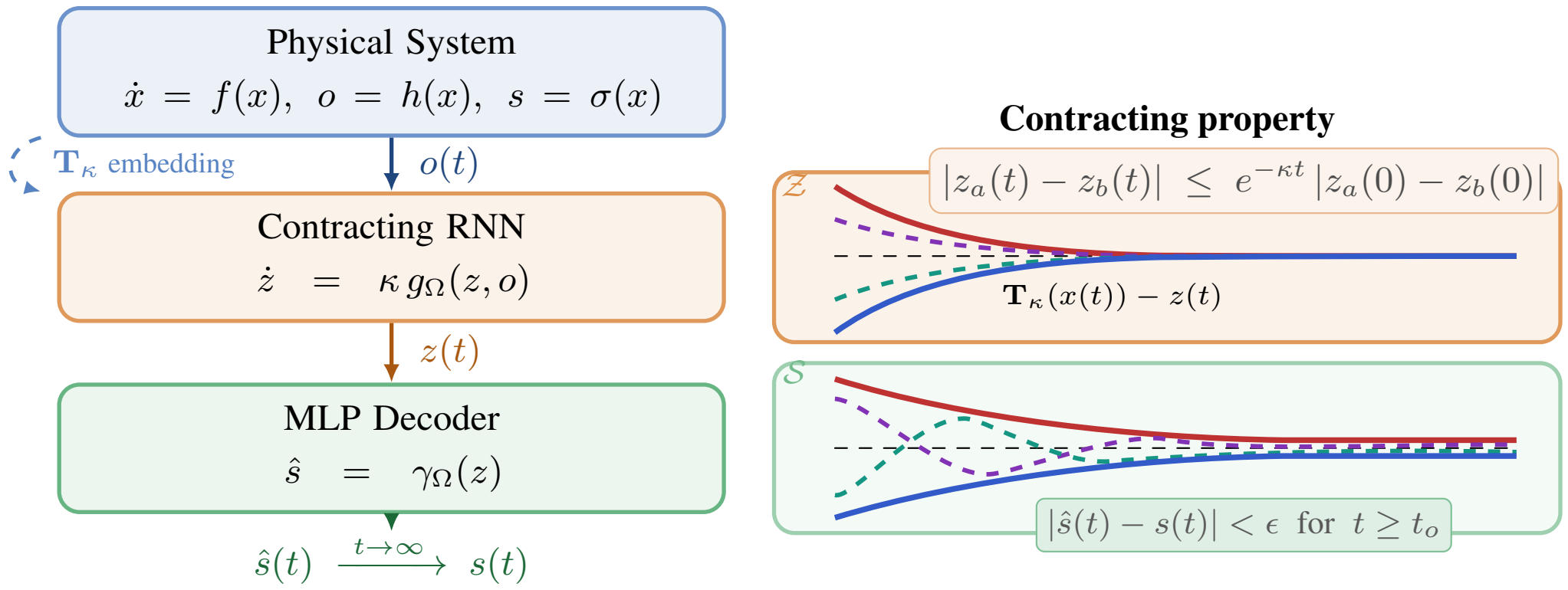
Reformulated with residual connection, relation \eqref{eq:rnn_standard} can be rewritten as:
\begin{equation}
 \LatentState_{k+1} = \LatentState_k + \TimeStep \cdot \BaseDynamics_\Params(\LatentState_k, \Observation_{k+1}),
 \label{eq:rnn_residual}
\end{equation}
where $\TimeStep > 0$ is a step-size parameter. This formulation highlights the incremental nature of the state update. When the sampling period of the observations tends to zero ($\TimeStep \to 0$), this recurrence converges to a continuous-time differential equation. This point of view naturally leads to \textit{Neural ODEs}\footnote{In the original work \cite{chen2018neural} the ODE was not necessarily applied to problems involving dynamical systems but rather built an internal dynamical system running over the layers of a ResNet-like neural network.} \cite{chen2018neural}, where the latent state and the output evolve continuously in time:
\begin{equation}
 \frac{d\LatentState}{dt}(t) = \BaseDynamics_\Params(\LatentState(t), \obsPhys(t)),\quad \hat{\TargetVar}(t) = \Decoder_\Params(\LatentState(t)).
 \label{eq:neural_ode}
\end{equation}
Analyzing the continuous-time limit allows one to use classical tools from dynamical systems and control theory to study stability and convergence properties. As we will show, these properties remain true for practical discrete implementations if the sampling step is small enough.

\myparagraph{Contributions} ---
While classical observer theory usually needs explicit knowledge of the functions $\GenerativeDrift$ and $\GenerativeObs$, this paper develops a theoretical framework to certify the representational capabilities of data-driven RNN-MLP estimators. 
We prove that, under mild assumptions, there exists a choice of the trainable weights $\Params$ that solves the estimation task with guaranteed exponential convergence.

To obtain these guarantees, we explicitly isolate a tunable scalar gain parameter $\Gain > 0$ in the latent dynamics  that acts as a global time-scale separation gain. More precisely, we write the estimator as 
\begin{equation}
 \frac{d\LatentState}{dt}(t) = \Gain \BaseDynamics_\Params(\LatentState(t), \obsPhys(t)),\quad \hat{\TargetVar}(t) = \Decoder_\Params(\LatentState(t))
 \label{eq:high_gain_ode}
\end{equation}
where $\BaseDynamics_\Params: \LatentSpace \times \ObsSpace \to \LatentSpace$ represents the normalized base latent dynamics parameterized by $\Params$.
The following contributions are obtained in this paper.
\begin{enumerate}[label=\textbf{\arabic*.},nosep,itemsep=1mm,labelindent=0mm,leftmargin=4mm,topsep=0mm]
 \item \textbf{Existence of the State-to-Latent Map:} We establish a structural result linking deep learning and the Kazantzis-Kravaris-Luenberger (KKL) theory \cite{AndrieuPraly2006}. We prove that if (i) the latent base dynamics generates a contraction, (ii) the latent gain is large enough, (iii) the physical state is differentially observable from $\Observation$, (iv) the latent dimension is large enough\footnote{An explicit bound is given depending on the dimension of the physical state $x$ and the observability property from $o$.}, then there exists a
smooth injective map $x\mapsto \Encoder_\Gain (x)$ which gives a representation of the physical
state in the latent space. \startmodifVA Contrary to previous existence results with nonlinear latent dynamics \cite{pachy2024existence}, the latent vector field may be coupled across latent coordinates, as in GRUs and LSTMs, rather than diagonal; the price is a rank condition on the latent dynamics, which is generic (Appendix \ref{Sec_AssumpFilterCapacity}).\stopmodif
 \item \textbf{Convergence and Tunability:} In an ideal setting where $\Decoder_\Params$ is a left-inverse of $\Encoder_\Gain$, we prove that 
 the estimation error provided by the estimator \eqref{eq:high_gain_ode} converges exponentially fast to zero whatever the initial condition of the latent dynamics. Moreover, the architecture is formally \textit{tunable}, which means the error can be made arbitrarily small in finite time.
 \item \textbf{MLP Complexity Bounds:} When the  decoder $\Encoder^{-1}_\Gain$ is approximated by a neural network $\Decoder_\Params$, we show that the tunability property is preserved and give a quantitative link between the latent gain $\Gain$ and the required MLP complexity. We show that while a high gain gives a faster convergence, it may increase the Lipschitz constant of the mapping, which requires a larger network.
 \item \textbf{Guarantees for Discrete Implementation:} We explicitly extend our continuous-time analysis to the practical case of discrete-time RNNs. We prove that the tunability property is preserved and we give a formal condition linking the required gain $\Gain$ and the maximum sampling step $\TimeStep$. 
 \item \textbf{Experimental Results}: We showcase the effect of the contracting property on a state estimation problem in robotics with real data and provide evidence for the tunability result in simulation.
\end{enumerate}

\section{Related Works}
\label{sec:related work}

\startmodifVA \myparagraph{Model-based observers.}\stopmodif
\startmodifVA When a model $(\GenerativeDrift,\GenerativeObs)$ is available, state\stopmodif estimation relies on analytical methods like Kalman, LMI-based or high-gain observers \cite{BerAndAst}. Among them, the Kazantzis-Kravaris-Luenberger (KKL) observer \cite{KazantzisKravaris1998, AndrieuPraly2006} immerses the physical state dynamics into a stable, observation-driven \textit{linear} filter, through a map solving a partial differential equation (PDE) that is usually analytically intractable. \startmodifVA Observer design is also closely related to \textit{synchronization}: an observer is a system that synchronizes with the plant it is driven by \cite{nijmeijer1997observer}, and the existence of an attracting map from the plant state to the observer state, as in Theorem \ref{thm:embedding} below, is known as \textit{generalized synchronization}. \textit{Contraction} is an important approach to obtain an observer (see for instance \cite{BerAndAst,pachy2024existence}) or a synchronizing system (see for instance \cite{andrieu2018synchronization}).\stopmodif

\startmodifVA \myparagraph{Data-driven observers and learned dynamical models.}\stopmodif
Learning dynamical systems from offline datasets is a task inherited from \textit{System Identification} \cite{lennart1999system}; learned models are then coupled with classical algorithms for analysis \citep{van2022kernel}, simulation \citep{pfaff2020learning,bauersfeld2021neurobem,janny2024space}, state estimation \citep{stubberud1991neural,de2007nonlinear}, prediction \citep{janny2021deep} or control \citep{bauersfeld2021neurobem,van2020data,zoboli2023deep}. \startmodifVA In the KKL line of work, machine learning approximates the immersion map, or rather its inverse, from data (supervised approaches on simulated datasets, autoencoder losses embedding the PDE, or a mix of both \cite{ramos2020numerical, peralez2021deep,janny2021deep,niazi2023learning,buisson2023towards,miao2023learning,tang2024synthesis}), but always with \textit{linear} latent dynamics.\stopmodif \startmodifVA \textit{Reservoir computing} \cite{lukovsevivcius2009reservoir} and reservoir observers \cite{lu2017reservoir} go one step further by driving a fixed random nonlinear recurrent network with the observations and training a readout: the \textit{echo state property} \cite{jaeger2001echo} is essentially a contraction assumption, but neither the injectivity of the induced map nor the tunability of the estimate come with guarantees.\stopmodif The works \cite{GIULI2026113005,MarFilAlNLal} suggest that contraction-like properties can be enforced during training and enable the design of control laws and observers.
\startmodifVA The only existence result for KKL observers with nonlinear contracting latent dynamics \cite{pachy2024existence} assumes a \textit{diagonal} latent vector field, each coordinate being an independent scalar contraction driven by the observation. This structure is the pivot of the proof (coordinate-wise regularity, Vandermonde rank argument), and no recurrent architecture in practical use has this form.\stopmodif

\myparagraph{Recurrent models} like \textit{LSTM} \cite{DBLP:journals/neco/HochreiterS97} and \textit{GRU}  \cite{cho-etal-2014-learning} \startmodifVA precisely mix nonlinearities across latent coordinates, so that their latent Jacobian is dense.\stopmodif
Their memory capacity scaling problem has been addressed by external neural memory \cite{DBLP:journals/corr/GravesWD14,endtoendmemorynetworks2015,neuralgpu2016} before transformers even arrived. Recurrence gained traction again, e.g. through state space models like \textit{S4} \cite{s4iclr2022}, \textit{Mamba} \cite{gu2024mamba} and \textit{LRU} \cite{lru2023}. Attention has been combined with recurrence, e.g. \textit{xLSTM} \cite{beck2024xlstm}, and full-fledged Transformers handling a memory bank are proposed by \textit{MooG} \cite{moog2024},  \textit{TTM} \cite{ryoo2023tokenturingmachines} and \textit{Kinaema} \cite{kinaema2025}. 
\startmodifVA We lift the diagonality restriction of \cite{pachy2024existence} and provide existence, injectivity and tunability guarantees for this class under a contraction condition and a generic rank condition, and derive an architecture satisfying the former by design.\stopmodif
\section{The Latent Embedding Architecture}
\label{sec:architecture}

To ensure the mathematical rigor, we introduce the following regularity and invariance assumptions.
\begin{assumption}[Blanket Assumption]
\label{ass:regularity}
Throughout this paper, we assume
the functions $\GenerativeDrift$ and $\GenerativeObs$ in \eqref{eq:generative_model} are defined on the entire system state space $\RR^n$, are continuously differentiable as many times as needed
and the physical state trajectories $\PhysicalState(t)$ obtained from \eqref{eq:generative_model} remain in positive times
in a compact set $\X \subset \PhysicalSpace$.
We assume also
the function $\BaseDynamics_\Params$ 
defining the latent dynamics \eqref{eq:high_gain_ode} is as many times continuously differentiable as needed.
\end{assumption}
Based on this assumption, we introduce a compact and convex inflation of the set $\X$, denoted $\Xa$, such that $\X$ is contained within the topological interior of $\Xa$. We then define the corresponding extended compact observation set $\Ycal = \GenerativeObs(\Xa) \subset \ObsSpace$.
This allows us to establish a safe operating region for the latent state. We define the compact set $\Zcal \subset \LatentSpace$ as a closed ball $\Zcal = \{\LatentState \in \LatentSpace : |\LatentState| \le R_z\}$, where the radius $R_z > 0$ is chosen large enough (specifically $R_z \ge 2\max_{\Observation \in \Ycal}|\BaseDynamics_\Params(0, \Observation)|$).

\subsection{Fading Memory via Contracting Latent Flows}
\label{sec:contraction}

To ensure the RNN acts as a fading memory filter that exponentially forgets its arbitrary initial condition $\LatentState_0$, we restrict its latent vector field to generate contracting flows.

\begin{assumption}[Monotonicity of the Latent Dynamics]
\label{ass:contraction} 
The base latent vector field $\BaseDynamics_\Params$ is uniformly monotonic on $\Zcal$ for any observation in $\Ycal$. More precisely, its Jacobian matrix satisfies:
\begin{equation}
 \frac{\partial \BaseDynamics_\Params}{\partial \LatentState}(\LatentState, \Observation) + \left(\frac{\partial \BaseDynamics_\Params}{\partial \LatentState}(\LatentState, \Observation)\right)^\top \le -2 I_m, \quad \forall (\LatentState, \Observation) \in \Zcal \times \Ycal.
 \label{eq:contraction_condition}
\end{equation}
\end{assumption}

For instance, a residual RNN cell with a linear dissipative term,
$\dot{\LatentState}_i = \Gain ( -\alpha_i \LatentState_i + \lambda_i \tanh(\LatentState_i + U_i \Observation) )$, globally satisfies \eqref{eq:contraction_condition} if
$\alpha_i - |\lambda_i| \ge 1$.
It also covers the case where there exist $P\in \RR^{m\times m}$ positive definite and $a>0$ such that 
$P\frac{\partial \BaseDynamics_\Params}{\partial \LatentState}(\LatentState, \Observation) + \left(\frac{\partial \BaseDynamics_\Params}{\partial \LatentState}(\LatentState, \Observation)\right)^\top P\le -a P$ 
for all $(z,o)$, modulo a linear change of coordinates of $z$ and a scaling on $\kappa$. This is exploited in Appendix \ref{app:rookie_arch} to propose a novel contracting GRU and handle multi-layered structures. \startmodifVA No constraint is put on the \textit{structure} of $\BaseDynamics_\Params$: a stable linear recurrence $\BaseDynamics_\Params(z,o) = A z + B o$ with $A + A^\top \le -2I_m$, as in state space models (S4, LRU), is a special case; the nonlinear coupled recurrences of GRUs and LSTMs are covered as well.\stopmodif

This monotonicity directly enforces the fading memory property: it guaranties that the operating region is forward-invariant and that all latent trajectories exponentially converge towards each other, effectively erasing the initial condition.

\begin{proposition}[Invariant and Contracting Flow]
\label{prop:invariance}
Under Assumptions \ref{ass:regularity} and \ref{ass:contraction}, the compact set $\Zcal$ is forward-invariant for the latent dynamics in \eqref{eq:high_gain_ode}, for any input $\obsPhys(t) \in \Ycal$. Moreover, for any $\Gain>0$ and for any two latent trajectories $\LatentState_a(t), \LatentState_b(t)$ in $\Zcal$, driven by the same observation $\obsPhys(t) \in \Ycal$, we have a contraction:
$$
|\LatentState_a(t) - \LatentState_b(t)| \le e^{-\Gain (t - t_0)} |\LatentState_a(t_0) - \LatentState_b(t_0)|, \quad \forall t \ge t_0.
$$
\end{proposition}
The proof is in Section \ref{app:proof_invariance}.
The scalar parameter $\Gain$ acts as a tunable gain that globally scales the vector field. By increasing $\Gain$, we accelerate the internal dynamics of the observer relative to the evolution of the input $\obsPhys(t)$, ensuring a convergence rate proportional to $e^{-\Gain t}$.

\subsection{Existence of the State-to-Latent Map}
\label{sec:embedding}

With forgetting its initialization, the latent state becomes entirely driven by the physical system. Specifically, $\LatentState(t)$ asymptotically converges to an image of the physical state, denoted $\Encoder_\Gain(\PhysicalState(t))$. See Section \ref{app:proof_embedding} for a proof of the result below.

\begin{theorem}[Existence of the State-to-Latent Map]
\label{thm:embedding}
Suppose Assumptions \ref{ass:regularity} and \ref{ass:contraction} are satisfied. For any gain $\Gain$ large enough ($\Gain > 2L_\GenerativeDrift$, with $L_\GenerativeDrift$ the Lipschitz constant of $\GenerativeDrift$ over $\Xa$), there exists a mapping $\Encoder_\Gain: \X \to \Zcal$ such that:
\begin{enumerate}
 \item \textbf{Invariance:} The physical trajectories are mapped to the latent trajectories:
 \begin{equation}
  \Encoder_\Gain(\PhysicalState(t)) = \LatentState(t) , \quad \forall t \ge 0\quad \text{if } \LatentState(0) = \Encoder_\Gain(\PhysicalState_0).
 \end{equation}
 \item \textbf{Attractivity:} For any initialization $\LatentState_0 \in \Zcal$, the latent state asymptotically tracks the image by $\Encoder_\Gain$ of the physical state:
 \begin{equation}
  |\LatentState(t) - \Encoder_\Gain(\PhysicalState(t))| \le e^{-\Gain t} |\LatentState_0 - \Encoder_\Gain(\PhysicalState_0)|, \quad \forall t \ge 0.
 \end{equation}
 \item \textbf{Regularity:} $\Encoder_\Gain$ is $C^1$ on $\X$ with Lipschitz partial derivatives.
\end{enumerate}
\end{theorem}

\subsection{Information Preservation: Injectivity of the State-to-Latent Map}
\label{sec:injectivity}

Theorem \ref{thm:embedding} guarantees that the RNN creates a latent representation of $x$ as $\LatentState = \Encoder_\Gain(\PhysicalState)$. However, for the MLP decoder to be able to reconstruct the physical state via $\Estimate = \Decoder_\Params(\LatentState)$, this representation must preserve the original information. Mathematically, it means that $\Encoder_\Gain$ must be \textit{injective}. 
The injectivity needs two properties: the physical system must give enough information in its observations (Observability), and the RNN must have a sufficient dimension and algebraic structure to process it (Filter Rank Condition).

\begin{assumption}[Differential Observability]
\label{ass:observability}
There exists an integer $\ell \ge 0$ such that the physical state $\PhysicalState$ is uniquely determined by the observation and its $\ell$ first time-derivatives. More precisely, there exists a constant $c_{obs} > 0$ such that the map $\PhysicalState \mapsto (\GenerativeObs(\PhysicalState), \Lie \GenerativeObs(\PhysicalState), \dots, \Lie^\ell \GenerativeObs(\PhysicalState))$ is Lipschitz injective on $\X$:
\\
\vskip -4.5em
\null
\begin{equation}\label{eq:observability}
 \sum_{i=0}^\ell |\Lie^i \GenerativeObs(\PhysicalState_a) - \Lie^i \GenerativeObs(\PhysicalState_b)| \ge c_{obs} |\PhysicalState_a - \PhysicalState_b|, \quad \forall \PhysicalState_a, \PhysicalState_b \in \X,
\end{equation}
where $\Lie^i \GenerativeObs$ is the $i$-th Lie derivative of $\GenerativeObs$ along the vector field $\GenerativeDrift$.
\end{assumption}

\begin{assumption}[Filter Rank and Capacity Condition]
\label{ass:filter_capacity}
The base dynamics $\BaseDynamics_\Params$ must satisfy the following conditions:
\begin{enumerate}
 \item \textbf{Capacity:} The dimension of the latent space satisfies $m \ge p(\ell + 1)$, where $p = \dim(\Observation)$.
 \item \textbf{Input-to-Dynamics Injectivity :} For any $\LatentState \in \Zcal$, the map $\Observation \mapsto \BaseDynamics_\Params(\LatentState, \Observation)$ is injective on $\Ycal$.
 \item \textbf{Rank Condition:} By Assumption \ref{ass:contraction}, for any $\Observation$ in $\Ycal$, there exists a unique equilibrium $\varphi_0(\Observation)$ such that $\BaseDynamics_\Params(\varphi_0(\Observation), \Observation) = 0$. Let the Jacobians be evaluated at this equilibrium:
 $$A(\Observation) = \frac{\partial \BaseDynamics_\Params}{\partial \LatentState}(\varphi_0(\Observation), \Observation), \quad B(\Observation) = \frac{\partial \BaseDynamics_\Params}{\partial \Observation}(\varphi_0(\Observation), \Observation).$$
 The $(m, p(\ell+1))$  matrix $C_\ell(\Observation)$ defined as:
 \begin{equation}
  C_\ell(\Observation) = \begin{bmatrix} B(\Observation) & A(\Observation)B(\Observation)& \cdots & A(\Observation)^{\ell}B(\Observation)\end{bmatrix}
 \end{equation}
 is left-invertible for all $\Observation \in \Ycal$.
\end{enumerate}
\end{assumption}
    The rank condition is related to the left-invertibility \cite{9835013} of the linear control system $\frac{d\xi}{dt} = A(\Observation)\xi +B(\Observation)u$ with $\Observation$ fixed.
    This is considered to be a mild assumption in nonlinear control, as
    almost any linear control system satisfies it \cite{brivadis2023further} (see also Appendix \ref{Sec_AssumpFilterCapacity} for a detailed analysis).
Under these assumptions, we can show that increasing the gain $\Gain$ forces the latent state to contain the necessary measurement derivatives, which makes the mapping invertible.

\begin{theorem}[Lipschitz Injectivity of $\Encoder_\Gain$]
\label{thm:injectivity}
Under Assumptions \ref{ass:regularity}, \ref{ass:contraction}, \ref{ass:observability}, and \ref{ass:filter_capacity}, there exist a constant $c_T > 0$ and a minimal gain $\underline{\Gain}$ such that for all $\Gain \ge \underline{\Gain}$, there exists $\Encoder_\Gain: \X \to \LatentSpace$ which satisfies the conclusion of Theorem \ref{thm:embedding} and which is Lipschitz injective:
\begin{equation}
 c_T \Gain^\ell |\Encoder_\Gain(\PhysicalState_a) - \Encoder_\Gain(\PhysicalState_b)| \ge |\PhysicalState_a - \PhysicalState_b|, \quad \forall \PhysicalState_a, \PhysicalState_b \in \X.
\end{equation}
As a consequence, there exists a globally Lipschitz left-inverse $\Encoder^{-1}_\Gain: \Encoder_\Gain(\X) \to \X$ mapping the latent state back to the physical state.
\end{theorem}
The proof is in Section \ref{app:proof_injectivity}.
Theorem \ref{thm:injectivity} is a fundamental result for our architecture. It proves that if we use a contracting Neural ODE with a high gain $\Gain$, the  image of $\Encoder_\Gain$ contains all the necessary information to uniquely compute $\PhysicalState(t)$. \startmodifVA Note that what has to be learned is not $\Encoder_\Gain$ but its inverse; the role of $\Encoder_\Gain$ in our analysis is to establish that this inverse exists, which makes learning the decoder well posed. Theorems \ref{thm:embedding} and \ref{thm:injectivity} being sufficient conditions, nothing says that a moderate gain fails; this is the regime our trained models operate in.

\begin{remark}[$\Encoder_\Gain$ as filtered output derivatives]
\label{rem:filtered_observability}
The map $\Encoder_\Gain$ is characterized by the nonlinear KKL equation $\frac{\partial \Encoder_\Gain}{\partial x}(x)\GenerativeDrift(x) = \Gain\, \BaseDynamics_\Params(\Encoder_\Gain(x), \GenerativeObs(x))$, which follows from the invariance property of Theorem \ref{thm:embedding}. Its relationship with the output and its successive time derivatives $(\GenerativeObs(x), \Lie\GenerativeObs(x),\dots,\Lie^\ell\GenerativeObs(x))$ appearing in Assumption \ref{ass:observability} is best seen in the linear case $\dot x = Fx$, $o = Hx$, $\BaseDynamics_\Params(z,o) = Az + Bo$ with $A$ Hurwitz. Then $\Encoder_\Gain$ is the unique solution of the Sylvester equation $\Encoder_\Gain F = \Gain(A \Encoder_\Gain + BH)$ and, for $\Gain$ large, it expands as
\begin{equation}\label{eq:sylvester_expansion}
\Encoder_\Gain\, x = -\sum_{i \ge 0} \Gain^{-i} A^{-(i+1)} B\, (H F^i x) ,
\end{equation}
where $HF^ix$ is the $i$-th derivative of the output. Hence $\Encoder_\Gain$ is not the observability matrix $(H; HF; \dots; HF^\ell)$ itself but a \textit{filtered} version of it: the $i$-th output derivative appears at scale $\Gain^{-i}$, weighted by the resolvent of the latent dynamics, and left-inverting the sum truncated at $i=\ell$ requires the left-invertibility of $[A^{-1}B,\dots,A^{-(\ell+1)}B]$, i.e., the rank condition of Assumption \ref{ass:filter_capacity}. This is the origin of the $\Gain^\ell$ factor in Theorem \ref{thm:injectivity} and of the requirement $m\ge p(\ell+1)$, and it explains the interest of the approach: instead of differentiating $o$ $\ell$ times, which is ill-posed under noise, the latent state carries the same information through a stable filter whose bandwidth is set by $\Gain$ (Proposition \ref{prop:iss}).
\end{remark}
\stopmodif
\section{Tunability and Decoder Complexity}
\label{sec:tunability}

In Section \ref{sec:architecture}, we established that a sufficiently large gain $\Gain$ guarantees the existence of a Lipschitz-continuous left-inverse map $\Encoder^{-1}_\Gain$. We now study the convergence of the estimation error when we use this inverse map to decode the latent state. Then, we analyze the practical case where this theoretical map is approximated by a neural network. %

\subsection{Convergence of the Ideal Observer}
\label{sec:ideal_convergence}

Let us first consider the ideal case where an exact left-inverse map $\Encoder^{-1}_\Gain$ is available. We define the state estimate as:
\begin{equation}
 \Estimate(t) = \Encoder^{-1}_\Gain(\LatentState(t)).
 \label{eq:ideal_estimate}
\end{equation}
We now prove that this estimator is formally \textit{tunable}. An observer is said to be tunable if, for any desired precision and any given time, we can choose its parameters to guarantee that the estimation error is smaller than this precision after this time.

\begin{theorem}[Exponential Convergence and Tunability]
\label{thm:ideal_tunability}
Under the assumptions of Theorem \ref{thm:injectivity} and given $\underline{\Gain}$ as in Theorem \ref{thm:injectivity}, there exists  $c'_T > 0$ such that for any $\Gain \ge \underline{\Gain}$, the state estimate $\Estimate(t)$ satisfies:
\begin{equation}
 |\Estimate(t) - \PhysicalState(t)| \le c_T' \Gain^\ell e^{-\Gain t} |\LatentState_0 - \Encoder_\Gain(\PhysicalState_0)|, \quad \forall t \ge 0,
 \label{eq:exponential_bound}
\end{equation}
for any physical trajectory $\PhysicalState(t)$ %
and any initial latent state $\LatentState_0 \in \Zcal$ and moreover, the architecture is tunable.
\end{theorem}
The proof is given in Section \ref{sec_ProofIdealTunability}.
This result highlights a fundamental trade-off. Increasing the gain $\Gain$ gives a faster exponential decay of the error. However, it may also increase the Lipschitz constant of the inverse map (which scales as $\Gain^\ell$), potentially amplifying the initial transient error before the exponential term dominates. \startmodifVA The same factor governs the sensitivity to measurement noise (Proposition \ref{prop:iss}).\stopmodif

\subsection{MLP Approximation and Complexity Bound}
\label{sec:mlp_complexity}

In practice, the theoretical map $\Encoder^{-1}_\Gain$ is not available analytically. Instead, we use a Multi-Layer Perceptron (MLP) as the decoder $\Decoder_\Params \approx \Encoder^{-1}_\Gain$, and the practical estimate is given by $\Estimate(t) = \Decoder_\Params(\LatentState(t))$. 

Note that the strict left-inverse $\Encoder^{-1}_\Gain$ is only defined on the image $\Encoder_\Gain(\X)$, which is a compact subset of $\Zcal$. By applying the McShane-Whitney extension theorem component-wise, this Lipschitz-continuous inverse can be extended to a globally Lipschitz function defined on the entire operating region $\Zcal$, preserving its Lipschitz constant $c_T\Gain^\ell$. The role of the MLP decoder $\Decoder_\Params$ is to approximate this extended map.

Since the extended $\Encoder^{-1}_\Gain$ is a Lipschitz-continuous function on a compact set, the Universal Approximation Theorem guarantees that an MLP can approximate it with arbitrary precision. We now quantify the required size of the network with respect to the gain $\Gain$ to ensure tunability.

\begin{theorem}[Neural Decoder Complexity]
\label{thm:neural_complexity}
Suppose the assumptions of Theorem \ref{thm:ideal_tunability} hold. For any desired precision $\epsilon > 0$ and any observation time $t_o > 0$, there exists a structural gain $\Gain > 0$ and a ReLU MLP decoder $\Decoder_\Params$ such that the neural estimate satisfies:
\begin{equation}
 |\Decoder_\Params(\LatentState(t)) - \PhysicalState(t)| \le \epsilon, \quad \forall t \ge t_o.
\end{equation}
Furthermore, the required complexity of the neural network (i.e., its total number of weights $\mathcal{W}$) grows polynomially with the gain $\Gain$ and the inverse precision $1/\epsilon$. More precisely, there exists a constant $C_{net} > 0$ (depending only on the dimensions $n$ and $m$) such that:
\begin{equation}
 \mathcal{W} \le C_{net} \left( \frac{\Gain^{\ell}}{\epsilon} \right)^{m} \log\left(\frac{\Gain^{\ell}}{\epsilon}\right).
 \label{eq:complexity_bound}
\end{equation}
\end{theorem}
The proof can be found in Section \ref{sec_ProofMLPComplexity}.
This theorem provides a formal justification for the  RNN-MLP architecture and shows that it remains tunable even when the exact inverse map is replaced by an MLP. However, it also clearly exposes the cost of this tunability: accelerating the continuous dynamics to improve the time-scale separation (increasing $\Gain$) makes the state-to-latent mapping more sensitive, requiring a larger network to learn its inverse with sufficient precision.

\startmodifVA 
\subsection{Robustness to measurement disturbances}
The following input-to-state estimate quantifies the sensitivity of the RNN-MLP estimator to measurement noise, and makes the classical high-gain trade-off explicit.
\begin{proposition}[Robustness to measurement disturbances]
\label{prop:iss}
Suppose the assumptions of Theorem \ref{thm:ideal_tunability} hold. Let \lb{$d_0>0$, let} $\bar L_{go} = \max\{ |\frac{\partial \BaseDynamics_\Params}{\partial \Observation}(\LatentState,\Observation)| : \LatentState\in\Zcal,\ \mathrm{dist}(\Observation,\Ycal)\le \lb{d_0}\}$ and $\bar d = \min\{\lb{d_0}, R_z/(4\bar L_{go})\}$. Let the latent dynamics in \eqref{eq:high_gain_ode} be driven by the corrupted observation $\obsPhys(t) + \delta(t)$, where $\delta$ is a Lebesgue measurable disturbance (no regularity in time is assumed) such that $\|\delta\|_\infty = \operatorname{ess\,sup}_{t\ge 0}|\delta(t)| \le \bar d$, and let the estimate be $\Estimate(t) = \Decoder_\Params(\LatentState(t))$, where the decoder $\Decoder_\Params$ satisfies $|\Decoder_\Params(\LatentState) - \Encoder^{-1}_\Gain(\LatentState)| \le \epsilon_{dec}$ for all $\LatentState \in \Zcal$, for some $\epsilon_{dec} \ge 0$ ($\epsilon_{dec} = 0$ for the exact inverse). Then $\Zcal$ remains forward invariant for the latent dynamics in \eqref{eq:high_gain_ode} and, for all $\Gain \ge \underline{\Gain}$ and any initial latent state $\LatentState_0 \in \Zcal$,
\begin{equation}\label{eq:iss_bound}
 |\Estimate(t) - \PhysicalState(t)| \le \epsilon_{dec} + c_T' \Gain^\ell e^{-\Gain t} |\LatentState_0 - \Encoder_\Gain(\PhysicalState_0)| + 2\, c_T' \Gain^\ell \bar L_{go}\, d(t), \quad \forall t \ge 0 ,
\end{equation}
where $d(t) = \operatorname{ess\,sup}_{s\in[0,t]} e^{-\Gain(t-s)/2}|\delta(s)| \le \|\delta\|_\infty$.
\end{proposition}
The proof is given in Section \ref{sec_ProofISS}. In particular, the ReLU MLP decoder of Theorem \ref{thm:neural_complexity} satisfies $|\Decoder_\Params(\LatentState(t)) - \PhysicalState(t)| \le \epsilon + 2c_T'\Gain^\ell \bar L_{go}\, d(t)$ for all $t \ge t_o$. The term $d(t)$ forgets past disturbances at rate $\Gain/2$: if $\delta$ vanishes after a time $t_1$, then $d(t) \le e^{-\Gain(t-t_1)/2}\|\delta\|_\infty$. For a persistent disturbance, the noise floor $2c_T'\Gain^\ell \bar L_{go} \|\delta\|_\infty$ is the high-gain effect: raising $\Gain$ accelerates the transient exponentially but raises the asymptotic error polynomially, with a degree set by the observability index $\ell$. The transient falls below the floor once $e^{-\Gain t_o} \approx \bar L_{go} \|\delta\|_\infty$, so for a target settling time $t_o$ the useful gain is $\Gain \approx t_o^{-1}\log(1/(\bar L_{go} \|\delta\|_\infty))$, beyond which any increase only raises the floor. In practice $\Gain$ is absorbed into the freely learned weights of $\BaseDynamics_\Params$ (as $(A,B)$ in numerical KKL methods); we fix $\Gain=1$ in the robotics experiments.\stopmodif

\section{Discrete-Time Implementation: Residual RNNs}
\label{sec:discrete}

This section formally establishes that the discrete-time implementation of our contracting latent dynamics, which naturally takes the form of a standard Residual Recurrent Neural Network (RNN), preserves the fundamental property of tunability established in the continuous-time case.

\myparagraph{System and Discrete Observations}.
While the theoretical analysis relies on the continuous-time generative model \eqref{eq:generative_model}, in practice, observations are collected at discrete time steps $t_k = k\TimeStep$, where $\TimeStep > 0$ is the sampling period. We denote the discrete observation sequence as:
\begin{equation}
 \obsPhys_k = \GenerativeObs(\flowf(\PhysicalState_0, t_k)).
\end{equation}

\myparagraph{Residual RNN as an Euler Discretization.}
The theoretical continuous-time observer is defined by $\dot{\LatentState} = \Gain \BaseDynamics_\Params(\LatentState, \Observation)$. Its discrete implementation via an explicit Euler numerical scheme is given by:
\begin{equation}
 \LatentState_{k+1} = \LatentState_k + \TimeStep \Gain \BaseDynamics_\Params(\LatentState_k, \obsPhys_k) = \text{RNN}_{\Params}(\LatentState_k, \obsPhys_k).
 \label{eq:euler_scheme}
\end{equation}
Note that corresponds to the architecture of a Residual RNN, where $\LatentState_k \in \LatentSpace$ acts as the hidden state. 
\begin{remark}[Contraction via Bounded Activations]
Consider the base dynamics $\BaseDynamics_\Params(\LatentState, \Observation) = -\LatentState + \tanh(W \LatentState + U \Observation + b)$, which is standard in RNN architectures. Since the hyperbolic tangent is bounded by $1$ element-wise, its Euclidean norm is bounded by $\sqrt{m}$. Let us define the compact set $\Zcal = \{ \LatentState \in \LatentSpace \mid |\LatentState| \le R \}$ with a radius $R > \sqrt{m}$. For any $\LatentState$ on the boundary of $\Zcal$, the linear dissipative term $-\LatentState$ strictly dominates the bounded term $\tanh(\dots)$, ensuring that the vector field points inward. The discrete update becomes $\LatentState_{k+1} = (1 - \TimeStep \Gain) \LatentState_k + \TimeStep \Gain \tanh(W \LatentState_k + U \obsPhys_k + b)$. The parameters $\Params = (W, U, b)$ can then be chosen to ensure the monotonicity condition \eqref{eq:contraction_condition} specifically for $\LatentState \in \Zcal$, which is a less restrictive constraint than global contraction.
\end{remark}
The discrete state estimate is finally decoded by the MLP:
\begin{equation}
 \Estimate_k = \Decoder_{\Params}(\LatentState_k) \approx \Encoder^{-1}_\Gain(\LatentState_k).
\end{equation}

\myparagraph{Discrete Tunability and Main Result.}
We now adapt our definition of tunability to the discrete context. Here, the sampling period $\TimeStep$ is treated as a design variable. Consequently, the concept of tunability requires the adjustment of this time step to achieve the desired accuracy.

\begin{definition}[Discrete Tunability]
The Residual RNN architecture is said to be tunable if, for any error threshold $\epsilon > 0$ and any observation time $t_o > 0$, there exist a structural gain $\Gain > 0$ and a maximum sampling step $\TimeStep^*$ such that, for all $\TimeStep \le \TimeStep^*$, the state estimate satisfies:
\begin{equation}
 |\Estimate_k - \PhysicalState(t_k)| \le \epsilon, \quad \forall k \text{ such that } t_k \ge t_o,
\end{equation}
for any physical initial condition $\PhysicalState_0 \in \X$ and any latent initialization $\LatentState_0 \in \Zcal$.
\end{definition}

\begin{theorem}[Tunability of the Residual RNN]
\label{thm:discrete_tunability}
Under the assumptions of Theorem \ref{thm:injectivity} (Observability, Latent Monotonicity, and Filter Rank Condition), the Residual RNN defined by \eqref{eq:euler_scheme} with the ideal decoder $\Estimate_k = \Encoder^{-1}_\Gain(\LatentState_k)$ is tunable.
\end{theorem}
The proof is in Section \ref{app:proof_euler}.
This theorem gives a formal justification for the use of RNNs in state estimation tasks: by choosing an architecture that respects local contraction, we can guarantee that an arbitrary precision is achievable provided the sampling frequency is sufficiently high.

\section{Experimental results}
\label{sec:experiments}

\vspace*{-1mm}
We provide experiments on two different types of data: a real robotics application trained from large-scale data, and a simulated system which allow us to verify theoretical claims.

\myparagraph{A robotics application: learning pose corrections from real large-scale data.}
Our objective is to estimate the pose of a mobile robot with respect to its starting position. It is equipped with sensors providing relative odometry $d_t = (\mathrm x_t, \mathrm y_t, \theta_t,v_t,\omega_t)^\top$ of pose and velocities obtained from wheel encoders and IMUs. We are interested in estimating the pose $\hat{s}_t$ in the coordinate frame of the episode start $t{=}0$, which classically can be obtained by kinematic integration of odometry readings over time (``\textit{Dead Reckoning''}), which is subject to drift: $\hat{s}_t^{raw} = \mathrm{Integrate}(\hat{s}^{raw}_{t-1}, d_t).$ %

\begin{figure}[t]
  \begin{minipage}[b]{0.19\textwidth}
    \centering
    \resizebox{!}{2.5cm}{%
    \begin{tikzpicture}
    \draw (0, 0) node[anchor=west,inner sep=0] {
        \includegraphics[width=4cm]{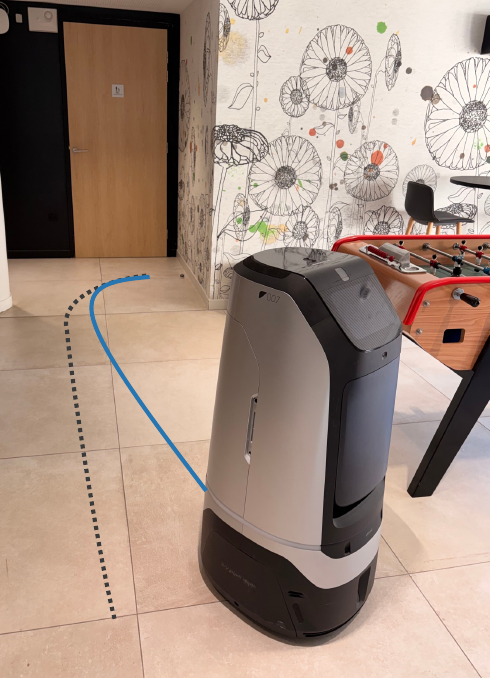}
    };    
    \draw (1.2,0.35) node[anchor=west,inner sep=0] {$s_0$};
    \draw (1,-2.4) node[anchor=west,inner sep=0] {$s_t^{raw}$};
    \draw (0.9,-1.4) node[anchor=west,inner sep=0] {\textcolor{colorCorr}{$\hat{s}^{corr}_t$}};
    \end{tikzpicture}}
  \end{minipage}
  \hfill
  \begin{minipage}[b]{0.31\textwidth}
    \centering
    \includegraphics[height=2.5cm]{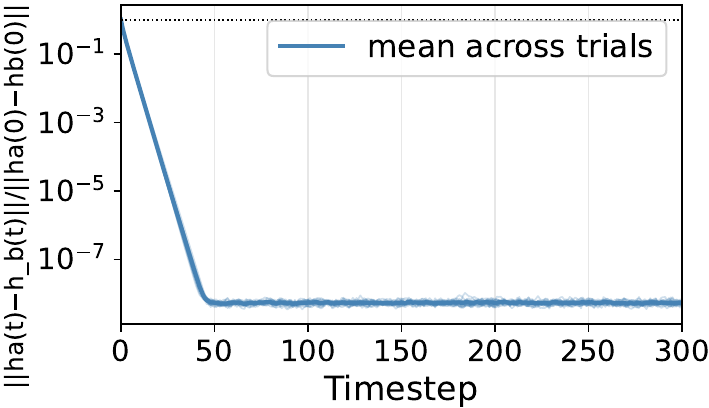}
  \end{minipage}
  \hfill
  \begin{minipage}[b]{0.46\textwidth}
    \centering
    \includegraphics[height=2.5cm]{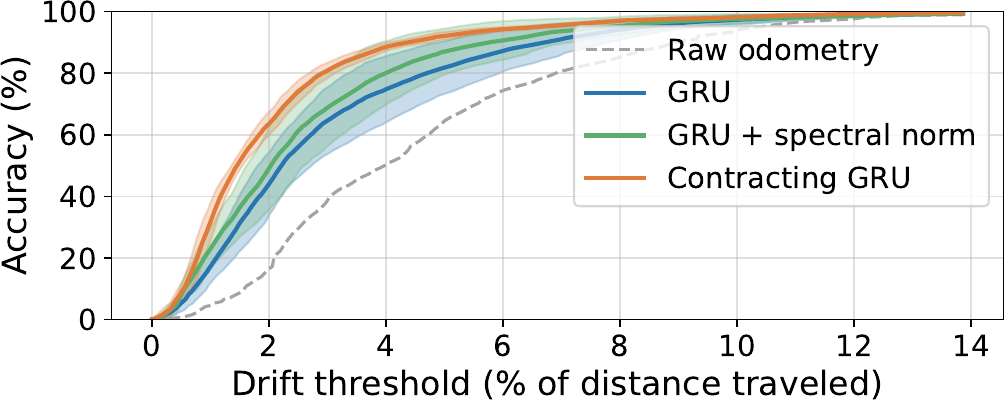}
  \end{minipage}
  \\[-4mm]
  \begin{minipage}[t]{0.19\textwidth}
    \caption{\label{fig:rookie}\raggedright\textbf{Robot}: raw \textit{vs} \textcolor{colorCorr}{corrected} trajectory.}
  \end{minipage}
  \hfill
  \begin{minipage}[t]{0.31\textwidth}
    \caption{\label{fig:gru_contracting}\textbf{Empirical contraction} of the \underline{classical} GRU. The network rapidly forgets its initial state.}
  \end{minipage}
  \hfill
  \begin{minipage}[t]{0.46\textwidth}
    \caption{\label{fig:comparison_gru_val}\textbf{Pose correction}: all models obtain strong gains but \textit{Contracting GRU} outperforms the unconstrained GRU and is more stable (mean $\pm$std-dev over 5 training runs).}
  \end{minipage}
  \vspace*{-5mm}
\end{figure}

We propose a learning based system to improve the odometry measurements through ``privileged information'' available during training but not during testing when the robot is deployed. We have collected a large-scale dataset of trajectories obtained with a real mobile service robot (cf. Figure \ref{fig:rookie}) in 4 different buildings on two continents spanning 14 
floors, 27km and 18h of navigation which was split into train/val/test, see Table \ref{tab:rookie_dataset} in Section \ref{app:rookie_datasets}. 
Beyond odometry the measurements also include localization from a Lidar based \textit{Adaptive Monte Carlo Localization} (AMCL) \cite{thrun2005probabilistic} system. These noisy pose readings $s^{\mathrm{amcl}}_t$ are thus available during training but not during testing. The goal is to learn a correction model such that the corrected trajectory $\hat{s}^{corr}_t \approx s^{\mathrm{amcl}}_t$, using only information available at runtime (no map).

The correction network learns a dynamical model of the robot and the odometry error capable of correcting pose measurements $d_t$ and replacing them with improved measurements $d_t + P \hat{y}_t$, where $\hat{y}_t \in \RR^3$ is a spatial correction term obtained by a recurrent network maintaining a hidden state $\LatentState_t$, and $P \in \RR^{5 \times 3}$ is a projection matrix mapping the spatial components to the full odometry state. The corrected trajectory is obtained by kinematic integration: 
\begin{align}
z_{t+1} = &  \varphi_{\Params}(\LatentState_t, {\Observation}_t) \\
\hat{y}_t = &  \gamma_\Omega(\LatentState_t), \\
\hat{{\TargetVar}}^{corr}_t = &  \mathrm{Integrate}(\hat{s}^{corr}_{t-1}, d_t + P \hat{y}_t),
\end{align}
where ${\Observation}_t$ is a 9-dimensional feature vector containing odometry readings, and command velocities,
\begin{equation}
  {\Observation}_t = \bigl[    
    \mathrm x_t,\;
    \mathrm y_t,\;
    \theta_t,\;
    v_t,\;
    \omega_t,\;
    v^{\mathrm{cmd}}_t,\;
    \omega^{\mathrm{cmd}}_t,\;
    v^{\mathrm{cmd}}_t - v_t,\;
    \omega^{\mathrm{cmd}}_t - \omega_t
  \bigr]
\end{equation}
The last two values are \emph{slip features} and encode the instantaneous mismatch between commanded and measured motion.
We propose a \textbf{trajectory integration loss} that directly penalizes accumulated drift. Trajectories are divided into non-overlapping windows of length $H{=}50$. Within each window, both the corrected and target body-frame displacements are integrated kinematically, and position MSE is computed on the resulting trajectories. Details on losses and training hyper-parameters are provided in Section \ref{app:rookie_losses}.

\myparagraph{Gated RNN-MLP architectures:}
we propose to address this problem with two different RNN architectures in their gated state-of-the-art variants: 
(i) ``\textbf{GRU}'', a vanilla GRU \cite{cho-etal-2014-learning} without any constraints on contraction, with 2-layers, hidden dim of 128, followed by a two-layer MLP head ($128{\to}64{\to}3$). 
(ii) ``\textbf{Contracting GRU}'', a two layer RNN with an update very much inspired by a GRU but with built-in contracting properties satisfying Assumption \ref{ass:contraction}. This is enforced with spectral normalization and choosing appropriate hyper-parameters, see Section \ref{app:rookie_arch} and \ref{app:rookie_hyperparameters}. (iii) \textbf{GRU + spectral norm} which ablates the effect of spectral normalization alone on the vanilla GRU. 
The filter rank condition of Assumption \ref{ass:filter_capacity} is numerically tested and validated in Section \ref{app:rookie_filter_rank}.

\myparagraph{The GRU learns contraction}. In Figure \ref{fig:gru_contracting} we provide empirical evidence that on this application the classical GRU without inductive bias on contraction properties does learn a contracting behavior. Two network copies $(a,b)$ are initialized with $z_a(0) \neq z_b(0)$ and driven by the same inputs. We show the normalized state distance $|z_a(t) - z_b(t)| / |z_a(0) - z_b(0)|$ on a log scale for 20 independent trials (light) and their mean. The network forgets its initial condition under any input.

\myparagraph{Results}. Figure \ref{fig:comparison_gru_val} gives a plot of  pose estimation accuracy as a function of the relative pose estimation error. An accuracy of 80\% at 2.0 means that 80\% of poses are correctly detected with an error which is lower than 2\% of the trajectory length. We can see that both ``GRU'' and ``Contracting GRU'' achieve strong gains, in particular for higher required precision (left part of the plot). However, the contracting GRU does outperform the classical GRU and, importantly, shows a far more stable training behavior when multiple seeds are compared.

\myparagraph{Simulated system.}
To complete the experimental part with a simulated system, we now consider a nonlinear oscillator  with unknown parameter:
\begin{equation}
    \dot{x}_1 = x_2, \quad
    \dot{x}_2 = -x_1 + (x_3 - x_1^2 - x_2^2)x_2 , \quad
    \dot{x}_3 =0
    \label{eq:vdp_aug}
\end{equation}
where $x = (x_1, x_2, x_3) \in \RR^3$ is the state (encompassing both the state $(x_1, x_2)$ of the oscillator and the unknown positive parameter $x_3$),
the observation is $o = x_1$
and
the target variable to be estimated is $s=x_3$.
In Appendix \ref{app:experiments_simulation}, we define a compact forward-invariant set $\mathcal{X}\subset \mathbb{R}^3$ of trajectories on which system \eqref{eq:vdp_aug} is differentially observable of order $\ell=3$ (see Assumption \ref{ass:observability}).

\begin{figure}[t!]
  \begin{minipage}[t]{0.45\textwidth}
    \centering
    \vspace{0pt}
    \includegraphics[width=\linewidth]{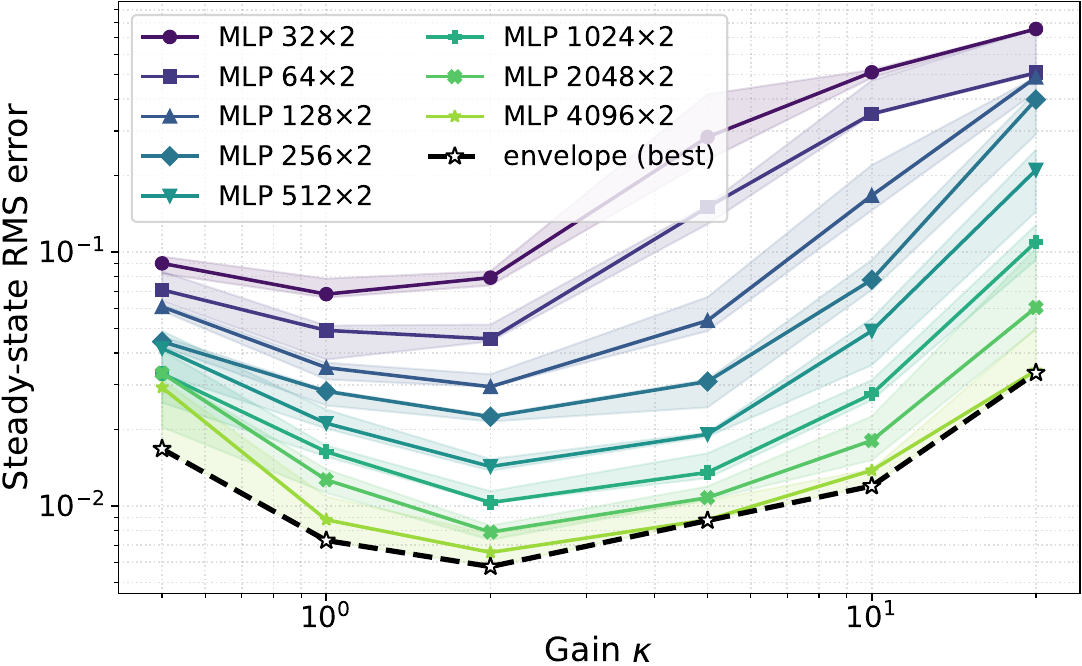}    \caption{\label{fig:gru_tuneability_capacity}Steady-state RMS vs. gain $\Gain$, for 2-layer MLP decoders of varying width $w$. %
    }     
  \end{minipage}
  \hfill
  \begin{minipage}[t]{0.43\textwidth}
    \centering
    \vspace{0pt}    
    \includegraphics[width=\linewidth]{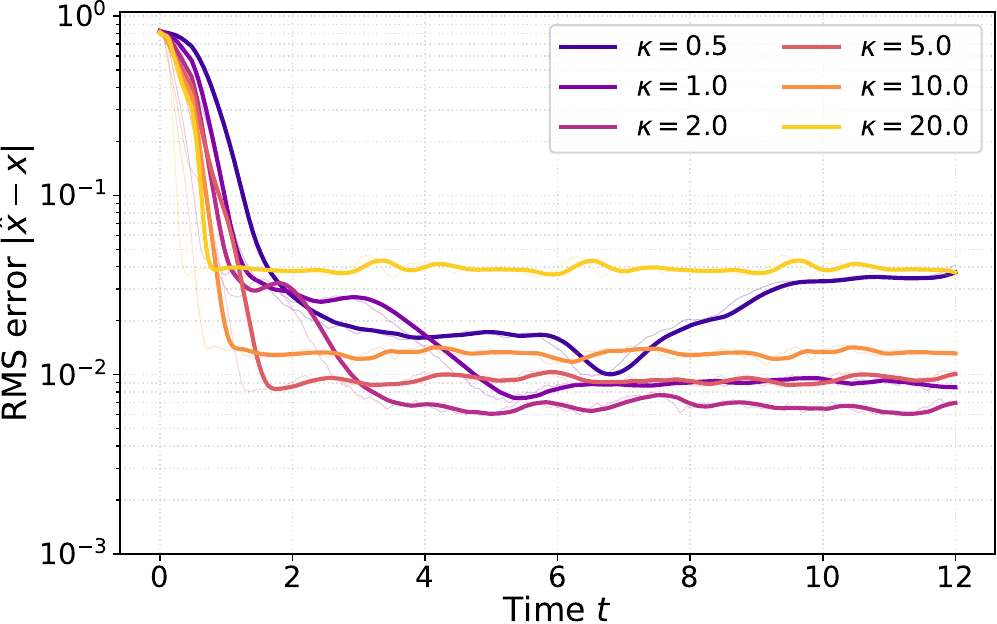}    \caption{\label{fig:tunability_largedecoder}Reconstruction error over time at varying $\Gain$, with the widest (=4096) MLP.}
  \end{minipage}
  \vspace*{-5mm}
\end{figure}

Figures \ref{fig:gru_tuneability_capacity} and \ref{fig:tunability_largedecoder} illustrate tunability, i.e. Theorem~\ref{thm:ideal_tunability}. The theoretical guarantee states that \emph{there exist} weights achieving the bound for large $\kappa$, so we train a separate observer for each value of $\Gain \in \{0.5, 1, 2, 5, 10, 20\}$ and each decoder width $w \in \{32, 64, 128, 256, 512, 1024, 2048, 4096\}$ and 3 seeds.
The capacity sweep (Fig.~\ref{fig:gru_tuneability_capacity}) shows that for every fixed width the error decreases up to $\Gain \approx 2$ and then re-grows, in line with the $\Gain^\ell$ decoder-Lipschitz cost predicted by Theorem~\ref{thm:neural_complexity}; widening the decoder lowers absolute error. 
The time-resolved view at the widest decoder (Fig.~\ref{fig:tunability_largedecoder}) confirms that larger $\Gain$ accelerates the transient, although a slight steady-state penalty remains at $\Gain = 20$.

\label{sec:experiments_simulation}

\vspace*{-3mm}
\startmodifVA \section{Discussion, limitations and conclusion}
\label{sec:conclusion}
\label{sec:limitations}
\vspace*{-3mm}
Viewing RNN-MLP estimators as time-discretized continuous systems, we extended nonlinear KKL observer theory from linear filters to contracting latent dynamics with dense Jacobians: the state-to-latent map preserves information under a generic rank condition, the architecture is tunable through a global gain $\Gain$, the decoder can be an MLP of quantified size, and these guarantees extend to the discrete-time implementation.
On a real pose-estimation task, a provably contracting model outperforms a baseline with increased training stability, and unconstrained GRUs converge to contracting dynamics, suggesting that contraction is an emergent property of well-trained sequence models.

\myparagraph{Scope of the assumptions.}
Differential observability (Assumption \ref{ass:observability}) is a property of the physical system, not of the method: for analytic observable systems it is necessary for the problem to be well posed, since no estimator, learned or model-based, can recover in arbitrarily short time what the observations do not contain. Contraction (Assumption \ref{ass:contraction}) holds by construction in the architecture of Appendix \ref{app:rookie_arch}, and the filter rank condition (Assumption \ref{ass:filter_capacity}) is generic (Appendix \ref{Sec_AssumpFilterCapacity}) and cheap to check after training by automatic differentiation (Appendix \ref{app:rookie_filter_rank}). Reconstructing only a function $\varsigma(x)$ of a non-observable state would require a functional observability framework \cite{montanari2022functional}, developed for KKL observers with linear latent dynamics \cite{spirito2022functional} but not yet with nonlinear ones.

\myparagraph{Noise and the curse of the observability index.}
Our generative model is deterministic. Proposition \ref{prop:iss} shows that contraction nevertheless yields robustness to bounded measurement disturbances, with an explicit noise floor in $\Gain^\ell$; the same exponent governs the decoder capacity $\mathcal{O}(\Gain^{\ell m})$ in Theorem \ref{thm:neural_complexity}. The observability index $\ell$ thus controls both faces of the trade-off between convergence speed and noise amplification or decoder complexity. A stochastic treatment (e.g. bounding the error variance under stochastic differential equations) is left for future work.

\myparagraph{Optimization and training dynamics.}
Our theorems establish the \textit{existence} of suitable parameters $\Params$ (any architecture enjoying a universal approximation theorem could replace the MLP in Theorem \ref{thm:neural_complexity}), not a constructive algorithm: enforcing the rank condition during training is non-trivial, and our evidence is empirical (Appendices \ref{app:rookie_arch}, \ref{app:rookie_filter_rank}). Training algorithms enforcing these conditions are a natural next step.\stopmodif

\newpage %

\bibliographystyle{plain} 
\bibliography{references,references_cw,references_cvpr2025} 

\begin{thebibliography}{10}

\bibitem{andrieu2018synchronization}
Vincent Andrieu, Bayu Jayawardhana, and Sophie Tarbouriech.
\newblock Some results on exponential synchronization of nonlinear systems.
\newblock {\em IEEE Transactions on Automatic Control}, 63(4):1213--1219, 2018.

\bibitem{AndrieuPraly2006}
Vincent Andrieu and Laurent Praly.
\newblock On the existence of a kazantzis--kravaris/luenberger observer.
\newblock {\em SIAM Journal on Control and Optimization}, 45(2):432--456, 2006.

\bibitem{jepa2}
Mido Assran, Adrien Bardes, David Fan, Quentin Garrido, Russell Howes, Mojtaba
  Komeili, Matthew Muckley, Ammar Rizvi, Claire Roberts, Koustuv Sinha, Artem
  Zholus, Sergio Arnaud, Abha Gejji, Ada Martin, Francois~Robert Hogan, Daniel
  Dugas, Piotr Bojanowski, Vasil Khalidov, Patrick Labatut, Francisco Massa,
  Marc Szafraniec, Kapil Krishnakumar, Yong Li, Xiaodong Ma, Sarath Chandar,
  Franziska Meier, Yann LeCun, Michael Rabbat, and Nicolas Ballas.
\newblock V-jepa 2: Self-supervised video models enable understanding,
  prediction and planning.
\newblock In {\em arxiv:2506.09985}, 2025.

\bibitem{bauersfeld2021neurobem}
Leonard Bauersfeld, Elia Kaufmann, Philipp Foehn, Sihao Sun, and Davide
  Scaramuzza.
\newblock Neurobem: Hybrid aerodynamic quadrotor model.
\newblock {\em arXiv preprint}, 2021.

\bibitem{beck2024xlstm}
Maximilian Beck, Korbinian P{\"o}ppel, Markus Spanring, Andreas Auer,
  Oleksandra Prudnikova, Michael Kopp, G{\"u}nter Klambauer, Johannes
  Brandstetter, and Sepp Hochreiter.
\newblock {xLSTM}: Extended long short-term memory.
\newblock In {\em Conference on Neural Information Processing Systems
  (NeurIPS)}, 2024.

\bibitem{BerAndAst}
Pauline Bernard, Vincent Andrieu, and Daniele Astolfi.
\newblock Observer design for continuous-time dynamical systems.
\newblock {\em Annual Reviews in Control}, 53:224--248, 2022.

\bibitem{brivadis2023further}
Lucas Brivadis, Vincent Andrieu, Pauline Bernard, and Ulysse Serres.
\newblock Further remarks on kkl observers.
\newblock {\em Systems \& Control Letters}, 172:105429, 2023.

\bibitem{buisson2023towards}
Mona Buisson-Fenet, Lukas Bahr, Valery Morgenthaler, and Florent Di~Meglio.
\newblock Towards gain tuning for numerical kkl observers.
\newblock {\em IFAC-PapersOnLine}, 56(2):4061--4067, 2023.

\bibitem{chen2018neural}
Ricky T.~Q. Chen, Yulia Rubanova, Jesse Bettencourt, and David~K. Duvenaud.
\newblock Neural ordinary differential equations.
\newblock In {\em Advances in Neural Information Processing Systems (NeurIPS)},
  volume~31, 2018.

\bibitem{cho-etal-2014-learning}
Kyunghyun Cho, Bart van Merri{\"e}nboer, Caglar Gulcehre, Dzmitry Bahdanau,
  Fethi Bougares, Holger Schwenk, and Yoshua Bengio.
\newblock Learning phrase representations using {RNN} encoder{--}decoder for
  statistical machine translation.
\newblock In {\em Conference on Empirical Methods in Natural Language
  Processing (EMNLP)}, 2014.

\bibitem{darouach2000functional}
Mohamed Darouach.
\newblock Existence and design of functional observers for linear systems.
\newblock {\em IEEE Transactions on Automatic Control}, 45(5):940--943, 2000.

\bibitem{GIULI2026113005}
Laura~Boca de~Giuli, Alessio~La Bella, Marcello Farina, and Riccardo
  Scattolini.
\newblock Stability and learning of networked systems with application to
  physics-informed neural networks.
\newblock {\em Automatica}, 189:113005, 2026.

\bibitem{de2007nonlinear}
Jos{\'e} de~Jes{\'u}s~Rubio and Wen Yu.
\newblock Nonlinear system identification with recurrent neural networks and
  dead-zone kalman filter algorithm.
\newblock {\em Neurocomputing}, 2007.

\bibitem{DBLP:conf/nips/DingFAP19}
Yiming Ding, Carlos Florensa, Pieter Abbeel, and Mariano Phielipp.
\newblock Goal-conditioned imitation learning.
\newblock In {\em Conference on Neural Information Processing Systems
  (NeurIPS)}, 2019.

\bibitem{fernando2010functional}
Tyrone~L. Fernando, Hieu~M. Trinh, and Les Jennings.
\newblock Functional observability and the design of minimum order linear
  functional observers.
\newblock {\em IEEE Transactions on Automatic Control}, 55(5):1268--1273, 2010.

\bibitem{DBLP:journals/corr/GravesWD14}
Alex Graves, Greg Wayne, and Ivo Danihelka.
\newblock Neural turing machines.
\newblock {\em arXiv preprint arXiv:1410.5401}, 2014.

\bibitem{gu2024mamba}
Albert Gu and Tri Dao.
\newblock Mamba: Linear-time sequence modeling with selective state spaces.
\newblock In {\em First Conference on Language Modeling}, 2024.

\bibitem{s4iclr2022}
Albert Gu, Karan Goel, and Christopher Ré.
\newblock Efficiently modeling long sequences with structured state spaces.
\newblock In {\em International Conference on Learning Representations (ICLR)},
  2022.

\bibitem{Hale1980}
J.~Hale.
\newblock {\em Ordinary differential equation}.
\newblock Krieger Publishing Company, Malabar, FL, second edition, 1980.

\bibitem{DBLP:journals/neco/HochreiterS97}
Sepp Hochreiter and J{\"{u}}rgen Schmidhuber.
\newblock Long short-term memory.
\newblock {\em Neural Computing}, 1997.

\bibitem{jaeger2001echo}
Herbert Jaeger.
\newblock The ``echo state'' approach to analysing and training recurrent
  neural networks.
\newblock Technical Report 148, German National Research Center for Information
  Technology (GMD), 2001.

\bibitem{janny2021deep}
Steeven Janny, Vincent Andrieu, Madiha Nadri, and Christian Wolf.
\newblock Deep {KKL}: Data-driven output prediction for non-linear systems.
\newblock In {\em 60th IEEE Conference on Decision and Control (CDC)}, pages
  4376--4381, 2021.

\bibitem{janny2024space}
Steeven Janny, Madiha Nadri, Julie Digne, and Christian Wolf.
\newblock Space and time continuous physics simulation from partial
  observations.
\newblock In {\em International Conference on Learning Representations (ICLR)},
  2024.

\bibitem{janny2025}
Steeven Janny, Hervé Poirier, Leonid Antsfeld, Guillaume Bono, Gianluca
  Monaci, Boris Chidlovskii, Francesco Giuliari, Alessio~Del Bue, and Christian
  Wolf.
\newblock Reasoning in visual navigation of end-to-end trained agents: a
  dynamical systems approach.
\newblock In {\em CVPR}, 2025.

\bibitem{neuralgpu2016}
{\L}ukasz Kaiser and Ilya Sutskever.
\newblock Neural gpus learn algorithms.
\newblock In {\em International Conference on Learning Representations (ICLR)},
  2016.

\bibitem{KazantzisKravaris1998}
N.~Kazantzis and C.~Kravaris.
\newblock Nonlinear observer design using {Lyapunov's} auxiliary theorem.
\newblock {\em Systems \& Control Letters}, 34(5):241--247, 1998.

\bibitem{Lakshmikantham-Leela}
V.~Lakshmikantham and S.~Leela.
\newblock {\em Differential and Integral Inequalities: Theory and
  Applications}, volume~1 of {\em Mathematics in Science and Engineering}.
\newblock Academic Press, New York, 1969.
\newblock Ordinary Differential Equations.

\bibitem{lennart1999system}
Ljung Lennart.
\newblock System identification: theory for the user.
\newblock {\em PTR Prentice Hall, Upper Saddle River, NJ}, 28:540, 1999.

\bibitem{StateRepFiliat2018}
T.~Lesort, N.~Díaz-Rodríguez, J.-F. Goudou, and D.~Filliat.
\newblock {State Representation Learning for Control: An Overview}.
\newblock {\em Neural Networks}, 2018.

\bibitem{9835013}
Michael~Di Loreto and Damien Eberard.
\newblock Strong left inversion of linear systems and input reconstruction.
\newblock {\em IEEE Transactions on Automatic Control}, 68(6):3612--3617, 2023.

\bibitem{lu2017reservoir}
Zhixin Lu, Jaideep Pathak, Brian Hunt, Michelle Girvan, Roger Brockett, and
  Edward Ott.
\newblock Reservoir observers: Model-free inference of unmeasured variables in
  chaotic systems.
\newblock {\em Chaos: An Interdisciplinary Journal of Nonlinear Science},
  27(4):041102, 2017.

\bibitem{lukovsevivcius2009reservoir}
Mantas Luko{\v{s}}evi{\v{c}}ius and Herbert Jaeger.
\newblock Reservoir computing approaches to recurrent neural network training.
\newblock {\em Computer science review}, 3(3):127--149, 2009.

\bibitem{MarFilAlNLal}
Yasmine Marani, Israel Filho, Tareq Al-Naffouri, and Taous~Meriem Laleg~Kirati.
\newblock Unsupervised physics-informed neural network-based nonlinear observer
  design for autonomous systems using contraction analysis.
\newblock In {\em 2025 European Control Conference (ECC)}, pages 2327--2332,
  2025.

\bibitem{miao2023learning}
Keyan Miao and Konstantinos Gatsis.
\newblock Learning robust state observers using neural odes.
\newblock In {\em Learning for Dynamics and Control Conference}, pages
  208--219. PMLR, 2023.

\bibitem{Mityagin2020}
Boris~S. Mityagin.
\newblock The zero set of a real analytic function.
\newblock {\em Mathematical Notes}, 107(3):529--530, 2020.

\bibitem{montanari2022functional}
Arthur~N. Montanari, Chao Duan, Luis~A. Aguirre, and Adilson~E. Motter.
\newblock Functional observability and target state estimation in large-scale
  networks.
\newblock {\em Proceedings of the National Academy of Sciences},
  119(1):e2113750119, 2022.

\bibitem{niazi2023learning}
Muhammad Umar~B Niazi, John Cao, Xudong Sun, Amritam Das, and Karl~Henrik
  Johansson.
\newblock Learning-based design of luenberger observers for autonomous
  nonlinear systems.
\newblock In {\em 2023 American Control Conference (ACC)}, pages 3048--3055.
  IEEE, 2023.

\bibitem{nijmeijer1997observer}
Henk Nijmeijer and Iven M.~Y. Mareels.
\newblock An observer looks at synchronization.
\newblock {\em IEEE Transactions on Circuits and Systems I: Fundamental Theory
  and Applications}, 44(10):882--890, 1997.

\bibitem{lru2023}
Antonio Orvieto, Samuel~L Smith, Albert Gu, Anushan Fernando, Caglar Gulcehre,
  Razvan Pascanu, and Soham De.
\newblock {Resurrecting Recurrent Neural Networks for Long Sequences}.
\newblock In {\em International Conference on Machine Learning (ICML)}, 2023.

\bibitem{pachy2024existence}
Victor Pachy, Vincent Andrieu, Pauline Bernard, Lucas Brivadis, and Laurent
  Praly.
\newblock On the existence of kkl observers with nonlinear contracting
  dynamics.
\newblock {\em IFAC-PapersOnLine}, 58(21):262--267, 2024.

\bibitem{peralez2021deep}
J.~Peralez and M.~Nadri.
\newblock Deep learning-based luenberger observer design for discrete-time
  nonlinear systems.
\newblock In {\em 2021 60th IEEE Conference on Decision and Control (CDC)},
  pages 1932--1937. IEEE, 2021.

\bibitem{pfaff2020learning}
Tobias Pfaff, Meire Fortunato, Alvaro Sanchez-Gonzalez, and Peter Battaglia.
\newblock Learning mesh-based simulation with graph networks.
\newblock In {\em International Conference on Learning Representations}, 2020.

\bibitem{humanoidlocomotiontokenprediction2024}
Ilija Radosavovic, Bike Zhang, Baifeng Shi, Jathushan Rajasegaran, Sarthak
  Kamat, Trevor Darrell, Koushil Sreenath, and Jitendra Malik.
\newblock Humanoid locomotion as next token prediction.
\newblock In {\em NeurIPS}, 2024.

\bibitem{ramos2020numerical}
L.~d.~C. Ramos, F.~Di~Meglio, V.~Morgenthaler, L.~F.~F. da~Silva, and
  P.~Bernard.
\newblock Numerical design of luenberger observers for nonlinear systems.
\newblock In {\em 2020 59th IEEE Conference on Decision and Control (CDC)},
  pages 5328--5333. IEEE, 2020.

\bibitem{ryoo2023tokenturingmachines}
Michael~S. Ryoo, Keerthana Gopalakrishnan, Kumara Kahatapitiya, Ted Xiao,
  Kanishka Rao, Austin Stone, Yao Lu, Julian Ibarz, and Anurag Arnab.
\newblock Token turing machines.
\newblock In {\em IEEE Conference on Computer Vision and Pattern Recognition
  (CVPR)}, 2023.

\bibitem{kinaema2025}
Mert~Bulent Sariyildiz, Philippe Weinzaepfel, Guillaume Bono, Gianluca Monaci,
  and Christian Wolf.
\newblock Kinaema: a recurrent sequence model for memory and pose in motion.
\newblock In {\em Conference on Neural Information Processing Systems
  (NeurIPS)}, 2025.

\bibitem{timeconstrastive2018}
Pierre Sermanet, Corey Lynch, Yevgen Chebotar, Jasmine Hsu, Eric Jang, Stefan
  Schaal, and Sergey Levine.
\newblock Time-contrastive networks: Self-supervised learning from video.
\newblock In {\em IEEE Conference on Computer Vision and Pattern Recognition
  (CVPR)}, 2018.

\bibitem{spirito2022functional}
Mario Spirito, Pauline Bernard, and Lorenzo Marconi.
\newblock On the existence of robust functional {KKL} observers.
\newblock In {\em 2022 American Control Conference (ACC)}, pages 2982--2987,
  2022.

\bibitem{curl2020}
Aravind Srinivas, Michael Laskin, and Pieter Abbeel.
\newblock Curl: Contrastive unsupervised representations for reinforcement
  learning.
\newblock In {\em International Conference on Machine Learning (ICML)}, 2020.

\bibitem{stubberud1991neural}
A~Stubberud, H~Wabgaonkar, and S~Stubberud.
\newblock A neural-network-based system identification technique.
\newblock In {\em Conference on Decision and Control}, 1991.

\bibitem{endtoendmemorynetworks2015}
Sainbayar Sukhbaatar, Arthur Szlam, Jason Weston, and Rob Fergus.
\newblock End-to-end memory networks.
\newblock In {\em Conference on Neural Information Processing Systems
  (NeurIPS)}, 2015.

\bibitem{Szarski}
Jacek Szarski.
\newblock {\em Differential Inequalities}, volume~43 of {\em Monografie
  Matematyczne}.
\newblock Polish Scientific Publishers, Warsaw, 1965.

\bibitem{tang2024synthesis}
Wentao Tang.
\newblock Synthesis of data-driven nonlinear state observers using
  lipschitz-bounded neural networks.
\newblock In {\em 2024 American Control Conference (ACC)}, pages 1713--1719.
  IEEE, 2024.

\bibitem{thrun2005probabilistic}
Sebastian Thrun, Wolfram Burgard, and Dieter Fox.
\newblock {\em Probabilistic robotics}.
\newblock MIT press Cambridge, 2005.

\bibitem{moog2024}
Sjoerd van Steenkiste, Daniel Zoran, Yi~Yang, Yulia Rubanova, Rishabh Kabra,
  Carl Doersch, Dilara Gokay, Joseph Heyward, Etienne Pot, Klaus Greff, Drew~A.
  Hudson, Thomas~Albert Keck, Joao Carreira, Alexey Dosovitskiy, Mehdi S.~M.
  Sajjadi, and Thomas Kipf.
\newblock Moving off-the-grid: Scene-grounded video representations.
\newblock In {\em Conference on Neural Information Processing Systems
  (NeurIPS)}, 2024.

\bibitem{van2020data}
Henk~J Van~Waarde, Jaap Eising, Harry~L Trentelman, and M~Kanat Camlibel.
\newblock Data informativity: A new perspective on data-driven analysis and
  control.
\newblock {\em IEEE Transactions on Automatic Control}, 65(11):4753--4768,
  2020.

\bibitem{van2022kernel}
Henk~J Van~Waarde and Rodolphe Sepulchre.
\newblock Kernel-based models for system analysis.
\newblock {\em IEEE Transactions on Automatic Control}, 68(9):5317--5332, 2022.

\bibitem{Yarotsky2017}
Dmitry Yarotsky.
\newblock Error bounds for approximations with deep relu networks.
\newblock {\em Neural Networks}, 94:103--114, 2017.
\newblock See also arXiv:1610.01145.

\bibitem{zoboli2023deep}
Samuele Zoboli, Steeven Janny, and Mattia Giaccagli.
\newblock Deep learning-based output tracking via regulation and contraction
  theory.
\newblock In {\em International Federation of Automatic Control (IFAC)}, 2023.

\end{thebibliography}

\newpage
\appendix

\begin{center}
 \Large \textbf{Supplementary Material for:\\ Provable State Estimation with Recurrent Models}
\end{center}

\startmodifVA \section{Roadmap of the Proofs}
\label{app:roadmap}

The appendix is long because the paper is a theoretical contribution and the proofs belong in it. This page gives the structure of the argument and the dependencies between its pieces, so that a reader can navigate the proofs or skip the ones they do not need.

\myparagraph{Overall structure.}
The main results form a chain, each one building on the previous:
\begin{center}
\small
Lemma \ref{lemma-dif-ineg} (Appendix \ref{sec:result-dif-ineq})
$\;\Rightarrow\;$ Prop.\ \ref{prop:invariance} (Appendix \ref{app:proof_invariance})
$\;\Rightarrow\;$ Thm.\ \ref{thm:embedding} (Appendix \ref{app:proof_embedding})
$\;\Rightarrow\;$ Thm.\ \ref{thm:injectivity} (Appendix \ref{app:proof_injectivity})
$\;\Rightarrow\;$ Thms.\ \ref{thm:ideal_tunability}, \ref{thm:neural_complexity} and Prop.\ \ref{prop:iss} (Appendix \ref{app:proof_decoder})
$\;\Rightarrow\;$ Thm.\ \ref{thm:discrete_tunability} (Appendix \ref{app:proof_euler}).
\end{center}

\myparagraph{Appendix \ref{sec:result-dif-ineq}: a differential inequality.}
The main technical tool used throughout is Lemma \ref{lemma-dif-ineg}: a matrix solution of $\dot{\EuScript M} = A(t)\EuScript M + B(t)$ whose symmetric part $A + A^\top$ is bounded above by $2a_+ I$ satisfies $|\EuScript M(t)| \le e^{a_+ (t-t_0)}|\EuScript M(t_0)| + \int_{t_0}^t e^{a_+(t-s)}|B(s)|ds$. Every contraction, attractivity and robustness estimate of the paper is an instance of this lemma with a different choice of $\EuScript M$, $A$ and $B$. \startmodifVA Lemma \ref{lemma-invariance-ball} complements it with an elementary invariance result for a ball, valid for inputs that are only measurable in time.\stopmodif

\myparagraph{Appendix \ref{app:proof_invariance}: contraction and invariance (Proposition \ref{prop:invariance}).}
The monotonicity of Assumption \ref{ass:contraction} is integrated along segments (mean value theorem) to show that (i) the vector field points inward on the boundary of the ball $\Zcal$, hence $\Zcal$ is forward invariant (Lemma \ref{lemma-invariance-ball}), and (ii) the difference of two trajectories obeys a linear time-varying equation with symmetric part $\le -2\Gain I$, hence contracts at rate $\Gain$ by Lemma \ref{lemma-dif-ineg}. A corollary (Appendix \ref{app:remark_invariance_perturbed}) shows that the invariance survives an additive perturbation of the vector field of magnitude at most $\bar{\BaseDynamics}$; it is reused in the injectivity and robustness proofs.

\myparagraph{Appendix \ref{app:proof_embedding}: existence and regularity of $\Encoder_\Gain$ (Theorem \ref{thm:embedding}).}
The map is defined as $\Encoder_\Gain(x_0) = \varpi(0)$, where $\varpi$ is the unique solution of the latent dynamics that is bounded on the whole time axis when driven by the output of the physical trajectory through $x_0$, extended to negative times. Existence and uniqueness of $\varpi$ follow from contraction (two bounded solutions must coincide as $\tau\to-\infty$). Invariance and attractivity are then direct consequences of Proposition \ref{prop:invariance}. Regularity ($C^1$ with Lipschitz derivative) is the delicate part: the variational equation is analyzed with Lemma \ref{lemma-dif-ineg}, and the condition $\Gain > 2 L_\GenerativeDrift$ ensures that the latent contraction dominates the expansion of the physical flow, so that the family of finite-horizon approximations is equi-Lipschitz and converges by the Ascoli-Arzel\`a theorem.

\myparagraph{Appendix \ref{app:proof_injectivity}: Lipschitz injectivity (Theorem \ref{thm:injectivity}).}
This is the heart of the paper and the point where the diagonality restriction of \cite{pachy2024existence} is lifted. Working in the observer time $\tau = \Gain t$, in which the observation is slowly varying (its $i$-th derivative is $\mathcal{O}(\Gain^{-i})$), the bounded solution $\varpi$ is approximated by an explicit sequence $\varphi_0, \dots, \varphi_\ell$ of functions of the observation and its first derivatives, built recursively by inverting the base dynamics under a constant forcing (implicit function $\psi$, Lemma \ref{lem:def_psi}; boundedness, Lemma \ref{lem:boundedness_phi}). Three facts are then established: the approximation map is Lipschitz injective with respect to the vector of observation derivatives, with constant of order $\Gain^{\ell}$, thanks to the rank condition of Assumption \ref{ass:filter_capacity} (Proposition \ref{prop:approx_map_injectivity}); the approximation error is of order $\Gain^{-(\ell+1)}$ (Proposition \ref{prop:approx_quality}); and the difference of two error trajectories obeys a fading-memory bound (Proposition \ref{prop:derivative_reconstruction}). The synthesis combines these with differential observability (Assumption \ref{ass:observability}) and a large enough $\Gain$ to conclude.

\myparagraph{Appendix \ref{app:proof_decoder}: decoding (Theorems \ref{thm:ideal_tunability}, \ref{thm:neural_complexity} and Proposition \ref{prop:iss}).}
The left inverse of $\Encoder_\Gain$ is extended to the whole latent space by the McShane-Whitney theorem, preserving its Lipschitz constant up to a factor $\sqrt{n}$. Tunability with the exact inverse is then a one-line consequence of attractivity; the MLP complexity bound follows from the approximation theorem of \cite{Yarotsky2017} for Lipschitz functions; and the robustness estimate is obtained by showing that $\Zcal$ remains invariant under the disturbance and comparing the observer driven by the corrupted output with the nominal one, along the lines of Lemma \ref{lemma-dif-ineg}.

\myparagraph{Appendix \ref{app:proof_euler}: discrete time (Theorem \ref{thm:discrete_tunability}).}
The residual RNN is an explicit Euler scheme for the continuous observer. A small-step analysis shows that the discrete update inherits the contraction for $\TimeStep\Gain$ small enough, and that the global Euler error on the finite horizon $[0,t_o]$ vanishes with $\TimeStep$; combined with Theorem \ref{thm:ideal_tunability} this gives discrete tunability.

\myparagraph{Appendices \ref{app:rookie} to \ref{Sec_AssumpFilterCapacity}: experiments and genericity.}
Appendix \ref{app:rookie} details the robotics experiments (losses, the Contracting GRU and its contraction certificate, datasets, hyper-parameters, empirical verification of the contraction constant and of the rank condition), Appendix \ref{app:experiments_simulation} the simulated system (in particular its differential observability), and Appendix \ref{Sec_AssumpFilterCapacity} proves that the rank condition of Assumption \ref{ass:filter_capacity} is generically satisfied by analytic recurrent models.\stopmodif

\section{A Result on Differential Inequalities}
\label{sec:result-dif-ineq}

To ease the reading of our proofs, we start by reminding the reader of a standard result on differential inequalities.

In what follows, we denote, for a matrix $\EuScript{M} \in \RR^{n \times m}$, the induced Euclidean norm:
\begin{equation}
 |\EuScript{M}| = \max_{v \in \RR^m : |v|=1} |\EuScript{M} v|
\end{equation}

\begin{lemma}[See for instance \cite{Szarski} or \cite{Lakshmikantham-Leela}]
\label{lemma-dif-ineg}
Let $\EuScript{M}$ be \startmodifVA an absolutely continuous \stopmodif matrix function satisfying the linear ODE
\begin{equation}
 \frac{d\EuScript{M}}{dt} = A(t) \EuScript{M} + B(t)
\end{equation}
\startmodifVA for almost all $t$, \stopmodif where $A: \RR \to \RR^{n \times n}$ and $B: \RR \to \RR^{n \times m}$ are \startmodifVA measurable and locally essentially bounded \stopmodif matrix functions satisfying\startmodifVA, for almost all $t$,\stopmodif
\begin{equation}
 2a_{\!_-} I_n \leq A(t) + A(t)^\top \leq 2 a_{\!_+} I_n
\end{equation}
where $a_{\!_+}$ and $a_{\!_-}$ are constant real numbers, with $a_{\!_-}$ possibly $-\infty$, and $I_n$ is the identity matrix of dimension $n$.

Then we have, for all $t_0 \leq t$,
\begin{equation}
 e^{a_{\!_-} (t-t_0)}|\EuScript{M}(t_0)| - \int_{t_0}^t e^{a_{\!_-} (t-s)}|B(s)| \, ds
 \leq 
 |\EuScript{M}(t)| \leq e^{a_{\!_+} (t - t_0)} |\EuScript{M}(t_0)| + \int_{t_0}^t e^{a_{\!_+} (t -s)} |B(s)| \, ds
\end{equation}
\end{lemma}

For example, when $a_{\!_-} = -a_{\!_+}$ and $B=0$, we get (using $t_0=s\leq 0$ and $t=0$ for the left inequality, and $t_0=0$ and $t=s\geq 0$ for the right inequality):
\begin{equation}
 e^{-a_{\!_+} (-s)}|\EuScript{M}(s)| \leq |\EuScript{M}(0)| \quad \forall s \leq 0,
 \qquad \text{and} \qquad 
 |\EuScript{M}(s)| \leq e^{a_{\!_+} s} |\EuScript{M}(0)| \quad \forall s \geq 0,
\end{equation}
which implies:
\begin{equation}
 |\EuScript{M}(s)| \leq e^{a_{\!_+} |s|} |\EuScript{M}(0)| \qquad \forall s \in \RR.
\end{equation}

\begin{proof}
From the ODE, we get readily for any unit vector $v \in \RR^m$\startmodifVA{} and almost all $t$\stopmodif,
\begin{equation*}
 \frac{d}{dt}\left\{v^\top\EuScript{M}^\top \EuScript{M} v\right\}
 = v^\top \EuScript{M}^\top (A(t)+A(t)^\top)\EuScript{M} v + 2 v^\top \EuScript{M}^\top B(t) v.
\end{equation*}
Using the bounds on the symmetric part of $A(t)$ and the Cauchy-Schwarz inequality for the second term, we have when $\EuScript{M} v \neq 0$:
\begin{equation*}
 2a_{\!_-} |\EuScript{M} v|^2 - 2|\EuScript{M}v| |B(t)v|
 \leq 
 2|\EuScript{M}v| \frac{d}{dt} \left\{ |\EuScript{M} v| \right\}
 \leq 
 2a_{\!_+} |\EuScript{M} v|^2 + 2|\EuScript{M}v| |B(t)v|.
\end{equation*}
Dividing by $2|\EuScript{M}v|$ yields:
\begin{equation*}
 a_{\!_-} |\EuScript{M} v| - |B(t)v|
 \leq 
 \frac{d}{dt} \left\{ |\EuScript{M} v| \right\}
 \leq 
 a_{\!_+} |\EuScript{M} v| + |B(t)v|.
\end{equation*}
\startmodifVA The function $t \mapsto |\EuScript{M}(t) v|$ is absolutely continuous, as the composition of the Lipschitz function $|\cdot|$ with an absolutely continuous function, hence differentiable almost everywhere. At a time where $\EuScript{M}(t) v = 0$ and this function is differentiable, its derivative vanishes, since $0$ is its minimum value; the above inequalities therefore hold for almost all $t$, including when $\EuScript{M} v = 0$. Since $t\mapsto e^{-a_{\!_+}t}|\EuScript{M}(t) v|$ and $t\mapsto e^{-a_{\!_-}t}|\EuScript{M}(t) v|$ are absolutely continuous, it follows that, for almost all $t$:\stopmodif
\begin{equation*}
 \frac{d}{dt} \left\{e^{-a_{\!_+}t}|\EuScript{M} v|\right\} \leq e^{-a_{\!_+}t}|B(t)v|
 \qquad \text{and} \qquad 
 - e^{-a_{\!_-}t}|B(t)v| \leq \frac{d}{dt} \left\{e^{-a_{\!_-}t}|\EuScript{M} v|\right\}.
\end{equation*}
By integration, we obtain for all $t_0 \leq t$ and for all $v$ such that $|v|=1$:
\begin{align*}
 e^{-a_{\!_+} t}|\EuScript{M}(t) v| - e^{-a_{\!_+} t_0}|\EuScript{M}(t_0) v| &\leq \int_{t_0}^t e^{-a_{\!_+}s}|B(s)v| \, ds, \\
 - \int_{t_0}^t e^{-a_{\!_-} s}|B(s)v| \, ds &\leq e^{-a_{\!_-} t}|\EuScript{M}(t) v| - e^{-a_{\!_-} t_0}|\EuScript{M}(t_0) v|.
\end{align*}
In particular, for $\bar{v}_+(t)$ maximizing $|\EuScript{M}(t) v|$, multiplying the first inequality by $e^{a_{\!_+} t}$ gives:
\begin{align*}
 |\EuScript{M}(t)|
 &\leq e^{a_{\!_+} (t-t_0)}|\EuScript{M}(t_0) \bar{v}_+(t)| + \int_{t_0}^t e^{a_{\!_+}(t-s)}|B(s) \bar{v}_+(t)| \, ds \\
 &\leq e^{a_{\!_+} (t-t_0)}|\EuScript{M}(t_0)| + \int_{t_0}^t e^{a_{\!_+} (t-s)}|B(s)| \, ds \qquad \forall t_0 \leq t.
\end{align*}
Similarly, for $\bar{v}_-(t_0)$ maximizing $|\EuScript{M}(t_0) v|$ (which implies $|\EuScript{M}(t)\bar{v}_-(t_0)| \leq |\EuScript{M}(t)|$), multiplying the second inequality by $e^{a_{\!_-} t}$ gives:
\begin{align*}
 |\EuScript{M}(t)| \geq |\EuScript{M}(t) \bar{v}_-(t_0)| 
 &\geq e^{a_{\!_-} (t-t_0)}|\EuScript{M}(t_0)\bar{v}_-(t_0)| - \int_{t_0}^t e^{a_{\!_-} (t-s)}|B(s) \bar{v}_-(t_0)| \, ds \\
 &\geq e^{a_{\!_-} (t-t_0)}|\EuScript{M}(t_0)| - \int_{t_0}^t e^{a_{\!_-} (t-s)}|B(s)| \, ds \qquad \forall t_0 \leq t.
\end{align*}
\end{proof}

\startmodifVA
We also use the following invariance result for a ball, which is a particular case of the Nagumo-type characterizations of invariant sets, in the setting of an input that is only measurable in time.

\begin{lemma}[Invariance of a ball under measurable inputs]
\label{lemma-invariance-ball}
Let $R > 0$, let $\mathcal{U} \subset \RR^p$ be compact, and let $G : \RR^m \times \RR^p \to \RR^m$ be continuous, with $\frac{\partial G}{\partial z}$ continuous on $\RR^m \times \RR^p$. Assume that
\begin{equation}\label{eq:ball_inward}
 z^\top G(z, u) \le 0, \qquad \forall z \in \RR^m \text{ with } |z| = R, \ \forall u \in \mathcal{U}.
\end{equation}
Let $u : [0, +\infty) \to \RR^p$ be Lebesgue measurable with $u(t) \in \mathcal{U}$ for almost all $t \ge 0$. Then the solution of $\frac{dz}{dt} = G(z, u(t))$, understood in the sense of Carath\'eodory, with $|z(0)| \le R$, is defined on $[0, +\infty)$ and satisfies $|z(t)| \le R$ for all $t \ge 0$.
\end{lemma}

\begin{proof}
The right-hand side $(t, z) \mapsto G(z, u(t))$ is measurable in $t$, continuous in $z$, and, on any compact set of $z$, bounded and Lipschitz in $z$ uniformly in $t$, since $u(t)$ lies in the compact set $\mathcal{U}$ for almost all $t$ and $\frac{\partial G}{\partial z}$ is continuous. By Carath\'eodory's theorem \cite[Chapter I.5]{Hale1980}, there is a unique solution, defined on a maximal interval $[0, t_{\max})$, and $t_{\max} = +\infty$ if it remains in a compact set.

\textit{Step 1: strict inequality.} Assume first that \eqref{eq:ball_inward} holds with a strict inequality. By continuity of $G$ and compactness, $\mu = -\max\{ z^\top G(z, u) : |z| = R, \ u \in \mathcal{U}\}$ is positive. 
Moreover, by continuity of $G$ and compactness again, there exists $\eta>0$ such that 
\begin{equation}\label{eq:ball_shell}
 z^\top G(z, u) \le -\frac{\mu}{2}, \qquad \forall z \text{ s.t. } R \le |z| \le R + \eta, \ \forall u \in \mathcal{U}.
\end{equation}
Assume by contradiction that $|z(t_+)| > R$ for some $t_+ \in [0, t_{\max})$, and let $t^* = \sup\{ t \in [0, t_+) : |z(s)| \le R, \ \forall s \in [0, t]\}$. Then $t^* \le t_+ < t_{\max}$ and, by continuity, $|z(t^*)| = R$, and there exists $t_1 \in (t^*, t_{\max})$, maybe equal to $t_+$, such that $|z(t_1)| > R$ and $|z(s)| \le R + \eta$ for all $s \in [t^*, t_1]$. Let $t_0 = \max\{ s \in [t^*, t_1] : |z(s)| \le R\}$, which exists since this set is closed and contains $t^*$. Then $|z(t_0)| = R$ and $R < |z(s)| \le R + \eta$ for all $s \in (t_0, t_1]$. The function $V(t) = |z(t)|^2$ is absolutely continuous and, by \eqref{eq:ball_shell}, $\frac{dV}{ds}(s) = 2 z(s)^\top G(z(s), u(s)) \le -\mu < 0$ for almost all $s \in (t_0, t_1)$. Integrating yields $V(t_1) < V(t_0) = R^2$, which contradicts $|z(t_1)| > R$. Hence $|z(t)| \le R$ for all $t \in [0, t_{\max})$, and therefore $t_{\max} = +\infty$.

\textit{Step 2: general case.} Fix $T \in (0, t_{\max})$. For $\varepsilon > 0$, let $z_\varepsilon$ be the solution, starting from $z(0)$, of $\frac{dz_\varepsilon}{dt} = G(z_\varepsilon, u(t)) - \varepsilon z_\varepsilon$. For $|z| = R$ and $u \in \mathcal{U}$, $z^\top (G(z, u) - \varepsilon z) \le -\varepsilon R^2 < 0$, so that Step 1, applied to $(z, u) \mapsto G(z, u) - \varepsilon z$, shows that $z_\varepsilon$ is defined on $[0, +\infty)$ and satisfies $|z_\varepsilon(t)| \le R$ for all $t \ge 0$. Let $\mathcal{K}$ be a closed ball containing the ball of radius $R$ and the compact set $z([0, T])$, and let $L$ be an upper bound of $|\frac{\partial G}{\partial z}|$ on the compact set $\mathcal{K} \times \mathcal{U}$. The difference $t \mapsto \Delta_\varepsilon(t) = z_\varepsilon(t) - z(t)$ is absolutely continuous, satisfies $\Delta_\varepsilon(0) = 0$ and, by the integral mean value theorem, for almost all $t \in [0, T]$,
\begin{equation*}
 \frac{d\Delta_\varepsilon}{dt}(t) = A_\varepsilon(t) \Delta_\varepsilon(t) - \varepsilon z_\varepsilon(t), \qquad A_\varepsilon(t) = \int_0^1 \frac{\partial G}{\partial z}\left( z(t) + s \Delta_\varepsilon(t), u(t) \right) ds ,
\end{equation*}
where the segment $\{z(t) + s\Delta_\varepsilon(t) : s \in [0,1]\}$ lies in the convex set $\mathcal{K}$ and $u(t) \in \mathcal{U}$, so that $A_\varepsilon$ is measurable, $|A_\varepsilon(t)| \le L$ and $A_\varepsilon(t) + A_\varepsilon(t)^\top \le 2L I_m$. Lemma \ref{lemma-dif-ineg}, with $a_{\!_+} = L$ and $|\varepsilon z_\varepsilon(t)| \le \varepsilon R$, gives
\begin{equation*}
 |z_\varepsilon(t) - z(t)| \le \varepsilon R \int_0^t e^{L(t-s)}\, ds \le \varepsilon R\, T e^{LT}, \qquad \forall t \in [0, T] .
\end{equation*}
Letting $\varepsilon \to 0$ and using that the ball is closed, we obtain $|z(t)| \le R$ for all $t \in [0, T]$. Since $T \in (0, t_{\max})$ is arbitrary, $z$ remains in a compact set on $[0, t_{\max})$, which implies $t_{\max} = +\infty$.
\end{proof}
\stopmodif

\section{Proof of Latent Contraction and Invariance (Proposition \ref{prop:invariance})}
\label{app:proof_invariance}

In this appendix, we prove Proposition \ref{prop:invariance}, which states that the compact set $\Zcal$ is forward-invariant and that any two trajectories driven by the same observation converge exponentially towards each other.

Recall that the latent state $\LatentState$ is governed by the dynamics:
\begin{equation*}
 \frac{d\LatentState}{dt} = \Gain \BaseDynamics_\Params(\LatentState, \obsPhys(t))
\end{equation*}
where $\obsPhys(t)$ takes values in the compact set $\Ycal$. We define the constant $\bar{\BaseDynamics}$ as the maximum magnitude of the unforced base vector field on $\Ycal$:
\begin{equation}
 \bar{\BaseDynamics} = \max_{\Observation \in \Ycal} |\BaseDynamics_\Params(0, \Observation)|.
\end{equation}
The compact operating region $\Zcal$ is defined as the closed ball of radius $R_z = 2\bar{\BaseDynamics}$, i.e., $\Zcal = \{\LatentState \in \LatentSpace : |\LatentState| \le 2\bar{\BaseDynamics}\}$.

\subsection{Forward Invariance of \texorpdfstring{$\Zcal$}{Z}}
To prove that $\Zcal$ is forward-invariant, it is sufficient to show that the vector field points inward on the boundary $\partial \Zcal = \{\LatentState \in \LatentSpace : |\LatentState| = 2\bar{\BaseDynamics}\}$. The outward normal vector at any point on the boundary is collinear with $\LatentState$. Thus, the vector field points inward if:
\begin{equation}
 \LatentState^\top \frac{d\LatentState}{dt} = \Gain \LatentState^\top \BaseDynamics_\Params(\LatentState, \Observation) \le 0 \qquad \forall \LatentState \in \partial\Zcal, \quad \forall \Observation \in \Ycal.
\end{equation}
Using the fundamental theorem of calculus (integral Mean Value Theorem), we can rewrite the base vector field as:
\begin{align*}
 \LatentState^\top \BaseDynamics_\Params(\LatentState, \Observation) &= \LatentState^\top [\BaseDynamics_\Params(\LatentState, \Observation) - \BaseDynamics_\Params(0, \Observation)] + \LatentState^\top \BaseDynamics_\Params(0, \Observation) \\
 &= \LatentState^\top \left[ \int_0^1 \frac{\partial \BaseDynamics_\Params}{\partial \LatentState}(s \LatentState, \Observation) \, ds \right] \LatentState + \LatentState^\top \BaseDynamics_\Params(0, \Observation).
\end{align*}
Since $\Zcal$ is convex and contains the origin, the segment $s\LatentState$ for $s \in [0,1]$ remains entirely in $\Zcal$. For any square matrix $M$, we have $\LatentState^\top M \LatentState = \frac{1}{2}\LatentState^\top (M + M^\top)\LatentState$. Applying this to the Jacobian integral and using the monotonicity condition (Assumption \ref{ass:contraction}), which guarantees that the symmetric part of the Jacobian is bounded by $-2 I_m$, we obtain:
\begin{align*}
 \LatentState^\top \left[ \int_0^1 \frac{\partial \BaseDynamics_\Params}{\partial \LatentState}(s \LatentState, \Observation) \, ds \right] \LatentState &= \frac{1}{2} \LatentState^\top \left[ \int_0^1 \left( \frac{\partial \BaseDynamics_\Params}{\partial \LatentState} + \left(\frac{\partial \BaseDynamics_\Params}{\partial \LatentState}\right)^\top \right)(s \LatentState, \Observation) \, ds \right] \LatentState \\
 &\le \frac{1}{2} \LatentState^\top \left[ \int_0^1 -2 I_m \, ds \right] \LatentState = -|\LatentState|^2.
\end{align*}
Using the Cauchy-Schwarz inequality ($\LatentState^\top \BaseDynamics_\Params(0, \Observation) \le |\LatentState| |\BaseDynamics_\Params(0, \Observation)| \le |\LatentState| \bar{\BaseDynamics}$), the inner product is bounded by:
\begin{equation*}
 \LatentState^\top \BaseDynamics_\Params(\LatentState, \Observation) \le -|\LatentState|^2 + \bar{\BaseDynamics} |\LatentState| = |\LatentState|(\bar{\BaseDynamics} - |\LatentState|).
\end{equation*}
Evaluating the scaled dynamics on the boundary $\partial \Zcal$ where $|\LatentState| = 2\bar{\BaseDynamics}$ and $\Gain > 0$, we get:
\begin{equation*}
 \LatentState^\top \frac{d\LatentState}{dt} \le \Gain \left[ 2\bar{\BaseDynamics} (\bar{\BaseDynamics} - 2\bar{\BaseDynamics}) \right] = -2\Gain\bar{\BaseDynamics}^2 \le 0.
\end{equation*}
\startmodifVA Since the inner product is non-positive on the boundary for all $\Observation \in \Ycal$, Lemma \ref{lemma-invariance-ball}, applied with $G = \Gain \BaseDynamics_\Params$, $\mathcal{U} = \Ycal$ and $u(t) = \obsPhys(t)$, guarantees that any trajectory starting in $\Zcal$ remains in it.\stopmodif{} Therefore, $\Zcal$ is forward-invariant.

\subsection{Exponential Contraction of the Flow}
Consider two initial conditions $\LatentState_{a0}, \LatentState_{b0} \in \Zcal$ and let $\LatentState_a(t)$ and $\LatentState_b(t)$ be the corresponding trajectories driven by the exact same observation signal $\obsPhys(t) \in \Ycal$.

Let $\Delta \LatentState(t) = \LatentState_a(t) - \LatentState_b(t)$. Its time derivative is given by:
\begin{equation*}
 \frac{d}{dt} \Delta \LatentState(t) = \Gain [\BaseDynamics_\Params(\LatentState_a(t), \obsPhys(t)) - \BaseDynamics_\Params(\LatentState_b(t), \obsPhys(t))].
\end{equation*}
By the integral Mean Value Theorem, this can be written as a linear time-varying system:
\begin{equation*}
 \frac{d}{dt} \Delta \LatentState(t) = A(t) \Delta \LatentState(t), \quad \text{where} \quad A(t) = \Gain \int_0^1 \frac{\partial \BaseDynamics_\Params}{\partial \LatentState}(s\LatentState_a(t) + (1-s)\LatentState_b(t), \obsPhys(t)) \, ds.
\end{equation*}
Because $\Zcal$ is forward-invariant and convex, the segment $s\LatentState_a(t) + (1-s)\LatentState_b(t)$ lies entirely within $\Zcal$ for all $s \in [0,1]$ and all $t \ge t_0$. Assumption \ref{ass:contraction} then implies that the symmetric part of the Jacobian inside the integral satisfies:
\begin{equation*}
 A(t) + A(t)^\top = \Gain \int_0^1 \left( \frac{\partial \BaseDynamics_\Params}{\partial \LatentState} + \left(\frac{\partial \BaseDynamics_\Params}{\partial \LatentState}\right)^\top \right) ds \le \Gain \int_0^1 -2 I_m \, ds = -2\Gain I_m.
\end{equation*}
We can now apply Lemma \ref{lemma-dif-ineg} with $\EuScript{M} = \Delta \LatentState$, $a_+ = -\Gain$, and $B(t) = 0$. This yields directly:
\begin{equation*}
 |\Delta \LatentState(t)| \le e^{-\Gain(t-t_0)} |\Delta \LatentState(t_0)| \qquad \forall t \ge t_0.
\end{equation*}
This concludes the proof of Proposition \ref{prop:invariance}.

\subsection{Corollary: Invariance under Additive Perturbations}
\label{app:remark_invariance_perturbed}
For the subsequent proofs (notably the construction of asymptotic approximations), we highlight a crucial corollary of the invariance analysis. Consider the augmented dynamics $\frac{d\LatentState}{dt} = \Gain [ \BaseDynamics_\Params(\LatentState, \Observation) - v ]$, where $v \in \LatentSpace$ is a constant additive perturbation. The boundary condition evaluation becomes:
\begin{align*}
 \LatentState^\top [ \BaseDynamics_\Params(\LatentState, \Observation) - v ] &\le -|\LatentState|^2 + |\LatentState| \left( |\BaseDynamics_\Params(0, \Observation)| + |v| \right) \\
 &\le |\LatentState| (\bar{\BaseDynamics} + |v| - |\LatentState|).
\end{align*}
If the perturbation satisfies $|v| \le \bar{\BaseDynamics}$, then on the boundary $|\LatentState| = 2\bar{\BaseDynamics}$, we obtain:
\begin{equation*}
 \LatentState^\top [ \BaseDynamics_\Params(\LatentState, \Observation) - v ] \le 2\bar{\BaseDynamics} (\bar{\BaseDynamics} + \bar{\BaseDynamics} - 2\bar{\BaseDynamics}) = 0.
\end{equation*}
Multiplying by $\Gain > 0$ preserves the non-positive sign. \startmodifVA Hence, for any fixed $\Observation \in \Ycal$ and $v \in \LatentSpace$ with $|v| \le \bar{\BaseDynamics}$, Lemma \ref{lemma-invariance-ball}, applied with $G(\LatentState, u) = \Gain [ \BaseDynamics_\Params(\LatentState, \Observation) - v ]$ and the constant input $u = (\Observation, v)$, shows that\stopmodif{} the compact set $\Zcal$ remains forward-invariant even when the vector field is shifted by any constant vector $v$ such that $|v| \le \bar{\BaseDynamics}$.

\section{Existence and Regularity of the State-to-Latent Map (Theorem \ref{thm:embedding})}
\label{app:proof_embedding}

In this appendix, we prove Theorem \ref{thm:embedding}. The proof is structured in four parts: defining the system bounds, establishing the bounds on the physical flow and observations, proving the existence of a unique bounded solution, and finally establishing its invariance, attractivity, and regularity.

It can be observed that, from Assumption \ref{ass:regularity}, the physical solutions initialized in the domain of interest coincide in positive time with that of the modified dynamics
\begin{equation}
\label{eq:syst_modified}
    \dot{\PhysicalState} = \Breve{\GenerativeDrift}(\PhysicalState) = \chi(\PhysicalState)\GenerativeDrift(\PhysicalState),
\end{equation}
with $\chi$ a smooth function such that
\begin{equation}\label{eq_ModifPhysicalSystem}
    \chi(\PhysicalState) = 1 \hspace{0.2cm}\text{if}\hspace{0.2cm} \PhysicalState \in \X, \hspace{0.5cm}\chi(\PhysicalState) = 0 \hspace{0.2cm}\text{if}\hspace{0.2cm} \PhysicalState \notin \Xa \nonumber
\end{equation}
where $\Xa$ is the compact inflation of $\X$ defined below Assumption \ref{ass:regularity}. Besides, Assumptions \ref{ass:regularity} (and \ref{ass:observability}) still hold for the modified dynamics \eqref{eq:syst_modified} with the output map $\GenerativeObs$. In the rest of the proof, we thus assume without loss of generality that $\GenerativeDrift = \Breve{\GenerativeDrift}$ and we consider that the modified dynamics ensures global (forward and backward) invariance of the extended compact set $\Xa$, as well as completeness and boundedness of all solutions over $\RR$.

To analyze the system in the natural time scale of the latent dynamics, we work in the observer time $\tau = \Gain t$. The physical flow initialized at $\PhysicalState_0$ is thus considered at time $t = \tau/\Gain$, and the corresponding observation fed to the network is:
\begin{equation}
 \obsLatent_{\PhysicalState_0}(\tau) = \GenerativeObs\left(\flowf\left(\PhysicalState_0, \frac{\tau}{\Gain}\right)\right).
\end{equation}
When the explicit dependence on the initial condition is not required for the analysis, we simply denote the observation signal in the latent time scale as $\obsLatent(\tau) = \obsPhys(\tau/\Gain)$. The latent dynamics in this time scale are $\frac{d\LatentState}{d\tau} = \BaseDynamics_\Params(\LatentState, \obsLatent_{\PhysicalState_0}(\tau))$. Let $\Phi_\LatentState(\LatentState_0, \tau_0, \tau; \PhysicalState_0)$ denote the flow of this latent system with $\LatentState_0$ being the initial condition at time $\tau_0$.

\subsection{Preamble: System and Filter Bounds}
Because of Assumption \ref{ass:regularity}, the physical trajectories remain in a compact set $\Xa$ and the base dynamics are evaluated on the compact set $\Zcal \times \Ycal$. Consequently, the following finite bounds are well-defined for the physical system:
\begin{align}
\label{bound-for-fh}
 L_\GenerativeObs &= \sup_{(\PhysicalState,v)\in \Xa \times \PhysicalSpace} \frac{1}{|v|}\left|\frac{\partial \GenerativeObs}{\partial \PhysicalState}(\PhysicalState)v\right|, \quad 
 &L_{\GenerativeObs x} &= \sup_{(\PhysicalState_a,\PhysicalState_b,v)\in \Xa^2 \times \PhysicalSpace} \frac{\left|\left(\frac{\partial \GenerativeObs}{\partial \PhysicalState}(\PhysicalState_a)-\frac{\partial \GenerativeObs}{\partial \PhysicalState}(\PhysicalState_b)\right)v\right|}{|\PhysicalState_a-\PhysicalState_b| |v|}, \\
 L_\GenerativeDrift &= \sup_{(\PhysicalState,v)\in \Xa \times \PhysicalSpace} \frac{1}{|v|}\left|\frac{\partial \GenerativeDrift}{\partial \PhysicalState}(\PhysicalState)v\right|, \quad 
 &L_{\GenerativeDrift x} &= \sup_{(\PhysicalState_a,\PhysicalState_b,v)\in \Xa^2 \times \PhysicalSpace} \frac{\left|\left(\frac{\partial \GenerativeDrift}{\partial \PhysicalState}(\PhysicalState_a)-\frac{\partial \GenerativeDrift}{\partial \PhysicalState}(\PhysicalState_b)\right)v\right|}{|\PhysicalState_a-\PhysicalState_b| |v|}.
\end{align}
Similarly, we define the bounds for the latent base dynamics $\BaseDynamics_\Params$ on $\Zcal \times \Ycal$:
\begin{align}
\label{bound-for-g}
 L_{go} &= \sup_{(\LatentState,\Observation)\in \Zcal\times\Ycal} \left| \frac{\partial \BaseDynamics_\Params}{\partial \Observation}(\LatentState,\Observation) \right|, \quad 
 &L_{goo} &= \sup_{(\LatentState,\Observation_a,\Observation_b)\in \Zcal\times\Ycal^2} \frac{\left| \frac{\partial \BaseDynamics_\Params}{\partial \Observation}(\LatentState,\Observation_a)-\frac{\partial \BaseDynamics_\Params}{\partial \Observation}(\LatentState,\Observation_b)\right|}{|\Observation_a-\Observation_b|}, \\
 L_{gz} &= \sup_{(\LatentState,\Observation)\in \Zcal\times\Ycal} \left| \frac{\partial \BaseDynamics_\Params}{\partial \LatentState}(\LatentState,\Observation) \right|, \quad 
 &L_{gzz} &= \sup_{(\LatentState_a,\LatentState_b, \Observation)\in \Zcal^2\times\Ycal} \frac{\left| \frac{\partial \BaseDynamics_\Params}{\partial \LatentState}(\LatentState_a,\Observation)-\frac{\partial \BaseDynamics_\Params}{\partial \LatentState}(\LatentState_b,\Observation)\right|}{|\LatentState_a-\LatentState_b|},
\end{align}
along with the cross-derivative bounds $L_{goz}$ and $L_{gzo}$ defined symmetrically.

\subsection{Bounds on the Physical Flow and Observations}
We establish the exponential growth bounds for the variations of the observation $\obsLatent_{\PhysicalState_0}(\tau)$ with respect to the initial condition $\PhysicalState_0$. The variational equation for the physical flow is:
\begin{equation*}
 \frac{\partial^2 \flowf}{\partial \PhysicalState \partial t}(\PhysicalState_0, t) = \frac{\partial \GenerativeDrift}{\partial \PhysicalState}(\flowf(\PhysicalState_0, t)) \frac{\partial \flowf}{\partial \PhysicalState}(\PhysicalState_0, t), \qquad \frac{\partial \flowf}{\partial \PhysicalState}(\PhysicalState_0, 0) = I_n.
\end{equation*}
With \eqref{bound-for-fh}, the symmetric part of the Jacobian is bounded by $2L_\GenerativeDrift I_n$. Invoking Lemma \ref{lemma-dif-ineg} yields:
\begin{equation}
 \left|\frac{\partial \flowf}{\partial \PhysicalState}(\PhysicalState_0, t)\right| \le e^{L_\GenerativeDrift |t|} \qquad \forall (t, \PhysicalState_0) \in \RR \times \Xa.
\end{equation}
To bound the second-order variations, let $\PhysicalState_a, \PhysicalState_b \in \Xa$. We have:
\begin{align*}
 \frac{\partial}{\partial t} \left[ \frac{\partial \flowf}{\partial \PhysicalState}(\PhysicalState_a, t) - \frac{\partial \flowf}{\partial \PhysicalState}(\PhysicalState_b, t) \right] = 
 &\left[ \frac{\partial \GenerativeDrift}{\partial \PhysicalState}(\flowf(\PhysicalState_a, t)) - \frac{\partial \GenerativeDrift}{\partial \PhysicalState}(\flowf(\PhysicalState_b, t)) \right] \frac{\partial \flowf}{\partial \PhysicalState}(\PhysicalState_a, t) \\
 &+ \frac{\partial \GenerativeDrift}{\partial \PhysicalState}(\flowf(\PhysicalState_b, t)) \left[ \frac{\partial \flowf}{\partial \PhysicalState}(\PhysicalState_a, t) - \frac{\partial \flowf}{\partial \PhysicalState}(\PhysicalState_b, t) \right].
\end{align*}
The norm of the first term is bounded by $L_{\GenerativeDrift x} |\flowf(\PhysicalState_a, t) - \flowf(\PhysicalState_b, t)| e^{L_\GenerativeDrift |t|}$. Using the Mean Value Theorem, $|\flowf(\PhysicalState_a, t) - \flowf(\PhysicalState_b, t)| \le e^{L_\GenerativeDrift |t|} |\PhysicalState_a - \PhysicalState_b|$. Thus, the first term is bounded by $L_{\GenerativeDrift x} e^{2L_\GenerativeDrift |t|} |\PhysicalState_a - \PhysicalState_b|$. Applying Lemma \ref{lemma-dif-ineg} again (with initial difference 0), we obtain:
\begin{equation}
 \left|\frac{\partial \flowf}{\partial \PhysicalState}(\PhysicalState_a, t) - \frac{\partial \flowf}{\partial \PhysicalState}(\PhysicalState_b, t)\right| \le \int_0^{|t|} e^{L_\GenerativeDrift(|t|-|s|)} L_{\GenerativeDrift x} e^{2L_\GenerativeDrift |s|} \, ds \, |\PhysicalState_a - \PhysicalState_b| \le \frac{L_{\GenerativeDrift x}}{L_\GenerativeDrift} e^{2L_\GenerativeDrift |t|} |\PhysicalState_a - \PhysicalState_b|.
\end{equation}
By applying the chain rule to $\obsLatent_{\PhysicalState_0}(\tau) = \GenerativeObs(\flowf(\PhysicalState_0, \tau/\Gain))$, we obtain the bounds on the observation derivatives. Let $\lambda = L_\GenerativeDrift / \Gain$. We get:
\begin{align}
 \left|\frac{\partial \obsLatent_{\PhysicalState_0}}{\partial \PhysicalState_0}(\tau)\right| &\le L_\GenerativeObs e^{\lambda |\tau|} \label{eq:exp_growth_u} \\
 \left|\frac{\partial \obsLatent_{\PhysicalState_a}}{\partial \PhysicalState_0}(\tau) - \frac{\partial \obsLatent_{\PhysicalState_b}}{\partial \PhysicalState_0}(\tau)\right| &\le L_{\Observation x}e^{2\lambda |\tau|} |\PhysicalState_a - \PhysicalState_b| \label{eq:exp_growth_del_u}
\end{align}
where
$L_{\Observation x}= \left(L_{\GenerativeObs x} + L_\GenerativeObs \frac{L_{\GenerativeDrift x}}{L_\GenerativeDrift} \right) $.
If $\Gain > 2L_\GenerativeDrift$, it implies  $\lambda < \frac{1}{2}$.

\subsection{Well-Definedness of the State-to-Latent Map}
\label{LPNsec1}

Let $\mathcal{F} = C^0_b(\RR, \Zcal)$ be the complete metric space of continuous bounded functions $\omega: \RR \to \Zcal$, equipped with the uniform norm $\|\omega\|_\infty = \sup_{\tau} |\omega(\tau)|$. 
Fix an arbitrary time shift $\theta > 0$. We define the operator $\mathbb{O}_\theta: \mathcal{F} \times \Xa \to \mathcal{F}$ by:
\begin{equation}
 \mathbb{O}_\theta(\omega, \PhysicalState_0)(\tau) = \Phi_\LatentState(\omega(\tau-\theta), \tau-\theta, \tau; \PhysicalState_0).
\end{equation}
By the forward invariance of $\Zcal$ (Proposition \ref{prop:invariance}), $\mathbb{O}_\theta$ maps $\mathcal{F}$ into itself. Furthermore, the contraction property of the flow guarantees that for any $\omega_a, \omega_b \in \mathcal{F}$:
\begin{align*}
 |\mathbb{O}_\theta(\omega_a, \PhysicalState_0)(\tau) - \mathbb{O}_\theta(\omega_b, \PhysicalState_0)(\tau)| &\le e^{-\theta} |\omega_a(\tau-\theta) - \omega_b(\tau-\theta)| \le e^{-\theta} \|\omega_a - \omega_b\|_\infty.
\end{align*}
Since $\theta > 0$, $e^{-\theta} < 1$, making $\mathbb{O}_\theta$ a strict contraction on $\mathcal{F}$ which is uniform in $x_0$ in $\Xa$. By the Banach Fixed-Point Theorem, there exists a unique fixed point $\varpi_{\PhysicalState_0} \in \mathcal{F}$.

A priori, the abstract fixed point $\varpi_{\theta, \PhysicalState_0} \in C^0_b(\RR, \Zcal)$ provided by Banach's theorem depends on the chosen time shift $\theta$, and it is not trivially a continuous-time solution of the differential equation everywhere. We now demonstrate that it is indeed the unique bounded exact solution.

Note that by iteration we get for any integer $m \in \NN$:
\begin{align*}
\varpi_{\theta, \PhysicalState_0}(\tau) &= \Phi_\LatentState(\varpi_{\theta, \PhysicalState_0}(\tau-\theta), \tau-\theta, \tau; \PhysicalState_0) \\
&= \Phi_\LatentState(\Phi_\LatentState(\varpi_{\theta, \PhysicalState_0}(\tau-2\theta), \tau-2\theta, \tau-\theta; \PhysicalState_0), \tau-\theta, \tau; \PhysicalState_0) \\
&= \Phi_\LatentState(\varpi_{\theta, \PhysicalState_0}(\tau-m\theta), \tau-m\theta, \tau; \PhysicalState_0).
\end{align*}
Namely, $\varpi_{\theta, \PhysicalState_0}$ is also the fixed point of $\mathbb{O}_{m\theta}$ for all integers $m$.

Now, consider the fractioned sequence $\theta_k = \frac{\theta}{k}$ for $k \in \NN^*$. By the exact same Banach theorem, there exists a unique fixed point $\varpi_{\theta_k, \PhysicalState_0} \in C^0_b(\RR, \Zcal)$ satisfying:
$$ \varpi_{\theta_k, \PhysicalState_0}(\tau) = \Phi_\LatentState(\varpi_{\theta_k, \PhysicalState_0}(\tau - \textstyle\frac{m}{k}\theta), \tau - \textstyle\frac{m}{k}\theta, \tau; \PhysicalState_0) \qquad \forall \tau \in \RR, \ \forall m \in \NN. $$
In particular, for $m=k$, we obtain that $\varpi_{\theta_k, \PhysicalState_0}$ is also a fixed point of $\mathbb{O}_\theta$. Uniqueness of the fixed point implies that $\varpi_{\theta_k, \PhysicalState_0} = \varpi_{\theta, \PhysicalState_0}$ for all $k \in \NN^*$, meaning:
\begin{equation}
\label{eq:dense_flow}
\varpi_{\theta, \PhysicalState_0}(\tau) = \Phi_\LatentState(\varpi_{\theta, \PhysicalState_0}(\tau - \textstyle\frac{m}{k}\theta), \tau - \textstyle\frac{m}{k}\theta, \tau; \PhysicalState_0) \qquad \forall \tau, \ \forall m \in \NN, \ \forall k \in \NN^*.
\end{equation}

To show that $\varpi_{\theta, \PhysicalState_0}$ is a true differential solution, let $\tau_0 < \tau$ be arbitrary and denote $m(k)$ the integer satisfying:
$$ \textstyle\frac{m(k)}{k}\theta \le \tau - \tau_0 < \textstyle\frac{m(k)+1}{k}\theta. $$
We have $\lim_{k\to\infty} \tau - \textstyle\frac{m(k)}{k}\theta = \tau_0$. Because the functions $\tau \mapsto \varpi_{\theta, \PhysicalState_0}(\tau)$ and $(\tau_0, \LatentState_0) \mapsto \Phi_\LatentState(\LatentState_0, \tau_0, \tau; \PhysicalState_0)$ are continuous, passing to the limit $k \to \infty$ in \eqref{eq:dense_flow} yields:
$$ \varpi_{\theta, \PhysicalState_0}(\tau) = \Phi_\LatentState(\varpi_{\theta, \PhysicalState_0}(\tau_0), \tau_0, \tau; \PhysicalState_0). $$

For $\tau \le \tau_0$, with
$$ \textstyle\frac{m(k)}{k}\theta \le \tau_0 - \tau < \textstyle\frac{m(k)+1}{k}\theta, $$
we have
$$ \lim_{k\to\infty} \tau + \textstyle\frac{m(k)}{k}\theta = \tau_0. $$
Also, since we have uniqueness of solutions, we can invert the flow in \eqref{eq:dense_flow} to get:
$$ \Phi_\LatentState(\varpi_{\theta, \PhysicalState_0}(s), s, s - \textstyle\frac{m(k)}{k}\theta; \PhysicalState_0) = \varpi_{\theta, \PhysicalState_0}(s - \textstyle\frac{m(k)}{k}\theta) \qquad \forall s \in \RR, \ \forall k \in \NN^*, $$
and consequently, by setting $s = \tau + \frac{m(k)}{k}\theta$:
$$ \Phi_\LatentState(\varpi_{\theta, \PhysicalState_0}(\tau + \textstyle\frac{m(k)}{k}\theta), \tau + \textstyle\frac{m(k)}{k}\theta, \tau; \PhysicalState_0) = \varpi_{\theta, \PhysicalState_0}(\tau). $$
So as above, by invoking continuity, the left hand side tends to $\Phi_\LatentState(\varpi_{\theta, \PhysicalState_0}(\tau_0), \tau_0, \tau; \PhysicalState_0)$ when $k \to \infty$, and we can conclude that:
$$ \varpi_{\theta, \PhysicalState_0}(\tau) = \Phi_\LatentState(\varpi_{\theta, \PhysicalState_0}(\tau_0), \tau_0, \tau; \PhysicalState_0). $$

Consequently, $\varpi_{\theta, \PhysicalState_0}$ is an exact continuous-time trajectory of the latent dynamics everywhere. Because it is globally bounded, and we previously showed that at most one bounded solution can exist, it is strictly independent of the choice of $\theta > 0$. We denote it simply as $\varpi_{\PhysicalState_0}$.

 We can thus  define the state-to-latent map $\Encoder_\Gain: \Xa \to \Zcal$ as:
 \begin{equation}
  \label{eq_DefEncoder}\Encoder_\Gain(\PhysicalState_0) = \varpi_{\PhysicalState_0}(0).
 \end{equation}

\subsection{Invariance and Attractivity}
By the semi-group property of the physical system, $\flowf(\PhysicalState_0, \frac{\tau+s}{\Gain}) = \flowf(\flowf(\PhysicalState_0, \frac{s}{\Gain}), \frac{\tau}{\Gain})$, which implies that the input observation satisfies $\Observation_{\flowf(\PhysicalState_0, s/\Gain)}(\tau) = \obsLatent_{\PhysicalState_0}(\tau+s)$. 
Consequently, $\tau \mapsto \varpi_{\PhysicalState_0}(\tau+s)$ and $\tau \mapsto \varpi_{\flowf(\PhysicalState_0, s/\Gain)}(\tau)$ are both bounded solutions on $\RR$ driven by the exact same observation sequence. By uniqueness of the bounded solution, they must be equal:
\begin{equation*}
 \varpi_{\PhysicalState_0}(\tau+s) = \varpi_{\flowf(\PhysicalState_0, s/\Gain)}(\tau) \qquad \forall (\tau, s).
\end{equation*}
Evaluating at $\tau=0$ and setting $s = \Gain t$, we get $\Encoder_\Gain(\flowf(\PhysicalState_0, t)) = \varpi_{\PhysicalState_0}(\Gain t)$. Since $\varpi_{\PhysicalState_0}$ is a valid flow trajectory passing through $\varpi_{\PhysicalState_0}(0) = \Encoder_\Gain(\PhysicalState_0)$, this proves the invariance property.

The attractivity follows directly from the contraction property of Proposition \ref{prop:invariance}:
\begin{equation*}
 |\Phi_\LatentState(\LatentState_0, 0, \Gain t; \PhysicalState_0) - \Encoder_\Gain(\flowf(\PhysicalState_0, t))| = |\Phi_\LatentState(\LatentState_0, 0, \Gain t; \PhysicalState_0) - \varpi_{\PhysicalState_0}(\Gain t)| \le e^{-\Gain t}|\LatentState_0 - \Encoder_\Gain(\PhysicalState_0)|.
\end{equation*}

\subsection{Regularity via Ascoli-Arzelà Theorem}

In this section, we restrict our attention to negative times $\tau \leq 0$.
We prove that the unique bounded map $\varpi$ is $C^1$ with a Lipschitz derivative on the compact set $\Xa$. The solution $\varpi_{\PhysicalState_0}$ being the limit of a sequence of bounded functions, it is sufficient to show that the sequence made of their partial derivatives is uniformly bounded and equicontinuous, allowing us to invoke the Ascoli-Arzelà theorem.

\vspace{1em}\noindent
\textbf{Step 1: The Iterative Sequence and its Formal Derivatives}\\
Let $\theta > 0$ be a fixed time shift. We define the iterative sequence of functions $\omega_k: \RR \times \Xa \to \Zcal$ by:
\begin{equation}
    \omega_0(\tau, \PhysicalState_0) \equiv 0, \qquad
    \omega_{k+1}(\tau, \PhysicalState_0) = \mathbb{O}_\theta(\omega_k, \PhysicalState_0)(\tau)
    =  \Phi_\LatentState(\omega_k(\tau-\theta, \PhysicalState_0), \tau-\theta, \tau; \PhysicalState_0).
\end{equation}
We have shown in Section \ref{LPNsec1}  that $\omega_k$ converges to $\varpi$ uniformly in
$(\tau, \PhysicalState_0)$ in $]-\infty,0]\times \Xa$.

Since the vector fields $\BaseDynamics_\Params$ and $\GenerativeDrift$, as well as the map $\GenerativeObs$, are $C^2$ by Assumption \ref{ass:regularity}, the flow $\Phi_\LatentState(\LatentState_0, \tau_0, \tau; \PhysicalState_0)$ is $C^1$ with respect to both $\LatentState_0$ and $\PhysicalState_0$. By induction, each $\omega_k$ is $C^1$ with respect to $\PhysicalState_0$. Differentiating the recurrence yields:
\begin{equation}
    \frac{\partial \omega_{k+1}}{\partial \PhysicalState_0}(\tau, \PhysicalState_0) = J_z(\tau) \frac{\partial \omega_k}{\partial \PhysicalState_0}(\tau-\theta, \PhysicalState_0) + J_x(\tau),
    \label{eq:deriv_recursion}
\end{equation}
where $J_z(\tau) = \frac{\partial \Phi_\LatentState}{\partial \LatentState_0}$ and $J_x(\tau) = \frac{\partial \Phi_\LatentState}{\partial \PhysicalState_0}$ are the Jacobians of the flow evaluated along the trajectory initialized at $\LatentState_0 = \omega_k(\tau-\theta, \PhysicalState_0)$ at time $\tau_0 = \tau-\theta$.

\vspace{1em}\noindent
\textbf{Step 2: First-Order Variational Equations and Bounds}\\
To bound the sequence \eqref{eq:deriv_recursion}, we first bound the Jacobians. They satisfy the variational equations:
\begin{align}
    \frac{d}{d\tau} J_z &= \frac{\partial \BaseDynamics_\Params}{\partial \LatentState}(\Phi_\LatentState, \obsLatent_{\PhysicalState_0}) J_z, \qquad J_z(\tau_0) = I_m, \label{eq:var_jz} \\
    \frac{d}{d\tau} J_x &= \frac{\partial \BaseDynamics_\Params}{\partial \LatentState}(\Phi_\LatentState, \obsLatent_{\PhysicalState_0}) J_x + \frac{\partial \BaseDynamics_\Params}{\partial \Observation}(\Phi_\LatentState, \obsLatent_{\PhysicalState_0}) \frac{\partial \obsLatent_{\PhysicalState_0}}{\partial \PhysicalState_0}(\tau), \qquad J_x(\tau_0) = 0. \label{eq:var_jx}
\end{align}
By Assumption \ref{ass:contraction}, the symmetric part of $\frac{\partial \BaseDynamics_\Params}{\partial \LatentState}$ is bounded by $-2 I_m$. Applying Lemma \ref{lemma-dif-ineg} directly to \eqref{eq:var_jz} gives:
\begin{equation}
    |J_z(\tau)| \le e^{-(\tau - \tau_0)}\qquad \forall \tau \geq \tau _0 .
    \label{eq:bound_dphi_dz}
\end{equation}
For $J_x$, we use the bound on the observation variations $|\frac{\partial \obsLatent_{\PhysicalState_0}}{\partial \PhysicalState_0}(\tau)| \le L_\GenerativeObs e^{\lambda|\tau|}$ from \eqref{eq:exp_growth_u}, where $\lambda = L_\GenerativeDrift / \Gain$. Applying Lemma \ref{lemma-dif-ineg} to \eqref{eq:var_jx} yields:
\begin{equation}
    |J_x(\tau)| \le \int_{\tau_0}^{\tau} e^{-(\tau - s)} L_{go} L_\GenerativeObs e^{-\lambda s} ds \le \frac{L_{go} L_\GenerativeObs}{1-\lambda} e^{\lambda|\tau|}
    \qquad \forall \tau _0\leq \tau \leq 0 .
    \label{eq:bound_dphi_dx}
\end{equation}

\vspace{1em}\noindent
\textbf{Step 3: Uniform Boundedness of the Sequence of Derivatives}\\
Substituting the bounds \eqref{eq:bound_dphi_dz} and \eqref{eq:bound_dphi_dx} into the norm of the recursion \eqref{eq:deriv_recursion} (with $\tau_0 = \tau - \theta$) and multiplying by 
$e^{-\lambda|\tau|}$
gives:
\begin{equation*}
    \left| \frac{\partial \omega_{k+1}}{\partial \PhysicalState_0}(\tau, \PhysicalState_0) \right| e^{-\lambda|\tau|}\le e^{-\theta} \left| \frac{\partial \omega_k}{\partial \PhysicalState_0}(\tau-\theta, \PhysicalState_0) \right|e^{-\lambda|\tau|} + \frac{L_{go} L_\GenerativeObs}{1-\lambda}.
\end{equation*}
and therefore, with $|\tau|\geq |\tau-\theta|-\theta$.
\begin{equation*}
    \sup_{\tau \le 0, \PhysicalState_0 \in \Xa} \left| \frac{\partial \omega_k}{\partial \PhysicalState_0}(\tau, \PhysicalState_0) \right| e^{-\lambda|\tau|}
    \le
    e^{-(1-\lambda)\theta} 
     \sup_{\tau \le 0, \PhysicalState_0 \in \Xa}
     \left| \frac{\partial \omega_k}{\partial \PhysicalState_0}(\tau-\theta, \PhysicalState_0) \right|e^{-\lambda|\tau-\theta|}
     +\frac{L_{go} L_\GenerativeObs}{1-\lambda} .
\end{equation*}
Because $\omega_0 \equiv 0$, we get by induction that the sequence is uniformly bounded for all $k \ge 0$:
\begin{equation}
    \sup_{\tau \le 0, \PhysicalState_0 \in \Xa} \left| \frac{\partial \omega_k}{\partial \PhysicalState_0}(\tau, \PhysicalState_0) \right| e^{-\lambda|\tau|} \le \frac{L_{go} L_\GenerativeObs}{(1-\lambda)(1 - e^{-(1-\lambda)\theta})} := C_{\omega'}
    \label{eq:uniform_bound_deriv}
\end{equation}
where $\lambda = L_\GenerativeDrift / \Gain < \frac{1  }{2}$.
This establishes the sequence $\left\{\frac{\partial \omega_k}{\partial \PhysicalState_0}\right\}$ is uniformly bounded 
in $[-q,0] \times \Xa$, for any $q>0$.

\vspace{1em}\noindent
\textbf{Step 4: Second-Order Variations and Lipschitz Bounds}\\
To apply Ascoli-Arzelà, we need to guarantee the equicontinuity of the derivatives, which can be done by bounding the second-order variations. Consider two parameters $\PhysicalState_{a0}, \PhysicalState_{b0} \in \Xa$ and define $\Delta J_z(\tau) = J_{za}(\tau) - J_{zb}(\tau)$. Subtracting the variational equations \eqref{eq:var_jz} evaluated along the two trajectories yields:
\begin{equation*}
\frac{d}{d\tau} \Delta J_z = \frac{\partial \BaseDynamics_\Params}{\partial \LatentState}\bigg|_a \Delta J_z + \underbrace{\left( \frac{\partial \BaseDynamics_\Params}{\partial \LatentState}\bigg|_a - \frac{\partial \BaseDynamics_\Params}{\partial \LatentState}\bigg|_b \right) J_{zb}}_{:= F_z(\tau)}.
\end{equation*}
We have $|F_z(\tau)| \le \big(L_{gzz}|\Delta \LatentState(\tau)| + L_{gzo}|\Delta \obsLatent(\tau)|\big)|J_{zb}(\tau)|$. And with \eqref{eq:bound_dphi_dz} and $|\Delta \obsLatent(\tau)| \le L_\GenerativeObs e^{\lambda|\tau|}|\Delta \PhysicalState_0|$, alongside the explicit latent flow variation bound $|\Delta \LatentState(\tau)| \le e^{-(\tau-\tau_0)}|\Delta \LatentState_0| + C_{\Phi x} e^{\lambda|\tau|} |\Delta \PhysicalState_0|$, we obtain:
\begin{align*}
|F_z(\tau)| &\le \left( L_{gzz} \left[ e^{-(\tau-\tau_0)}|\Delta \LatentState_0| + C_{\Phi x} e^{\lambda|\tau|} |\Delta \PhysicalState_0| \right] + L_{gzo} L_\GenerativeObs e^{\lambda|\tau|} |\Delta \PhysicalState_0| \right) e^{-(\tau-\tau_0)} \\
&= L_{gzz} e^{-2(\tau-\tau_0)} |\Delta \LatentState_0| + \lambda \wp_{zx} e^{\lambda|\tau|} e^{-(\tau-\tau_0)} |\Delta \PhysicalState_0|,
\end{align*}
where we define the structural constant $\wp_{zx} = \frac{1}{\lambda} (L_{gzz} C_{\Phi x} + L_{gzo} L_\GenerativeObs)$. Applying Lemma \ref{lemma-dif-ineg} to the differential equation governing $\Delta J_z$ with the initial condition $\Delta J_z(\tau_0) = 0$ requires evaluating the convolution integral $\int_{\tau_0}^{\tau} e^{-(\tau-s)} |F_z(s)| \, ds$. Because the integration domain is strictly non-positive ($\tau_0 \le s \le \tau \le 0$), we substitute $|s| = -s$. The exact integrations therefore yield:
\begin{align*}
\int_{\tau_0}^\tau e^{-(\tau-s)} e^{-2(s-\tau_0)} ds &\le e^{-(\tau-\tau_0)}, \\
\int_{\tau_0}^\tau e^{-(\tau-s)} e^{-\lambda s} e^{-(s-\tau_0)} ds &= \frac{1}{\lambda} e^{-(\tau-\tau_0)} \big(e^{\lambda|\tau_0|} - e^{\lambda|\tau|}\big).
\end{align*}
This directly provides the structural Lipschitz bound for the state Jacobian:
\begin{equation}
\left| \Delta J_z(\tau) \right| \le L_{gzz} e^{-(\tau-\tau_0)} |\LatentState_{a0} - \LatentState_{b0}| + \wp_{zx} e^{-(\tau-\tau_0)} \big(e^{\lambda|\tau_0|} - e^{\lambda|\tau|}\big) |\PhysicalState_{a0} - \PhysicalState_{b0}|.
\label{eq:lip_dphi_dz}
\end{equation}

A similar procedure is applied to the parameter Jacobian difference $\Delta J_x(\tau)$. Subtracting the equations \eqref{eq:var_jx} reveals a forcing term $F_x(\tau)$ containing four distinct differences. Bounding them using the known structural constants gives:
\begin{align*}
|F_x(\tau)| &\le \left( L_{gzz}|\Delta \LatentState(\tau)| + L_{gzo}|\Delta \obsLatent(\tau)| \right) |J_{xb}(\tau)| + L_{go} \left|\Delta \frac{\partial \obsLatent}{\partial \PhysicalState_0}(\tau)\right|
\\
&\quad + \left( L_{gzo}|\Delta \LatentState(\tau)| + L_{goo}|\Delta \obsLatent(\tau)| \right) \left|\frac{\partial \obsLatent}{\partial \PhysicalState_0}\right|_b \\
&\le \lambda \wp_{zx} e^{\lambda|\tau|} e^{-(\tau-\tau_0)} |\Delta \LatentState_0| + \wp_{xx} e^{2\lambda|\tau|} |\Delta \PhysicalState_0|,
\end{align*}
where $\wp_{xx} = L_{gzz} C_{\Phi x}^2 + 2 L_{gzo} C_{\Phi x} L_\GenerativeObs + L_{go} L_{\Observation x} + L_{goo} L_\GenerativeObs^2$. Integrating this via Lemma \ref{lemma-dif-ineg} requires bounding the cross-term integral:
\begin{equation*}
\int_{\tau_0}^\tau e^{-(\tau-s)} e^{-\lambda s} e^{-(s-\tau_0)} ds = \frac{1}{\lambda} e^{-(\tau-\tau_0)} e^{\lambda|\tau_0|} - \frac{1}{\lambda} e^{-(\tau-\tau_0)} e^{\lambda|\tau|} \le \frac{1}{\lambda} e^{-(1-\lambda)(\tau-\tau_0)} e^{\lambda|\tau|},
\end{equation*}
where we used the identity $\lambda|\tau_0| = \lambda|\tau| + \lambda(\tau-\tau_0)$ derived from the fact that $s \le \tau \le 0$. The second term integrates as $\int_{\tau_0}^\tau e^{-(\tau-s)} e^{-2\lambda s} ds \le \frac{1}{1-2\lambda} e^{2\lambda|\tau|}$. Combining these bounds strictly yields:
\begin{equation}
\left| \Delta J_x(\tau) \right| \le \wp_{zx} e^{-(1-\lambda)(\tau-\tau_0)} e^{\lambda|\tau|} |\LatentState_{a0} - \LatentState_{b0}| + \frac{\wp_{xx}}{1-2\lambda} e^{2\lambda|\tau|} |\PhysicalState_{a0} - \PhysicalState_{b0}|.
\label{eq:lip_dphi_dx}
\end{equation}
Note that the integration bounding the parameter sensitivity requires the exponent $1 - 2\lambda$ to be strictly positive, which natively enforces the required high-gain condition $\Gain > 2L_\GenerativeDrift$.

\vspace{1em}\noindent
\textbf{Step 5: Equicontinuity}\\
By injecting the Lipschitz bounds \eqref{eq:lip_dphi_dz} and \eqref{eq:lip_dphi_dx} back into the difference of the recursive equation \eqref{eq:deriv_recursion}, multiplying by
$e^{-2\lambda|\tau|}$ and utilizing the uniform bound $C_{\omega'}$ obtained in \eqref{eq:uniform_bound_deriv}, algebraic recombination leads to:
\begin{equation*}
    \left| \Delta \frac{\partial \omega_{k+1}}{\partial \PhysicalState_0} (\tau )\right|e^{-2\lambda|\tau|}
    \le
    e^{-\theta} \left| \Delta \frac{\partial \omega_k}{\partial \PhysicalState_0}(\tau-\theta) \right|e^{-2\lambda|\tau|} + \wp_{p} |\PhysicalState_{a0} - \PhysicalState_{b0}|,
\end{equation*}
where $\wp_p$ is a positive constant grouping the previous structural terms.
With the same arguments as those used in Step 3, we conclude, for all $k\geq 0$,
\begin{equation}
    \sup_{\tau \le 0, (\PhysicalState_{a0}, \PhysicalState_{b0})\in \Xa}
    \left| \Delta \frac{\partial \omega_{k+1}}{\partial \PhysicalState_0} \right|
    e^{-2\lambda|\tau|} \le \frac{\wp_{p}}{1-e^{-(1-2\lambda)\theta}}
    |\PhysicalState_{a0} - \PhysicalState_{b0}|
\end{equation}
This establishes the sequence $\left\{\frac{\partial \omega_k}{\partial \PhysicalState_0}\right\}$
is Lipschitz in $[-q,0]\times\Xa$, uniformly in $k$, for any $q>0$.
This implies this sequence is 
equicontinuous on 
$[-q,0]\times \Xa$, for any $q>0$.

\vspace{1em}\noindent
\textbf{Step 6: Convergence}\\

For $q=1$, Arzela-Ascoli 
theorem guarantees the existence of a subsequence $\{k_{1,j}\}$ such that
$\left\{\frac{\partial \omega_{k_{1,j}}}{\partial \PhysicalState_0}\right\}$ converges uniformly on $[-1, 0]\times \Xa $ to a Lipschitz limit, say 
$D\varpi_1$. For $q=2$ we obtain a subsequence $\{k_{2,j}\}$ of the previous subsequence such that
$\left\{\frac{\partial \omega_{k_{2,j}}}{\partial \PhysicalState_0}\right\}$ converges uniformly on $[-2, 0]\times \Xa $ to a Lipschitz limit, say 
$D\varpi_2$ which coincides with $D\varpi_1$ on $[-1,0]$. We go on recursively for all integer $q$

We remark that, since $\Xa$ is convex, the Mean Value Theorem gives, for any $\PhysicalState_0$ in $\Xa$, any unit vector $v$ and any non zero $\delta$ such that $\PhysicalState_0+\delta v$ is also in $\Xa$,
\begin{align}
\nonumber
\frac{\left|
\omega _{k}(\tob,\PhysicalState_0+\delta v)
-
\omega _{k}(\tob,\PhysicalState_0)
-
\delta \frac{\partial \omega _{k}}{\partial \PhysicalState_0}(\tob,\PhysicalState_0)v
\right|
}{\delta }
&\nonumber=
\left|\int_0^1
\left[\frac{\partial \omega _{k}}{\partial \PhysicalState_0}(\tob,\PhysicalState_0+s\delta  v)
-
\frac{\partial \omega _{k}}{\partial \PhysicalState_0}(\tob,\PhysicalState_0)
\right] vds\right|
\\\label{LP36-new}
&\leq
\frac{ \wp_p}{1-e^{- (1-2\lambda)\theta }} e^{2\lambda |\tob|}\,  \frac{\delta  }{2}.
\end{align}
The convergence of subsequences $\{\omega _{k_{q,j}}\}$ and
$\left\{\frac{\partial \omega _{k_{q,j}}}{\partial \PhysicalState_0}\right\}$ to $\varpi $ and
$D\varpi_q $, respectively, being uniform on $[-q, 0]\times \Xa$, by taking the limit
in (\ref{LP36-new}) when first $k_{q,j}$ goes to infinity and then when $\delta $ goes to $0$, we get that the limit
of the derivatives $D\varpi_q $ is the derivative
$\frac{\partial \varpi}{\partial \PhysicalState_0}$ of the limit $\varpi$.

This establishes that, for each $\tau \leq 0$, the function
$\PhysicalState_0\mapsto \varpi (\tau,\PhysicalState_0)$ is $C^1$ with Lipschitz derivative
in $\Xa$.

\section{Injectivity of the State-to-Latent Map (Theorem \ref{thm:injectivity})}
\label{app:proof_injectivity}

This appendix is dedicated to the proof of Theorem \ref{thm:injectivity}. The proof relies on an asymptotic analysis of the contracting latent dynamics when driven by slow-varying inputs. We construct a sequence of functions that approximates the exact unique bounded solution of the filter, and we show that this approximation is injective with respect to the input and its time derivatives.

\subsection{Preamble and Definitions}
\label{app:sec:preamble}

\paragraph{Restrictions and Bounds.}
We assume the base dynamics $\BaseDynamics_\Params: \LatentSpace \times \ObsSpace \to \LatentSpace$ satisfy the regularity of Assumption \ref{ass:regularity} and the monotonicity property of Assumption \ref{ass:contraction}. Consequently, all the bounds defined in \eqref{bound-for-g} are available on the compact set $\Zcal \times \Ycal$. We denote $\bar{\Observation} = \max_{\Observation \in \Ycal} |\Observation|$.

\paragraph{Input Derivatives and Dummy Variables.}
We are interested in the first $\ell$ derivatives with respect to the observer time $\tau$ of the input $\obsLatent(\tau)$, with $\ell$ fixed by the observability index (Assumption \ref{ass:observability}).
We introduce functions whose arguments will ultimately be the time sequence:
\begin{equation*}
 \tau \mapsto \left(\obsLatent(\tau), \frac{d\obsLatent}{d\tau}(\tau), \dots, \frac{d^i\obsLatent}{d\tau^i}(\tau)\right).
\end{equation*}
To establish properties independent of specific time trajectories, we introduce dummy variables $\eta_i \in \ObsSpace$ acting as generic arguments. We use the notation compact block notations:
\begin{equation}
 \bfobsLatent_i(\tau) = \left(\frac{d\obsLatent}{d\tau}(\tau), \dots, \frac{d^i\obsLatent}{d\tau^i}(\tau)\right)^\top, \qquad \bfdum_i = (\eta_1, \dots, \eta_i)^\top.
\end{equation}
We define a weighted magnitude function $\rho$ to scale the successive time derivatives appropriately:
\begin{equation}
 \rho(\bfdum_{\ell+1}) = \max\left\{|\eta_1|, |\eta_2|^{1/2}, \dots, |\eta_{\ell+1}|^{1/(\ell+1)}\right\}.
 \label{eq:def_rho}
\end{equation}
For any $\varepsilon > 0$, we define the compact domain $\bfcaldum(\varepsilon)$ as:
\begin{equation}
 \bfcaldum(\varepsilon) = \left\{(\eta_0, \bfdum_{\ell+1}) : |\eta_0| \le \bar{\Observation}, \rho(\bfdum_{\ell+1}) \le \varepsilon \right\}.
 \label{eq:def_u}
\end{equation}
An input function $\tau \mapsto \obsLatent(\tau)$ is said to be \textit{admissible} if there exists a sufficiently small $\varepsilon_{ad} > 0$ such that:
\begin{equation}
 (\obsLatent(\tau), \bfobsLatent_{\ell+1}(\tau)) \in \bfcaldum(\varepsilon_{ad}) \qquad \forall \tau \in \RR.
 \label{eq:def_admissible}
\end{equation}

\protect
\paragraph{The Operator \texorpdfstring{$\protect\edot{-}$}{-}.}
To manipulate time derivatives algebraically, we introduce the operator 
$\edot{-}$. It maps a continuously differentiable function 
$(\eta_0, \bfdum_i) \mapsto p(\eta_0, \bfdum_i)$ to a function $(\eta_0, \bfdum_{i+1}) \mapsto \edot{p}(\eta_0, \bfdum_{i+1})$ defined as:
\begin{equation}
 \edot{p}(\eta_0, \bfdum_{i+1}) = \sum_{k=0}^i \frac{\partial p}{\partial \eta_k}(\eta_0, \bfdum_i) \eta_{k+1}.
 \label{eq:operator_edot}
\end{equation}
By construction, $\edot{\eta}_i = \eta_{i+1}$. Evaluated along a valid time trajectory where $\eta_i = \obsLatent_i(\tau)=\frac{d^i \obsLatent}{d\tau^i}(\tau)$, this operator provides the exact time derivative:
\begin{equation}
 \edot{p}(\obsLatent(\tau), \bfobsLatent_{i+1}(\tau)) = \frac{d}{d\tau} \left\{ p(\obsLatent(\tau), \bfobsLatent_i(\tau)) \right\}.
 \label{eq:edot_time_eval}
\end{equation}

\paragraph{The Implicit Function $\psi$.}
We need to invert the base dynamics with respect to the latent state when subject to a constant additive forcing $v$. We define the function $\psi$ implicitly as the solution to:
\begin{equation}
 \BaseDynamics_\Params(\psi(v, \eta_0), \eta_0) = v.
 \label{eq:def_psi}
\end{equation}

\begin{lemma}
\label{lem:def_psi}
Under Assumptions \ref{ass:regularity} and \ref{ass:contraction}, there exists a unique function $\psi : \{(v, \eta_0) : |v| \le \bar{\BaseDynamics}, |\eta_0| \le \bar{\Observation}\} \to \Zcal$ which solves \eqref{eq:def_psi}. Furthermore, $\psi$ inherits the smoothness of $\BaseDynamics_\Params$.
\end{lemma}
\begin{proof}
Consider the auxiliary dynamics $\frac{d\xi}{d\tau} = \BaseDynamics_\Params(\xi, \eta_0) - v$. As demonstrated in Appendix \ref{app:remark_invariance_perturbed}, for any fixed $v \in \LatentSpace$ satisfying $|v| \le \bar{\BaseDynamics}$, the compact convex set $\Zcal$ remains forward-invariant. Therefore, there exists an equilibrium point $\xi_{eq} \in \Zcal$ (see e.g., \cite[Theorem I.8.2]{Hale1980}). This equilibrium satisfies $\BaseDynamics_\Params(\xi_{eq}, \eta_0) = v$. Due to the contraction property inside $\Zcal$ (Assumption \ref{ass:contraction}), this equilibrium is unique. We thus well-define the map $(v, \eta_0) \mapsto \psi(v, \eta_0) := \xi_{eq}$. Its smoothness follows directly from the Implicit Function Theorem, since $\BaseDynamics_\Params$ is smooth and its Jacobian $\frac{\partial \BaseDynamics_\Params}{\partial \LatentState}$ is negative definite (hence invertible) everywhere in $\Zcal \times \Ycal$.
\end{proof}

\subsection{Definition of the Approximation Sequence}

As established previously, the  state-to-latent map $\Encoder_\Gain$ defined in \eqref{eq_DefEncoder} relies on the unique bounded latent trajectory.  To lighten the notation in the following approximation analysis, we drop the explicit dependence on the initial physical state $\PhysicalState_0$ and simply denote this trajectory as $\varpi(\tau)$ when driven by a given observation signal. 
However, because $\varpi(\tau)$ is the solution to a differential equation, its injectivity with respect to the physical state cannot be analyzed directly. To overcome this, we construct an algebraic approximation of $\varpi(\tau)$. Because the contracting latent dynamics heavily filter the input, the latent state relies primarily on the instantaneous observation $\obsLatent(\tau)$ and its recent history. We thus approximate the exact state $\varpi(\tau)$ using a sequence of static maps $\varphi_i$ that depend explicitly on the observation and its first $i$ time derivatives.

We define this sequence of approximation functions $\varphi_i$ via the following recursion:
\begin{align}
\varphi_{-1} &= 0, \nonumber \\
\varphi_0(\eta_0) &= \psi(0, \eta_0), \nonumber \\
\varphi_i(\eta_0, \bfdum_i) &= \psi(\edot{\varphi}_{i-1}(\eta_0, \bfdum_i), \eta_0) \qquad \forall i \in \{1, \dots, \ell+1\}.
 \label{eq:def_phi_i}
\end{align}
Two direct consequences follow from this definition:
\begin{enumerate}
 \item Since the definition \eqref{eq:operator_edot} of the operator $\edot{-}$ gives readily $\edot{\varphi}_{i-1}(\eta_0, 0, \dots, 0) = 0$, we have:
 \begin{equation}
  \varphi_i(\eta_0, 0, \dots, 0) = \psi(0, \eta_0) = \varphi_0(\eta_0).
  \label{eq:phi_zero_eval}
 \end{equation}
 \item The implicit relation \eqref{eq:def_psi} of $\psi$ directly yields the fundamental identity:
 \begin{equation}
  \edot{\varphi}_{i-1}(\eta_0, \bfdum_i) = \BaseDynamics_\Params(\varphi_i(\eta_0, \bfdum_i), \eta_0).
  \label{eq:phi_fundamental_identity}
 \end{equation}
\end{enumerate}

\begin{lemma}[Well-Definedness and Bounds on the Approximations]
\label{lem:boundedness_phi}
Suppose Assumptions \ref{ass:regularity} and \ref{ass:contraction} are satisfied. There exists an admissibility threshold $\varepsilon_{ad} > 0$ such that the functions $\varphi_i$ are well-defined on the compact domain $\bfcaldum(\varepsilon_{ad})$ and take values in $\Zcal$. Furthermore, if $\BaseDynamics_\Params$ is $C^q$ (with $q \ge \ell+1$), then $\varphi_i$ is $C^{q-i}$, and there exist constants $\bar{\phi}$ and $\overline{\partial \phi}$ such that for all $i \in \{0, \dots, \ell\}$, and all $(\eta_0, \bfdum_{\ell+1}) \in \bfcaldum(\varepsilon_{ad})$:
\begin{align}
 |\varphi_i(\eta_0, \bfdum_i)| &\le 2\bar{\BaseDynamics} \label{eq:lemma_bound_phi_i} \\
 |\varphi_{i+1}(\eta_0, \bfdum_{i+1}) - \varphi_i(\eta_0, \bfdum_i)| &\le \rho(\bfdum_{\ell+1})^{i+1} \bar{\phi} \label{eq:lemma_bound_diff_phi}
\end{align}
Moreover, for any two points $(\eta_{a0}, \bfdum_{a(\ell+1)})$ and $(\eta_{b0}, \bfdum_{b(\ell+1)})$ in $\bfcaldum(\varepsilon_{ad})$:
\begin{align}
 &\left| [\varphi_{i+1}(\eta_{a0}, \bfdum_{a(i+1)}) - \varphi_i(\eta_{a0}, \bfdum_{ai})] - [\varphi_{i+1}(\eta_{b0}, \bfdum_{b(i+1)}) - \varphi_i(\eta_{b0}, \bfdum_{bi})] \right| \nonumber \\
 &\quad \le \overline{\partial \phi} \sum_{k=0}^{i+1} \max\left\{\rho(\bfdum_{a(\ell+1)}), \rho(\bfdum_{b(\ell+1)})\right\}^{i+1-k} |\eta_{ak} - \eta_{bk}|. \label{eq:lemma_bound_diff_diff_phi}
\end{align}
\end{lemma}

\begin{proof}
We proceed in three steps: well-definedness by induction, bounding the approximation differences via homogeneity scaling, and finally establishing the local Lipschitz bound.

\paragraph{1. Well-definedness (Induction on $i$).}
\textit{Base case ($i=0$):} By Lemma \ref{lem:def_psi}, the function $\eta_0 \mapsto \varphi_0(\eta_0) = \psi(0, \eta_0)$ solving $\BaseDynamics_\Params(\varphi_0(\eta_0), \eta_0) = 0$ is well-defined, unique, takes values in $\Zcal$, and is $C^q$. We set $\varepsilon_0 = 1$. The bound \eqref{eq:lemma_bound_phi_i} is trivially satisfied since $\Zcal$ is a ball of radius $2\bar{\BaseDynamics}$.

\textit{Inductive step:} Assume that, for some $i \le \ell$, there exists $\varepsilon_i \in (0, 1]$ such that for all $\varepsilon \le \varepsilon_i$, the functions $\varphi_0, \dots, \varphi_i$ are well-defined on the compact set $\bfcaldum(\varepsilon_i)$, take values in $\Zcal$, and are $C^{q-j}$ respectively. By this inductive hypothesis, the partial derivatives of $\varphi_i$ are continuous and bounded on $\bfcaldum(\varepsilon_i)$. Let us define the constant:
\begin{equation*}
 L_{\varphi_i} := (i+1) \max_{k \in \{0, \dots, i\}} \left( \sup_{(\eta_0, \bfdum_i) \in \bfcaldum(\varepsilon_i)} \left| \frac{\partial \varphi_i}{\partial \eta_k}(\eta_0, \bfdum_i) \right| \right).
\end{equation*}
Recalling the definition \eqref{eq:operator_edot} of the operator $\edot{-}$, we have:
\begin{equation*}
 |\edot{\varphi}_i(\eta_0, \bfdum_{i+1})| = \left| \sum_{k=0}^i \frac{\partial \varphi_i}{\partial \eta_k}(\eta_0, \bfdum_i) \eta_{k+1} \right| \le \frac{1}{i+1} L_{\varphi_i} \sum_{k=0}^i |\eta_{k+1}|.
\end{equation*}
For any $(\eta_0, \bfdum_{i+1}) \in \bfcaldum(\varepsilon_i)$ with $\varepsilon_i \le 1$, the definition of $\rho$ ensures $|\eta_{k+1}| \le \rho(\bfdum_{i+1})^{k+1} \le \rho(\bfdum_{i+1})$. This yields:
\begin{equation}
 |\edot{\varphi}_i(\eta_0, \bfdum_{i+1})| \le L_{\varphi_i} \rho(\bfdum_{i+1}).
 \label{eq:forcing_bound}
\end{equation}
The subsequent function is defined as $\varphi_{i+1}(\eta_0, \bfdum_{i+1}) = \psi(\edot{\varphi}_i(\eta_0, \bfdum_{i+1}), \eta_0)$. Lemma \ref{lem:def_psi} guarantees this is well-defined and evaluates inside $\Zcal$ provided that the forcing magnitude $|\edot{\varphi}_i| \le \bar{\BaseDynamics}$. By choosing the refined admissibility threshold:
\begin{equation*}
 \varepsilon_{i+1} = \min\left(\varepsilon_i, \frac{\bar{\BaseDynamics}}{L_{\varphi_i}}\right),
\end{equation*}
we ensure that $\rho(\bfdum_{i+1}) \le \varepsilon_{i+1} \implies |\edot{\varphi}_i| \le \bar{\BaseDynamics}$. Thus $\varphi_{i+1}$ is well-defined and $C^{q-(i+1)}$. Setting $\varepsilon_{ad} = \varepsilon_{\ell+1}$ concludes the induction, proving \eqref{eq:lemma_bound_phi_i} up to $\ell+1$.

\paragraph{2. Scaling and Difference Bound.}
To prove \eqref{eq:lemma_bound_diff_phi}, we apply the Hadamard Lemma to $\psi$ between the two forcings:
\begin{align*}
 \varphi_{i+1}(\eta_0, \bfdum_{i+1}) - \varphi_i(\eta_0, \bfdum_i) &= \psi(\edot{\varphi}_i(\eta_0, \bfdum_{i+1}), \eta_0) - \psi(\edot{\varphi}_{i-1}(\eta_0, \bfdum_i), \eta_0) \\
 &= K_i(\eta_0, \bfdum_{i+1}) \left[ \edot{\varphi}_i(\eta_0, \bfdum_{i+1}) - \edot{\varphi}_{i-1}(\eta_0, \bfdum_i) \right],
\end{align*}
where $K_i(\eta_0, \bfdum_{i+1}) = \int_0^1 \frac{\partial \psi}{\partial v}(s\edot{\varphi}_i + (1-s)\edot{\varphi}_{i-1}, \eta_0) \, ds$ is $C^{q-i-1}$ and bounded on $\bfcaldum(\varepsilon_{ad})$.

Let $r = \rho(\bfdum_{\ell+1}) \le \varepsilon_{ad}$. We define the scaled variables $\xi_k = \frac{\eta_k}{r^k}$, denoting $\bfdumr_i = (\xi_1, \dots, \xi_i)^\top$, and $\xi_0 = \eta_0$. The original variables satisfy $\eta_k = r^k \xi_k$.
For $i=0$, since $\edot{\varphi}_{-1} = 0$, we have $\edot{\varphi}_0(\xi_0, r\xi_1) = \frac{\partial \psi}{\partial \Observation}(0, \xi_0) r\xi_1 := r\check{p}_0(\xi_0, \xi_1)$ where $\check{p}_0$ is $C^{q-1}$. Thus:
\begin{equation*}
 \varphi_1(\xi_0, r\xi_1) - \varphi_0(\xi_0) = r K_0(\xi_0, r\xi_1)\check{p}_0(\xi_0, \xi_1) := r \tilde{p}_1(r, \xi_0, \xi_1).
\end{equation*}
By induction, if a function $p$ can be factorized as $p(\xi_0, r\xi_1, \dots, r^i\xi_i) = r^i \tilde{p}(r, \xi_0, \dots, \xi_i)$, then applying the $\edot{-}$ operator yields:
\begin{align}
 \edot{p}(\xi_0, r\xi_1, \dots, r^{i+1}\xi_{i+1}) &= \sum_{k=0}^i \frac{\partial p}{\partial \eta_k}(\xi_0, r\xi_1, \dots, r^i\xi_i) r^{k+1} \xi_{k+1} \nonumber \\
 &= r^i \sum_{k=0}^i \frac{\partial \tilde{p}}{\partial \xi_k}(r, \xi_0, \dots, \xi_i) \frac{\partial \xi_k}{\partial \eta_k} r^{k+1} \xi_{k+1} \nonumber \\
 &= r^{i+1} \sum_{k=0}^i \frac{\partial \tilde{p}}{\partial \xi_k}(r, \xi_0, \dots, \xi_i) \xi_{k+1} := r^{i+1} \check{p}(r, \xi_0, \bfdumr_{i+1}). \label{eq:edot_factorization}
\end{align}
Following this recursive factorization, there exist $C^{q-i-1}$ functions $\tilde{p}_{i+1}$ such that:
\begin{equation*}
 \varphi_{i+1}(\xi_0, r\xi_1, \dots, r^{i+1}\xi_{i+1}) - \varphi_i(\xi_0, r\xi_1, \dots, r^i\xi_i) = r^{i+1} \tilde{p}_{i+1}(r, \xi_0, \bfdumr_{i+1}).
\end{equation*}
Whenever $(\eta_0, \bfdum_{\ell+1}) \in \bfcaldum(\varepsilon_{ad})$, the scaled vector $(\xi_0, \bfdumr_{i+1})$ belongs to the compact set $\{(\xi_0, \bfdumr_{i+1}) : |\xi_0| \le \bar{\Observation}, \rho(\bfdumr_{\ell+1}) = 1\}$. By continuity, $\tilde{p}_{i+1}$ is bounded on this compact domain by a constant $\bar{\phi}$. Replacing $r$ by $\rho(\bfdum_{\ell+1})$ yields \eqref{eq:lemma_bound_diff_phi}.

\paragraph{3. Lipschitz of the Approximation Differences.}
To establish \eqref{eq:lemma_bound_diff_diff_phi}, we apply the Hadamard Lemma to the factorized function $\tilde{p}_{i+1}$ with respect to the scaled variables $\xi_k$:
\begin{equation*}
 \tilde{p}_{i+1}(r, \xi_{a0}, \bfdumr_{a(i+1)}) - \tilde{p}_{i+1}(r, \xi_{b0}, \bfdumr_{b(i+1)}) = \sum_{k=0}^{i+1} \pi_k(r, \xi_{a0}, \bfdumr_{a(i+1)}, \xi_{b0}, \bfdumr_{b(i+1)}) (\xi_{ak} - \xi_{bk}),
\end{equation*}
where $\pi_k$ are continuous functions. Multiplying both sides by $r^{i+1}$ gives the difference in the original unscaled functions $p = \varphi_{i+1} - \varphi_i$:
\begin{align*}
 p(\eta_{a0}, \bfdum_{a(i+1)}) - p(\eta_{b0}, \bfdum_{b(i+1)}) &= r^{i+1} \sum_{k=0}^{i+1} \pi_k(\dots) (\xi_{ak} - \xi_{bk}) \\
 &= \sum_{k=0}^{i+1} r^{i+1-k} \pi_k(\dots) (r^k \xi_{ak} - r^k \xi_{bk}) \\
 &= \sum_{k=0}^{i+1} \pi_k(\dots) r^{i+1-k} (\eta_{ak} - \eta_{bk}).
\end{align*}
Let $r = \max\{\rho(\bfdum_{a(\ell+1)}), \rho(\bfdum_{b(\ell+1)})\}$. Since $r \le \varepsilon_{ad}$, the scaled arguments lie in a compact set, meaning there exists a uniform upper bound $\overline{\partial \phi}$ for all functions $|\pi_k(\dots)|$. Factoring out this bound immediately yields \eqref{eq:lemma_bound_diff_diff_phi}.
\end{proof}

\subsection{Lipschitz Injectivity of the Approximation Map}

The next crucial step is to prove that the approximation map $\varphi_\ell$ is injective on $\bfcaldum(\varepsilon)$ when $\varepsilon$ is sufficiently small. This ensures that the time-derivatives of the input observation can be uniquely recovered from the latent state approximation.

\begin{proposition}[Lipschitz Injectivity of the Approximation]
\label{prop:approx_map_injectivity}
Suppose Assumptions \ref{ass:regularity}, \ref{ass:contraction}, and \ref{ass:filter_capacity} are satisfied. Then, there exist real numbers $\varepsilon_{inj} \in (0, \varepsilon_{ad}]$ and $L_{inj} > 0$ such that, for any two pairs $(\eta_{a0}, \bfdum_{a\ell})$ and $(\eta_{b0}, \bfdum_{b\ell})$ in the compact set $\bfcaldum(\varepsilon_{inj})$, we have:
\begin{equation}
 |\eta_{a0} - \eta_{b0}| + |\bfdum_{a\ell} - \bfdum_{b\ell}| \le L_{inj} |\varphi_\ell(\eta_{a0}, \bfdum_{a\ell}) - \varphi_\ell(\eta_{b0}, \bfdum_{b\ell})|.
 \label{eq:approx_map_lipschitz_injectivity}
\end{equation}
\end{proposition}

\begin{proof}
The proof is organized into four main steps. First, we study the injectivity of the base map $\varphi_0$. Second, we derive a recursive formula for the Jacobian of $\varphi_i$, leading to an explicit expression for $J_\ell(\eta_0, 0)$ and establishing its left-invertibility. Third, we prove the injectivity of $\varphi_\ell$ on $\bfcaldum(\varepsilon_{inj})$ via a contradiction argument. Finally, we deduce the uniform Lipschitz injectivity bound.

Let $\varepsilon_{ad}$ be the admissibility threshold given by Lemma \ref{lem:boundedness_phi}, guaranteeing that the sequence of approximations $\varphi_i$ is well-defined, smooth, and takes values in $\Zcal$.
Let $J_i$ denote the Jacobian matrix of $\varphi_i$ with respect to all its arguments:
\begin{equation*}
 J_i(\eta_0, \bfdum_i) = \begin{bmatrix} \frac{\partial \varphi_i}{\partial \eta_0} & \frac{\partial \varphi_i}{\partial \eta_1} & \dots & \frac{\partial \varphi_i}{\partial \eta_i} \end{bmatrix}.
\end{equation*}
We are particularly interested in evaluating $J_\ell$ on the zero-derivative manifold, i.e., $J_\ell(\eta_0, 0)$. Let $A(\eta_0) = \frac{\partial \BaseDynamics_\Params}{\partial \LatentState}(\varphi_0(\eta_0), \eta_0)$ and $B(\eta_0) = \frac{\partial \BaseDynamics_\Params}{\partial \Observation}(\varphi_0(\eta_0), \eta_0)$. By the monotonicity property (Assumption \ref{ass:contraction}), $A(\eta_0)+A(\eta_0)^\top$ is strictly negative definite and therefore 
$A(\eta_0)$ is invertible.

\paragraph{1. Study of the base map $\varphi_0$.}
The map $\varphi_0(\eta_0)$ is defined by the implicit relation $\BaseDynamics_\Params(\varphi_0(\eta_0), \eta_0) = 0$. Assume there exist $\eta_{a0}, \eta_{b0} \in \Ycal$ such that $\varphi_0(\eta_{a0}) = \varphi_0(\eta_{b0}) := z^*$. This implies $\BaseDynamics_\Params(z^*, \eta_{a0}) = \BaseDynamics_\Params(z^*, \eta_{b0})$. By the Observation Injectivity property assumed in the Filter Rank Condition (Assumption \ref{ass:filter_capacity}), this requires $\eta_{a0} = \eta_{b0}$. Hence, $\varphi_0$ is injective.
Moreover, differentiating the implicit relation using the chain rule yields:
\begin{equation*}
 \frac{\partial \BaseDynamics_\Params}{\partial \LatentState}(\varphi_0(\eta_0), \eta_0) \frac{\partial \varphi_0}{\partial \eta_0}(\eta_0) + \frac{\partial \BaseDynamics_\Params}{\partial \Observation}(\varphi_0(\eta_0), \eta_0) = 0 \implies A(\eta_0) J_0(\eta_0) + B(\eta_0) = 0.
\end{equation*}
This explicitly gives $J_0(\eta_0) = -A(\eta_0)^{-1} B(\eta_0)$.

\paragraph{2. Jacobian Recursion and Left-Invertibility.}
We now derive a recursive formula for $J_{i+1}(\eta_0, 0)$. The map $\varphi_{i+1}$ satisfies the fundamental identity \eqref{eq:phi_fundamental_identity}: $\BaseDynamics_\Params(\varphi_{i+1}(\eta_0, \bfdum_{i+1}), \eta_0) = \edot{\varphi}_i(\eta_0, \bfdum_{i+1})$.
We differentiate this identity with respect to each block component $\eta_j$ of the augmented vector $(\eta_0, \dots, \eta_{i+1})$ and then evaluate the result at $\bfdum_{i+1} = 0$.

\begin{itemize}
 \item \textbf{Derivative w.r.t. $\eta_0$:}
 \begin{equation*}
  \frac{\partial \BaseDynamics_\Params}{\partial \LatentState} \frac{\partial \varphi_{i+1}}{\partial \eta_0}(\eta_0, \bfdum_{i+1}) + \frac{\partial \BaseDynamics_\Params}{\partial \Observation} = \frac{\partial}{\partial \eta_0} \left( \sum_{k=0}^{i} \frac{\partial \varphi_i}{\partial \eta_k}(\eta_0, \bfdum_i) \eta_{k+1} \right).
 \end{equation*}
 Evaluating at $\bfdum_{i+1} = 0$, the right-hand side vanishes because every term in the sum contains a factor $\eta_{k+1}$ evaluated at $0$. With $\varphi_{i+1}(\eta_0, 0) = \varphi_0(\eta_0)$, this yields:
 \begin{equation*}
  A(\eta_0) \frac{\partial \varphi_{i+1}}{\partial \eta_0}(\eta_0, 0) + B(\eta_0) = 0 \implies \frac{\partial \varphi_{i+1}}{\partial \eta_0}(\eta_0, 0) = -A(\eta_0)^{-1} B(\eta_0).
 \end{equation*}

 \item \textbf{Derivative w.r.t. $\eta_j$ (for $j \in \{1, \dots, i+1\}$):}
 \begin{equation*}
  \frac{\partial \BaseDynamics_\Params}{\partial \LatentState} \frac{\partial \varphi_{i+1}}{\partial \eta_j}(\eta_0, \bfdum_{i+1}) + 0 = \frac{\partial}{\partial \eta_j} \left( \sum_{k=0}^{i} \frac{\partial \varphi_i}{\partial \eta_k}(\eta_0, \bfdum_i) \eta_{k+1} \right).
 \end{equation*}
 Evaluating at $\bfdum_{i+1} = 0$, the sum on the right-hand side drops all terms except the one where the derivative explicitly hits $\eta_j$ (i.e., when $k+1 = j$). This gives:
 \begin{equation*}
  A(\eta_0) \frac{\partial \varphi_{i+1}}{\partial \eta_j}(\eta_0, 0) = \frac{\partial \varphi_i}{\partial \eta_{j-1}}(\eta_0, 0) \implies \frac{\partial \varphi_{i+1}}{\partial \eta_j}(\eta_0, 0) = A(\eta_0)^{-1} \frac{\partial \varphi_i}{\partial \eta_{j-1}}(\eta_0, 0).
 \end{equation*}
\end{itemize}

Combining these blocks, we obtain the recursive block-matrix relation:
\begin{align*}
 J_{i+1}(\eta_0, 0) &= \begin{bmatrix} \frac{\partial \varphi_{i+1}}{\partial \eta_0}(\eta_0, 0) & \frac{\partial \varphi_{i+1}}{\partial \eta_1}(\eta_0, 0) & \dots & \frac{\partial \varphi_{i+1}}{\partial \eta_{i+1}}(\eta_0, 0) \end{bmatrix} \\
 &= A(\eta_0)^{-1} \begin{bmatrix} -B(\eta_0) & \frac{\partial \varphi_i}{\partial \eta_0}(\eta_0, 0) & \dots & \frac{\partial \varphi_i}{\partial \eta_i}(\eta_0, 0) \end{bmatrix} \\
 &= A(\eta_0)^{-1} \begin{bmatrix} -B(\eta_0) & J_i(\eta_0, 0) \end{bmatrix}.
\end{align*}
Unrolling this recursion from $i=0$ up to $\ell$ yields the explicit form:
\begin{align*}
 J_\ell(\eta_0, 0) &= - A(\eta_0)^{-(\ell+1)} \begin{bmatrix} A(\eta_0)^\ell B(\eta_0) & A(\eta_0)^{\ell-1} B(\eta_0) & \dots & B(\eta_0) \end{bmatrix},\\
 &=- A(\eta_0)^{-(\ell+1)}C_\ell(\eta_0) P ,
\end{align*}
where $C_\ell(\eta_0)$ is 
the matrix defined in the Filter Rank Condition (Assumption \ref{ass:filter_capacity})
and $P\in\RR^{p(\ell+1)\times p(\ell+1)}$ is an invertible permutation matrix
. Since $C_\ell(\eta_0)$ is  left-invertible by assumption, $J_\ell(\eta_0, 0)$ is also left-invertible for all $\eta_0 \in \Ycal$. By the Local Inversion Theorem, $\varphi_\ell$ is locally injective in an open neighborhood around every point $(\eta_0, 0)$.

\paragraph{3. Existence of $\varepsilon_{inj}$ via Contradiction.}
We prove that there exists $\varepsilon_{inj} \in (0, \varepsilon_{ad}]$ such that $\varphi_\ell$ is injective on the entire set $\bfcaldum(\varepsilon_{inj})$. We proceed by contradiction. Assume no such $\varepsilon_{inj}$ exists.
Then, for every $k \in \mathbb{N}^*$, we can pick $\epsilon_k = 1/k$. There must exist two distinct points $(\eta_{ak0}, \bfdum_{ak\ell}) \neq (\eta_{bk0}, \bfdum_{bk\ell})$ satisfying:
\begin{enumerate}
 \item $(\eta_{ak0}, \bfdum_{ak\ell}) \in \bfcaldum(\epsilon_k)$ and $(\eta_{bk0}, \bfdum_{bk\ell}) \in \bfcaldum(\epsilon_k)$,
 \item $\varphi_\ell(\eta_{ak0}, \bfdum_{ak\ell}) = \varphi_\ell(\eta_{bk0}, \bfdum_{bk\ell})$.
\end{enumerate}
Since all points lie in the compact set $\bfcaldum(\varepsilon_{ad})$, we can extract convergent subsequences. Let the limits be $(\eta_{ak0}, \bfdum_{ak\ell}) \to (\eta_{a*0}, \bfdum_{a*\ell})$ and $(\eta_{bk0}, \bfdum_{bk\ell}) \to (\eta_{b*0}, \bfdum_{b*\ell})$.
Because $\rho(\bfdum_{ak\ell}) \le 1/k \to 0$ and $\rho(\bfdum_{bk\ell}) \le 1/k \to 0$, the limits of the derivative vectors must be strictly zero: $\bfdum_{a*\ell} = 0$ and $\bfdum_{b*\ell} = 0$.
By the continuity of $\varphi_\ell$, taking the limit in the equality yields:
\begin{equation*}
 \varphi_\ell(\eta_{a*0}, 0) = \lim_{k\to\infty} \varphi_\ell(\eta_{ak0}, \bfdum_{ak\ell}) = \lim_{k\to\infty} \varphi_\ell(\eta_{bk0}, \bfdum_{bk\ell}) = \varphi_\ell(\eta_{b*0}, 0).
\end{equation*}
Using \eqref{eq:phi_zero_eval}, we know that $\varphi_\ell(\eta_0, 0) = \varphi_0(\eta_0)$. Thus, $\varphi_0(\eta_{a*0}) = \varphi_0(\eta_{b*0})$.
By the global injectivity of $\varphi_0$ (proved in Step 1), this forces $\eta_{a*0} = \eta_{b*0} := \eta_{*0}$. Both sequences thus converge to the exact same limit point $(\eta_{*0}, 0)$.

However, we established in Step 2 that $J_\ell(\eta_{*0}, 0)$ is left-invertible, meaning $\varphi_\ell$ is locally injective in some neighborhood $\mathcal{N}(\eta_{*0}, 0)$. For $k$ sufficiently large, both distinct points $(\eta_{ak0}, \bfdum_{ak\ell})$ and $(\eta_{bk0}, \bfdum_{bk\ell})$ will simultaneously enter this neighborhood while still being mapped to the identical value $\varphi_\ell(\dots)$. This strictly contradicts the local injectivity. Thus, a strictly positive $\varepsilon_{inj}$ ensuring global injectivity on $\bfcaldum(\varepsilon_{inj})$ must exist.

\paragraph{4. Lipschitz Injectivity Bound.}
The function $\varphi_\ell$ is injective and continuously differentiable on the compact convex set $\bfcaldum(\varepsilon_{inj})$, and its Jacobian is full column rank everywhere on this set. Standard analysis results guarantee that the inverse map is globally Lipschitz on the image set $\varphi_\ell(\bfcaldum(\varepsilon_{inj}))$. This guarantees the existence of the constant $L_{inj} > 0$ such that \eqref{eq:approx_map_lipschitz_injectivity} holds.
\end{proof}

\subsection{Approximation Error Dynamics and Fading Memory}

We now study the properties of the true bounded solution $\varpi(\tau)$ relative to its approximation $\varphi_\ell$. We define the approximation error as $e_\ell(\obsLatent(\tau), \bfobsLatent_\ell(\tau)) = \varpi(\tau) - \varphi_\ell(\obsLatent(\tau), \bfobsLatent_\ell(\tau))$.

\begin{proposition}[Approximation Error Magnitude]
\label{prop:approx_quality}
Suppose Assumptions \ref{ass:regularity} and \ref{ass:contraction} are satisfied. There exists a constant $\bar{\phi}$ such that, for any admissible input function $\tau \mapsto \obsLatent(\tau)$, we have:
\begin{equation}
 \sup_{\tau \in \RR} |\varpi(\tau) - \varphi_\ell(\obsLatent(\tau), \bfobsLatent_\ell(\tau))| \le L_{gz} \bar{\phi} \left[\sup_{\tau \in \RR} \rho(\bfobsLatent_{\ell+1}(\tau))\right]^{\ell+1}.
 \label{eq:approx_error_bound}
\end{equation}
\end{proposition}

\begin{proof}
As established in \eqref{eq:edot_time_eval} and \eqref{eq:phi_fundamental_identity}, when the argument of $\varphi_\ell$ is the valid time trajectory, its derivative is exactly:
\begin{equation*}
 \frac{d}{d\tau} \left\{ \varphi_\ell(\obsLatent(\tau), \bfobsLatent_\ell(\tau)) \right\} = \edot{\varphi}_\ell(\obsLatent(\tau), \bfobsLatent_{\ell+1}(\tau)) = \BaseDynamics_\Params(\varphi_{\ell+1}(\obsLatent(\tau), \bfobsLatent_{\ell+1}(\tau)), \obsLatent(\tau)).
\end{equation*}
Using the shorthand notation $\varphi_i(\tau) = \varphi_i(\obsLatent(\tau), \bfobsLatent_i(\tau))$, the dynamics of the approximation error are:
\begin{equation}
 \frac{d e_\ell}{d\tau}(\tau) = \frac{d\varpi}{d\tau}(\tau) - \frac{d\varphi_\ell}{d\tau}(\tau) = \BaseDynamics_\Params(\varpi(\tau), \obsLatent(\tau)) - \BaseDynamics_\Params(\varphi_{\ell+1}(\tau), \obsLatent(\tau)).
 \label{eq:edot_raw_diff}
\end{equation}
Applying the Mean Value Theorem to the first term yields:
\begin{align*}
 \BaseDynamics_\Params(\varpi(\tau), \obsLatent(\tau)) &= \BaseDynamics_\Params(\varphi_\ell(\tau) + e_\ell(\tau), \obsLatent(\tau)) \\
 &= \BaseDynamics_\Params(\varphi_\ell(\tau), \obsLatent(\tau)) + \underbrace{\left( \int_0^1 \frac{\partial \BaseDynamics_\Params}{\partial \LatentState}(\varphi_\ell(\tau) + s e_\ell(\tau), \obsLatent(\tau)) \, ds \right)}_{:= A_\ell(\tau)} e_\ell(\tau).
\end{align*}
Similarly, with the notation $\phi_{\ell+1}(\tau) = \varphi_{\ell+1}(\tau) - \varphi_\ell(\tau)$:
\begin{align*}
 \BaseDynamics_\Params(\varphi_{\ell+1}(\tau), \obsLatent(\tau)) &= \BaseDynamics_\Params(\varphi_\ell(\tau) + \phi_{\ell+1}(\tau), \obsLatent(\tau)) \\
 &= \BaseDynamics_\Params(\varphi_\ell(\tau), \obsLatent(\tau)) + \underbrace{\left( \int_0^1 \frac{\partial \BaseDynamics_\Params}{\partial \LatentState}(\varphi_\ell(\tau) + s \phi_{\ell+1}(\tau), \obsLatent(\tau)) \, ds \right)}_{:= B_\ell(\tau)} \phi_{\ell+1}(\tau).
\end{align*}
Note that by Lemma \ref{lem:boundedness_phi}, $|\phi_{\ell+1}(\tau)| \le \bar{\phi} \rho(\bfobsLatent_{\ell+1}(\tau))^{\ell+1}$ as long as the input is admissible.
Substituting these expansions back into \eqref{eq:edot_raw_diff}, the common term $\BaseDynamics_\Params(\varphi_\ell, \obsLatent)$ cancels out:
\begin{equation}
 \frac{d e_\ell}{d\tau}(\tau) = A_\ell(\tau) e_\ell(\tau) - B_\ell(\tau) \phi_{\ell+1}(\tau).
 \label{eq:edot_final}
\end{equation}
Because the sets $\Zcal$ and $\Ycal$ are convex and invariant, the paths of integration remain inside $\Zcal$. By the contraction assumption, $A_\ell(\tau) + A_\ell(\tau)^\top \le -2 I_m$. Moreover, $|B_\ell(\tau)| \le L_{gz}$. Applying Lemma \ref{lemma-dif-ineg} gives:
\begin{equation*}
 |e_\ell(\tau)| \le e^{-(\tau - \tau_0)} |e_\ell(\tau_0)| + \int_{\tau_0}^\tau e^{-(\tau - s)} L_{gz} \bar{\phi} \rho(\bfobsLatent_{\ell+1}(s))^{\ell+1} \, ds.
\end{equation*}
Since both $\varpi(\tau)$ and $\varphi_\ell(\tau)$ are bounded on $\RR$, $e_\ell(\tau)$ is bounded. Letting $\tau_0 \to -\infty$, the initial condition term vanishes, and bounding the supremum of $\rho$ directly yields \eqref{eq:approx_error_bound}.
\end{proof}

We now examine how this approximation error responds to input variations. We consider two admissible input histories, $\tau\mapsto \obsLatent_a{}(\tau)$ and $\tau\mapsto \obsLatent_b{}(\tau)$, generating solutions $\tau\mapsto \varpi_a(\tau)$ and $\tau\mapsto \varpi_b(\tau)$. Let $\Delta e_\ell{} = e_{\ell, a}{} - e_{\ell, b}{}$,
$\Delta \bfobsLatent_k{}= \bfobsLatent_{ak}{} - \bfobsLatent_{bk}{}$
and $\Delta \obsLatent_k{} = \obsLatent_{ak}{} - \obsLatent_{bk}{}$
where $\obsLatent_{ak}{}=\frac{d ^k\obsLatent_{a}}{d\tau^k}{}$.

\begin{corollary}[Lipschitz Property of the Error w.r.t. Input History]
\label{cor:lipschitz_e_ell}
Under the assumptions of Proposition \ref{prop:approx_map_injectivity}, there exist constants $c_e, c_u, c_{\ell+1} > 0$ such that for any decay rate $\lambda \in [0, 1/2)$ and bound $K > 0$, there exist global constants $L_{e\bfobsLatent}$ and $L_{e\obsLatent_{\ell+1}}$ such that, if the inputs take values in $\bfcaldum(\rho_{fm})$ with
\begin{equation*}
\rho_{fm} \in \left[0, \min\left\{1, \left(\frac{1/2 - \lambda}{c_e}\right)^{\frac{1}{\ell+1}}\right\}\right),
\end{equation*}
and satisfy the fading memory property for all $s \le 0$:
\begin{equation}
e^{\lambda s} \sum_{k=0}^\ell \rho_{fm}^{\ell-k} |\Delta \obsLatent_k(s)| \le K \sum_{k=0}^\ell \rho_{fm}^{\ell-k} |\Delta \obsLatent_k(0)|,
\label{eq:fading_memory_assumption}
\end{equation}
then the approximation error difference at $\tau = 0$ is bounded by:
\begin{equation}
|\Delta e_\ell(0)| \le L_{e\bfobsLatent} \rho_{fm} \sum_{k=0}^\ell \rho_{fm}^{\ell-k} |\Delta \obsLatent_k(0)| + L_{e\obsLatent_{\ell+1}} \sup_{s \le 0} e^{\lambda s} |\Delta \obsLatent_{\ell+1}(s)|.
\label{eq:lipschitz_e_ell}
\end{equation}

\end{corollary}

\begin{proof}
Subtracting the dynamics \eqref{eq:edot_final} evaluated for the two trajectories yields:
\begin{align*}
 \frac{d \Delta e_\ell}{d\tau}(\tau) &= [A_{a\ell}(\tau) e_{\ell, a}(\tau) - B_{a\ell}(\tau) \phi_{a(\ell+1)}(\tau)] - [A_{b\ell}(\tau) e_{\ell, b}(\tau) - B_{b\ell}(\tau) \phi_{b(\ell+1)}(\tau)] \\
 &= A_{a\ell}(\tau) \Delta e_\ell(\tau) + F(\tau),
\end{align*}
where $F(\tau) = \Delta A_\ell(\tau) e_{\ell, b}(\tau) - B_{a\ell}(\tau) \Delta \phi_{\ell+1}(\tau) - \Delta B_\ell(\tau) \phi_{b(\ell+1)}(\tau)$, with $\Delta A_\ell = A_{a\ell} - A_{b\ell}$ and $\Delta B_\ell = B_{a\ell} - B_{b\ell}$.

Because the sets $\Zcal$ and $\Ycal$ are convex and invariant, the paths of integration remain inside $\Zcal$. The monotonicity property guarantees $A_{a\ell}(\tau) + A_{a\ell}(\tau)^\top \le -2 I_m$. Applying Lemma \ref{lemma-dif-ineg} gives:
\begin{equation}
 |\Delta e_\ell(\tau)| \le e^{-(\tau-\tau_0)} |\Delta e_\ell(\tau_0)| + \int_{\tau_0}^\tau e^{-(\tau-s)} |F(s)| \, ds.
 \label{eq:gronwall_delta_e}
\end{equation}

Using the bounds from \eqref{bound-for-g} and Lemma \ref{lem:boundedness_phi}, since all intermediate variables lie in the invariant set $\Zcal$ and observations in $\Ycal$, we can explicitly bound each component:
\begin{align*}
|\Delta A_\ell| &\le L_{gzz} [|\Delta \varphi_{\ell}| + |\Delta e_{\ell}|] + L_{gzo} |\Delta \obsLatent| \\
|e_{\ell,b}(\tau)| &\le L_{gz} \bar{\phi}\rho(\bfobsLatent_{b(\ell+1)})^{\ell+1} \quad \text{(Proposition \ref{prop:approx_quality})} \\
 |B_{a\ell}| &\le L_{gz} \\
 |\Delta \phi_{\ell+1}| &\le \overline{\partial \phi}\sum_{k=0}^{\ell+1} \max\left\{\rho(\bfobsLatent_{a(\ell+1)}), \rho(\bfobsLatent_{b(\ell+1)})\right\}^{\ell+1-k}|\Delta \obsLatent_k| \quad \text{(Lemma \ref{lem:boundedness_phi})} \\
 |\Delta B_\ell| &\le L_{gzz}[|\Delta \varphi_{\ell}| + |\Delta \phi_{\ell+1}|] + L_{gzo} |\Delta \obsLatent| \\
 |\phi_{b(\ell+1)}| &\le \bar{\phi} \rho(\bfobsLatent_{b(\ell+1)})^{\ell+1} \quad \text{(Lemma \ref{lem:boundedness_phi})} \\
 |\Delta \varphi_{\ell}| &\le L_\varphi \sum_{k=0}^{\ell} |\Delta \obsLatent_k| \\
 |\Delta \obsLatent| &= |\Delta \obsLatent_0| \le \sum_{k=0}^{\ell} |\Delta \obsLatent_k| \le \sum_{k=0}^{\ell} \rho_{fm}^{-k}|\Delta \obsLatent_k|.
\end{align*}

Because we assume $\sup_{\tau}\max\left\{\rho(\bfobsLatent_{a(\ell+1)}(\tau)), \rho(\bfobsLatent_{b(\ell+1)}(\tau))\right\} \le \rho_{fm}$, we can substitute these bounds into the expression of $F(\tau)$ and distribute the terms:
\begin{align*}
 |F(\tau)| &\le \left(L_{gzz} [|\Delta \varphi_{\ell}| + |\Delta e_{\ell}|] + L_{gzo}|\Delta \obsLatent|\right) L_{gz} \bar{\phi} \rho_{fm}^{\ell+1} \\
 &\quad + L_{gz} \overline{\partial \phi}\sum_{k=0}^{\ell+1} \rho_{fm}^{\ell+1-k}|\Delta \obsLatent_k(\tau)| \\
 &\quad + \left(L_{gzz}\left[|\Delta \varphi_{\ell}| + \overline{\partial \phi}\sum_{k=0}^{\ell+1} \rho_{fm}^{\ell+1-k} |\Delta \obsLatent_k(\tau)|\right] + L_{gzo} |\Delta \obsLatent|\right) \bar{\phi}\rho_{fm}^{\ell+1} \\
\end{align*}
This gives,
\begin{align*}
|F(\tau)| &\le L_{gzz} L_{gz} \bar{\phi} \rho_{fm}^{\ell+1}|\Delta \varphi_{\ell}| + L_{gzz}\bar{\phi}\rho_{fm}^{\ell+1}|\Delta \varphi_{\ell}| + L_{gzz}L_{gz} \bar{\phi} \rho_{fm}^{\ell+1}|\Delta e_{\ell}(\tau)| \\
 &\quad + L_{gzo}L_{gz} \bar{\phi} \rho_{fm}^{\ell+1}|\Delta \obsLatent| + L_{gzo}\bar{\phi}\rho_{fm}^{\ell+1} |\Delta \obsLatent| \\
 &\quad + L_{gz} \overline{\partial \phi}\sum_{k=0}^{\ell+1} \rho_{fm}^{\ell+1-k}|\Delta \obsLatent_k(\tau)| + L_{gzz}\bar{\phi}\rho_{fm}^{\ell+1} \overline{\partial \phi}\sum_{k=0}^{\ell+1} \rho_{fm}^{\ell+1-k} |\Delta \obsLatent_k(\tau)| \\
\end{align*}
and,
\begin{align*}
|F(\tau)| &\le L_{gzz} [L_{gz} + 1]\bar{\phi} \rho_{fm}^{\ell+1} L_\varphi \sum_{k=0}^{\ell} |\Delta \obsLatent_k(\tau)| + L_{gzz}L_{gz} \bar{\phi} \rho_{fm}^{\ell+1} |\Delta e_{\ell}(\tau)| \\
 &\quad + L_{gzo}[L_{gz} + 1]\bar{\phi} \rho_{fm}^{\ell+1} |\Delta \obsLatent| \\
 &\quad + [L_{gz} + L_{gzz}\bar{\phi} \rho_{fm}^{\ell+1}]\overline{\partial \phi} \sum_{k=0}^{\ell} \rho_{fm}^{\ell+1-k}|\Delta \obsLatent_k(\tau)| + [L_{gz} + L_{gzz}\bar{\phi} \rho_{fm}^{\ell+1}]\overline{\partial \phi}|\Delta \obsLatent_{\ell +1}(\tau)| \\
 &\le L_{gzz}L_{gz} \bar{\phi} \rho_{fm}^{\ell+1} |\Delta e_{\ell}(\tau)| \\
 &\quad + \rho_{fm} \left([L_{gzz} L_\varphi + L_{gzo}][L_{gz} + 1]\bar{\phi} + [L_{gz} + L_{gzz}\bar{\phi} \rho_{fm}^{\ell+1}]\overline{\partial \phi} \right) \sum_{k=0}^{\ell} \rho_{fm}^{\ell-k}|\Delta \obsLatent_k(\tau)| \\
 &\quad + [L_{gz} + L_{gzz}\bar{\phi} \rho_{fm}^{\ell+1}]\overline{\partial \phi}|\Delta \obsLatent_{\ell +1}(\tau)| \\
 &= c_e \rho_{fm}^{\ell+1} |\Delta e_{\ell}(\tau)| + c_u \rho_{fm} \sum_{k=0}^{\ell} \rho_{fm}^{\ell-k}|\Delta \obsLatent_k(\tau)| + c_{\ell+1}|\Delta \obsLatent_{\ell +1}(\tau)|,
\end{align*}
where we logically bounded $\rho_{fm}^{\ell+1} \le \frac{1/2 - \lambda}{c_e}$ to define the generic constants independently of the specific observer gain:
\begin{align*}
c_e &= L_{gzz} L_{gz} \bar{\phi}, \\
c_u &= [L_{gzz} L_\varphi + L_{gzo}][L_{gz} + 1]\bar{\phi} + \left[ L_{gz} + L_{gzz} \bar{\phi} \frac{1/2 - \lambda}{c_e} \right] \overline{\partial \phi}, \\
 c_{\ell+1} &= \left[ L_{gz} + L_{gzz} \bar{\phi} \frac{1/2 - \lambda}{c_e} \right] \overline{\partial \phi}.
\end{align*}

Using this bound in \eqref{eq:gronwall_delta_e} and multiplying both sides by $e^\tau$ yields:
\begin{align*}
 e^{\tau}|\Delta e_\ell(\tau)| &\le e^{\tau_0} |\Delta e_\ell(\tau_0)| + c_e \rho_{fm}^{\ell+1} \int_{\tau_0}^\tau e^{s} |\Delta e_{\ell}(s)| ds \\
 &\quad + \int_{\tau_0}^\tau e^{s} \left( c_u \rho_{fm} \sum_{k=0}^{\ell} \rho_{fm}^{\ell-k}|\Delta \obsLatent_k(s)| + c_{\ell+1}|\Delta \obsLatent_{\ell+1}(s)| \right) ds.
\end{align*}
Applying Grönwall's inequality allows us to explicitly solve for the error magnitude:
\begin{align*}
 &e^{\tau}|\Delta e_\ell(\tau)| \le \left[ e^{\tau_0} + c_e \rho_{fm}^{\ell+1} \int_{\tau_0}^\tau {\startmodifVA e^{\tau_0} e^{c_e \rho_{fm}^{\ell+1}(\tau-r)} dr\stopmodif} \right] |\Delta e_\ell(\tau_0)| \\
 &\quad + c_e \rho_{fm}^{\ell+1} \int_{\tau_0}^\tau e^{c_e \rho_{fm}^{\ell+1}(\tau-r)} \left( \int_{\tau_0}^r e^{s} \left( c_u \rho_{fm} \sum_{k=0}^{\ell} \rho_{fm}^{\ell-k}|\Delta \obsLatent_k(s)| + c_{\ell+1}|\Delta \obsLatent_{\ell+1}(s)| \right) ds \right) dr \\
 &\quad + \int_{\tau_0}^\tau e^{s} \left( c_u \rho_{fm} \sum_{k=0}^{\ell} \rho_{fm}^{\ell-k}|\Delta \obsLatent_k(s)| + c_{\ell+1}|\Delta \obsLatent_{\ell+1}(s)| \right) ds.
\end{align*}

This elegantly simplifies into a single consolidated convolution:
\begin{align*}
 |\Delta e_\ell(\tau)| &\le e^{-(1-c_e \rho_{fm}^{\ell+1})(\tau-\tau_0)} |\Delta e_\ell(\tau_0)|
 \\
 &+ \int_{\tau_0}^\tau e^{-(1-c_e \rho_{fm}^{\ell+1})(\tau-s)} \left( c_u \rho_{fm} \sum_{k=0}^{\ell} \rho_{fm}^{\ell-k}|\Delta \obsLatent_k(s)| + c_{\ell+1}|\Delta \obsLatent_{\ell+1}(s)| \right) ds.
\end{align*}

Since both $\varpi$ and $\varphi_\ell$ are bounded on $\RR$, the approximation error $|\Delta e_\ell|$ is bounded. We can thus safely let the initial time $\tau_0 \to -\infty$, causing the initial condition term to vanish. Evaluating at the current time $\tau=0$ and utilizing the fading memory assumption \eqref{eq:fading_memory_assumption}, we obtain:
\begin{align*}
 |\Delta e_\ell(0)| &\le \int_{-\infty}^0 e^{(1-c_e \rho_{fm}^{\ell+1})s} \left( c_u \rho_{fm} K \sum_{k=0}^{\ell} \rho_{fm}^{\ell-k}|\Delta \obsPhys_k(0)| + c_{\ell+1}e^{\lambda s}|\Delta \obsLatent_{\ell +1}(s)| \right) e^{-\lambda s} ds \\
 &\le \frac{1}{1-c_e \rho_{fm}^{\ell+1}-\lambda} \left[ c_u \rho_{fm} K \sum_{k=0}^{\ell} \rho_{fm}^{\ell-k}|\Delta \obsPhys_k(0)| + c_{\ell+1}\sup_{s\le 0} e^{\lambda s}|\Delta \obsLatent_{\ell+1}(s)| \right].
\end{align*}
By enforcing the strict threshold $\rho_{fm} < \left(\frac{1/2 - \lambda}{c_e}\right)^{\frac{1}{\ell+1}}$, we guarantee that the integration denominator $1 - c_e \rho_{fm}^{\ell+1} - \lambda > 1/2$. Thus, the fractional multiplier is strictly bounded by $2$. This yields the final expression \eqref{eq:lipschitz_e_ell} with the constants explicitly given as:
\begin{align*}
 L_{e\bfobsLatent} &= 2c_u K = 2 \left([L_{gzz} L_\varphi + L_{gzo}][L_{gz} + 1]\bar{\phi} + \left[L_{gz}+L_{gzz}\bar{\phi} \frac{1/2-\lambda}{c_e}\right]\overline{\partial \phi}\right) K, \\
 L_{e\obsLatent_{\ell+1}} &= 2c_{\ell+1} = 2 \left[L_{gz} + L_{gzz}\bar{\phi} \frac{1/2-\lambda}{c_e}\right]\overline{\partial \phi}.
\end{align*}
\end{proof}

\begin{proposition}[General Input Derivative Reconstruction Bound]
\label{prop:derivative_reconstruction}
Suppose Assumptions \ref{ass:regularity}, \ref{ass:contraction}, and \ref{ass:filter_capacity} hold. Let $\lambda \in [0, 1/2)$ and $K > 0$, and let $c_e, L_{inj}, \varepsilon_{inj}, L_{e\bfobsLatent}, L_{e\obsLatent_{\ell+1}}$ be the constants defined previously.
If two admissible inputs $\obsLatent_a(\tau)$ and $\obsLatent_b(\tau)$ take values in $\bfcaldum(\rho_{fm})$ where:
\begin{equation}
 \rho_{fm} < \min\left\{1, \varepsilon_{inj}, \left(\frac{1/2-\lambda}{c_e}\right)^{\frac{1}{\ell+1}}, \frac{1}{2\sqrt{\ell+1} L_{inj} L_{e\bfobsLatent}} \right\},
 \label{eq:rho_fm_threshold}
\end{equation}
and satisfy the fading memory property \eqref{eq:fading_memory_assumption}, then the following bound holds:
\begin{equation}
\sum_{k=0}^\ell \rho_{fm}^{\ell-k} |\Delta \obsLatent_k(0)| \le 2\sqrt{\ell+1} L_{inj} \left[ |\Delta \varpi(0)| + L_{e\obsLatent_{\ell+1}} \sup_{s \le 0} e^{\lambda s} |\Delta \obsLatent_{\ell+1}(s)| \right].
\label{eq:derivative_reconstruction_bound}
\end{equation}
\end{proposition}

\begin{proof}
By definition, $\Delta e_\ell(\tau) = \Delta \varpi(\tau) - \Delta \varphi_\ell(\tau)$. 
Since the inputs evaluated up to order $\ell$ belong to $\bfcaldum(\varepsilon_{inj})$ (because $\rho_{fm} \le \varepsilon_{inj}$), Proposition \ref{prop:approx_map_injectivity} applies:
\begin{equation*}
 |\Delta \obsLatent(\tau)| + |\Delta \bfobsLatent_\ell(\tau)| \le L_{inj} |\Delta \varphi_\ell(\tau)| \le L_{inj} \left( |\Delta \varpi(\tau)| + |\Delta e_\ell(\tau)| \right).
\end{equation*}
Evaluating at $\tau=0$ and substituting the bound for $|\Delta e_\ell(0)|$ from Corollary \ref{cor:lipschitz_e_ell} yields:
\begin{align*}
 |\Delta \obsLatent(0)| + |\Delta \bfobsLatent_\ell(0)| &\le L_{inj} |\Delta \varpi(0)| \\&+ L_{inj} L_{e\bfobsLatent} \rho_{fm} \sum_{k=0}^\ell \rho_{fm}^{\ell-k} |\Delta \obsPhys_k(0)| + L_{inj} L_{e\obsLatent_{\ell+1}} \sup_{s \le 0} e^{\lambda s} |\Delta \obsLatent_{\ell+1}(s)|.
\end{align*}
Because $\rho_{fm} < 1$, we can use the Cauchy-Schwarz inequality to bound the weighted sum on the left hand side:
\begin{equation*}
 \sum_{k=0}^\ell \rho_{fm}^{\ell-k} |\Delta \obsPhys_k(0)| \le \sqrt{\sum_{k=0}^\ell \rho_{fm}^{2(\ell-k)}} \sqrt{\sum_{k=0}^\ell |\Delta \obsPhys_k(0)|^2} \le \sqrt{\ell+1} \left( |\Delta \obsLatent(0)| + |\Delta \bfobsLatent_\ell(0)| \right).
\end{equation*}
Combining these inequalities, we obtain:
\begin{equation*}
 \left[ \frac{1}{\sqrt{\ell+1}} - L_{inj} L_{e\bfobsLatent} \rho_{fm} \right] \sum_{k=0}^\ell \rho_{fm}^{\ell-k} |\Delta \obsPhys_k(0)| \le L_{inj} |\Delta \varpi(0)| + L_{inj} L_{e\obsLatent_{\ell+1}} \sup_{s \le 0} e^{\lambda s} |\Delta \obsLatent_{\ell+1}(s)|.
\end{equation*}
Thanks to the strict threshold \eqref{eq:rho_fm_threshold}, the bracketed term is strictly greater than $\frac{1}{2\sqrt{\ell+1}}$. Multiplying by $2\sqrt{\ell+1}$ directly yields \eqref{eq:derivative_reconstruction_bound}.
\end{proof}

\subsection{Synthesis: Proof of Theorem \ref{thm:injectivity}}
We are now ready to close the proof of Theorem \ref{thm:injectivity}. Recall from Appendix \ref{app:proof_embedding} that $\Encoder_\Gain$ defined in equation \eqref{eq_DefEncoder} is uniquely defined via the bounded solution of the latent dynamics. By construction, this specific mapping already satisfies the invariance, attractivity, and regularity properties established in Theorem \ref{thm:embedding}. We now prove that for a sufficiently large gain $\Gain$, this exact same mapping is also Lipschitz injective.

We consider the trajectories $\flowf(\PhysicalState_{a0}, t)$ and $\flowf(\PhysicalState_{b0}, t)$ initialized in 
the set $\X$
where Assumption \ref{ass:observability} holds (recall that the modification given in  \eqref{eq:syst_modified} preserve the satisfaction of this assumption)
and denote
$\Delta \PhysicalState (t)=\flowf(\PhysicalState_{a0},t)-\flowf(\PhysicalState_{b0},t)$.
We analyze the observations in the observer time scale $\tau = \Gain t$.
The observation derivatives are  given by scaled Lie derivatives:
\begin{equation}
 \obsLatent_{\PhysicalState_0, i}(\tau) = \frac{d^i \obsLatent_{\PhysicalState_0}}{d\tau^i}(\tau) = \frac{1}{\Gain^i} \Lie^i \GenerativeObs\left(\flowf\left(\PhysicalState_0, \frac{\tau}{\Gain}\right)\right).
 \label{eq:scaled_lie_derivatives}
\end{equation}

\paragraph{1. Verification of the Assumptions.}
We define the upper bound for the physical measurement derivatives on $\X$:
\begin{equation*}
 \bar{\obsLatent}' = \sup_{\PhysicalState \in \X} \max\left\{ |\Lie \GenerativeObs(\PhysicalState)|, |\Lie^2 \GenerativeObs(\PhysicalState)|^{1/2}, \dots, |\Lie^{\ell+1} \GenerativeObs(\PhysicalState)|^{1/(\ell+1)} \right\}.
\end{equation*}
Using \eqref{eq:scaled_lie_derivatives}, the magnitude function \eqref{eq:def_rho} satisfies:
\begin{equation*}
 \rho\left( \bfobsLatent_{\PhysicalState_0, \ell+1}(\tau) \right) \le \frac{\bar{\obsLatent}'}{\Gain}.
\end{equation*}
By choosing the gain $\Gain$ large enough such that $\Gain \ge \bar{\obsLatent}' / \varepsilon_{ad}$, the input is guaranteed to be admissible. We set $\rho_{fm} = \bar{\obsLatent}' / \Gain$. 

Next, we verify the fading memory property \eqref{eq:fading_memory_assumption}. 
The differential observability on $\X$ (Assumption \ref{ass:observability}) gives
$$
c_{obs}|\Delta \PhysicalState (0)| = c_{obs}|\PhysicalState_{a0}-\PhysicalState_{b0}| \le \sum_{i=0}^\ell |\Lie^i \GenerativeObs(\PhysicalState_{a0}) - \Lie^i \GenerativeObs(\PhysicalState_{b0})| = \sum_{i=0}^\ell  |\Delta \obsLatent_i(0)|$$
Scaling this with $\Gain \geq 1$ yields:
\begin{equation}
 c_{obs} |\Delta \PhysicalState(0)| \le \sum_{i=0}^\ell \Gain^i |\Delta \obsLatent_i(0)|.
 \label{eq:obs_scaled}
\end{equation}
By the global Lipschitz property of the Lie derivatives (with constant $L_H = \sum_{i=0}^\ell \text{Lip}(\Lie^i \GenerativeObs)$), we have $\Gain^j |\Delta \obsLatent_j(s)| \le \text{Lip}(\Lie^j \GenerativeObs) |\Delta \PhysicalState(s/\Gain)|$. Because the physical flow (recall that the modification given in  \eqref{eq:syst_modified})
is governed by a Lipschitz vector field with constant $L_\GenerativeDrift$, the backward-in-time distance is bounded by:
\begin{equation}
\label{LPN1}
 |\Delta \PhysicalState(s/\Gain)| \le e^{-\frac{L_\GenerativeDrift}{\Gain} s} |\Delta \PhysicalState(0)| \qquad \forall s \le 0.
\end{equation}
Setting $\lambda = L_\GenerativeDrift / \Gain$, this implies:
\begin{equation*}
 e^{\lambda s} \sum_{i=0}^\ell \Gain^i |\Delta \obsLatent_i(s)| \le L_H |\Delta \PhysicalState(0)| \le \frac{L_H}{c_{obs}} \sum_{i=0}^\ell \Gain^i |\Delta \obsLatent_i(0)| \qquad \forall s \le 0.
\end{equation*}
Multiplying by $\Gain^{-\ell}$ and substituting $\rho_{fm} = \bar{\obsLatent}' / \Gain$, we satisfy \eqref{eq:fading_memory_assumption} with $K = \frac{L_H}{c_{obs}} \max\{(\bar{\obsLatent}')^\ell, (\bar{\obsLatent}')^{-\ell}\}$. All conditions for Proposition \ref{prop:derivative_reconstruction} are met provided $\Gain$ is chosen large enough to enforce \eqref{eq:rho_fm_threshold}.

\paragraph{2. Combination of Observability and Asymptotic Bounds.}
Starting from the scaled observability inequality \eqref{eq:obs_scaled} and pulling out $\Gain^{-\ell}$:
\begin{align}
 \frac{\min\{1, (\bar{\obsLatent}')^\ell\}}{\Gain^\ell} c_{obs} |\Delta \PhysicalState(0)| &\le \frac{1}{\Gain^\ell} \sum_{i=0}^\ell \Gain^i (\bar{\obsLatent}')^{\ell-i} |\Delta \obsLatent_i(0)| \nonumber \\
 &= \rho_{fm}^\ell \sum_{i=0}^\ell \rho_{fm}^{-i} |\Delta \obsLatent_i(0)| = \sum_{i=0}^\ell \rho_{fm}^{\ell-i} |\Delta \obsLatent_i(0)|. \label{eq:final_chain_start}
\end{align}
Applying Proposition \ref{prop:derivative_reconstruction}, and recalling (with \eqref{eq_DefEncoder}) that $\Delta \varpi(0) = \Encoder_\Gain(\PhysicalState_{a0}) - \Encoder_\Gain(\PhysicalState_{b0})$, we obtain:
\begin{equation}
 \sum_{i=0}^\ell \rho_{fm}^{\ell-i} |\Delta \obsLatent_i(0)| \le 2\sqrt{\ell+1} L_{inj} \left[ |\Encoder_\Gain(\PhysicalState_{a0}) - \Encoder_\Gain(\PhysicalState_{b0})| + R_{rem} \right],
\end{equation}
where $R_{rem} = L_{e\obsLatent_{\ell+1}} \sup_{s \le 0} e^{\lambda s} |\Delta \obsLatent_{\ell+1}(s)|$.
Using the Lipschitz constant of the $(\ell+1)$-th Lie derivative and (\ref{LPN1}):
\begin{align*}
 e^{\lambda s} |\Delta \obsLatent_{\ell+1}(s)| &= e^{\frac{L_\GenerativeDrift}{\Gain}s} \frac{1}{\Gain^{\ell+1}} |\Lie^{\ell+1} \GenerativeObs(\flowf(\PhysicalState_{a0}, s/\Gain)) - \Lie^{\ell+1} \GenerativeObs(\flowf(\PhysicalState_{b0}, s/\Gain))| \\
 &\le \frac{\text{Lip}(\Lie^{\ell+1} \GenerativeObs)}{\Gain^{\ell+1}} e^{\frac{L_\GenerativeDrift}{\Gain}s} |\Delta \PhysicalState(s/\Gain)| \\
 &\le \frac{\text{Lip}(\Lie^{\ell+1} \GenerativeObs)}{\Gain^{\ell+1}} |\Delta \PhysicalState(0)|.
\end{align*}
Substituting this bound into $R_{rem}$ and combining with \eqref{eq:final_chain_start} yields:
\begin{align*}
 \frac{\min\{1, (\bar{\obsLatent}')^\ell\} c_{obs}}{\Gain^\ell} |\Delta \PhysicalState(0)| &\le 2\sqrt{\ell+1} L_{inj} |\Encoder_\Gain(\PhysicalState_{a0}) - \Encoder_\Gain(\PhysicalState_{b0})| \\&+ 2\sqrt{\ell+1} L_{inj} L_{e\obsLatent_{\ell+1}} \frac{\text{Lip}(\Lie^{\ell+1} \GenerativeObs)}{\Gain^{\ell+1}} |\Delta \PhysicalState(0)|.
\end{align*}
Multiplying the entire equation by $\Gain^\ell$ and grouping the terms proportional to $|\Delta \PhysicalState(0)|$:
\begin{align*}
 \Bigg( \min\{1, (\bar{\obsLatent}')^\ell\} c_{obs} &- \frac{2\sqrt{\ell+1} L_{inj} L_{e\obsLatent_{\ell+1}} \text{Lip}(\Lie^{\ell+1} \GenerativeObs)}{\Gain} \Bigg) |\Delta \PhysicalState(0)| \\&\le 2\sqrt{\ell+1} L_{inj} \Gain^\ell |\Encoder_\Gain(\PhysicalState_{a0}) - \Encoder_\Gain(\PhysicalState_{b0})|.
\end{align*}
If we select the minimal gain $\underline{\Gain}$ such that:
\begin{equation*}
 \underline{\Gain} > \frac{4\sqrt{\ell+1} L_{inj} L_{e\obsLatent_{\ell+1}} \text{Lip}(\Lie^{\ell+1} \GenerativeObs)}{\min\{1, (\bar{\obsLatent}')^\ell\} c_{obs}},
\end{equation*}
then the bracketed coefficient on the left is strictly positive and bounded below by $\frac{1}{2}\min\{1, (\bar{\obsLatent}')^\ell\} c_{obs}$. 
Dividing both sides by this strictly positive constant yields:
\begin{equation*}
 |\PhysicalState_{a0} - \PhysicalState_{b0}| =
 |\Delta \PhysicalState(0)| 
 \le c_T \Gain^\ell |\Encoder_\Gain(\PhysicalState_{a0}) - \Encoder_\Gain(\PhysicalState_{b0})|,
\end{equation*}
where $c_T = \frac{4\sqrt{\ell+1} L_{inj}}{\min\{1, (\bar{\obsLatent}')^\ell\} c_{obs}}$. Since $\Encoder_\Gain$ is the mapping constructed in Theorem \ref{thm:embedding}, this completely proves that it satisfies both the conclusions of Theorem \ref{thm:embedding} and the Lipschitz injectivity bound, concluding the proof of Theorem \ref{thm:injectivity}. \qed

\section{Decoder Approximation and Complexity (Theorems \ref{thm:ideal_tunability} and \ref{thm:neural_complexity})}
\label{app:proof_decoder}

This appendix provides the  proofs for the convergence of the ideal observer (Theorem \ref{thm:ideal_tunability}) and the neural decoder complexity bound (Theorem \ref{thm:neural_complexity}). A fundamental prerequisite for both proofs is to formally extend the exact inverse map from the embedded manifold to the entire latent operating region.

\subsection{Global Extension of the Inverse Map (McShane-Whitney)}
From Theorem \ref{thm:injectivity}, the map $\Encoder_\Gain: \X \to \LatentSpace$ is Lipschitz injective on $\X$ for $\Gain \ge \underline{\Gain}$, satisfying:
\begin{equation*}
 |\PhysicalState_a - \PhysicalState_b| \le c_T \Gain^\ell |\Encoder_\Gain(\PhysicalState_a) - \Encoder_\Gain(\PhysicalState_b)| \qquad \forall \PhysicalState_a, \PhysicalState_b \in \X.
\end{equation*}
This defines a strict left-inverse $\Encoder_\Gain^{\rm inv}: \Encoder_\Gain(\X) \to \X$. To ensure that our decoder can process any state within the latent operating region $\Zcal$, we must extend this inverse map to the entire space $\LatentSpace$ while preserving its Lipschitz continuity. 

We construct the extended map, denoted $\Encoder^{-1}_\Gain: \LatentSpace \to \PhysicalSpace$, component by component using the McShane-Whitney extension theorem. For each dimension $i \in \{1, \dots, n\}$, we define:
\begin{equation}
 \Encoder^{-1}_\Gain(\LatentState)_i = \sup_{\chi \in \X} \left\{ \chi_i - c_T \Gain^\ell |\LatentState - \Encoder_\Gain(\chi)| \right\}.
 \label{eq:mcshane_extension}
\end{equation}

\paragraph{Exact Inverse on $\X$:}
If $\LatentState$ is on the embedded manifold, i.e., $\LatentState = \Encoder_\Gain(\PhysicalState)$ for some $\PhysicalState \in \X$, then the injectivity bound directly implies:
\begin{equation*}
 \chi_i - \PhysicalState_i \le |\chi - \PhysicalState| \le c_T \Gain^\ell |\Encoder_\Gain(\chi) - \Encoder_\Gain(\PhysicalState)| \qquad \forall \chi \in \X.
\end{equation*}
Rearranging yields $\chi_i - c_T \Gain^\ell |\Encoder_\Gain(\PhysicalState) - \Encoder_\Gain(\chi)| \le \PhysicalState_i$ for all $\chi \in \X$. The supremum over $\chi$ in \eqref{eq:mcshane_extension} is therefore upper-bounded by $\PhysicalState_i$. However, taking $\chi = \PhysicalState$ exactly achieves this bound (since the distance term vanishes). Thus, $\Encoder^{-1}_\Gain(\Encoder_\Gain(\PhysicalState))_i = \PhysicalState_i$ for all $i$, proving that $\Encoder^{-1}_\Gain(\Encoder_\Gain(\PhysicalState)) = \PhysicalState$ on $\X$.

\paragraph{Global Lipschitz Continuity:}
Let $\LatentState_a, \LatentState_b \in \LatentSpace$ be two arbitrary points. Without loss of generality, assume $\Encoder^{-1}_\Gain(\LatentState_b)_i \ge \Encoder^{-1}_\Gain(\LatentState_a)_i$. Because $\X$ is compact, these suprema are finite, and we have:
\begin{align*}
 0 &\le \Encoder^{-1}_\Gain(\LatentState_b)_i - \Encoder^{-1}_\Gain(\LatentState_a)_i \\
 &= \sup_{\chi_b \in \X} \left\{ \chi_{bi} - c_T \Gain^\ell |\LatentState_b - \Encoder_\Gain(\chi_b)| \right\} - \sup_{\chi_a \in \X} \left\{ \chi_{ai} - c_T \Gain^\ell |\LatentState_a - \Encoder_\Gain(\chi_a)| \right\}.
\end{align*}
Since $\sup_{\chi_a} \{ \dots \}$ bounds the expression for any specific choice, we can substitute $\chi_a = \chi_b$ in the negative term to obtain an upper bound:
\begin{align*}
 \Encoder^{-1}_\Gain(\LatentState_b)_i - \Encoder^{-1}_\Gain(\LatentState_a)_i &\le \sup_{\chi_b \in \X} \left\{ \chi_{bi} - c_T \Gain^\ell |\LatentState_b - \Encoder_\Gain(\chi_b)| - \left( \chi_{bi} - c_T \Gain^\ell |\LatentState_a - \Encoder_\Gain(\chi_b)| \right) \right\} \\
 &\le c_T \Gain^\ell \sup_{\chi_b \in \X} \left\{ |\LatentState_a - \Encoder_\Gain(\chi_b)| - |\LatentState_b - \Encoder_\Gain(\chi_b)| \right\}.
\end{align*}
By the reverse triangle inequality, this evaluates to:
\begin{equation}
 |\Encoder^{-1}_\Gain(\LatentState_b)_i - \Encoder^{-1}_\Gain(\LatentState_a)_i| \le c_T \Gain^\ell |\LatentState_a - \LatentState_b| \qquad \forall i \in \{1, \dots, n\}.
\end{equation}
Summing the squares over all $n$ components yields the Euclidean Lipschitz bound for the vector map:
\begin{equation}
 |\Encoder^{-1}_\Gain(\LatentState_b) - \Encoder^{-1}_\Gain(\LatentState_a)| \le \sqrt{n} c_T \Gain^\ell |\LatentState_b - \LatentState_a| := L_{inv} |\LatentState_b - \LatentState_a|.
 \label{eq:extended_lipschitz}
\end{equation}

\subsection{Proof of Theorem \ref{thm:ideal_tunability} }
\label{sec_ProofIdealTunability}
Let $\LatentState(t) = \Phi_\LatentState(\LatentState_0, 0, \Gain t; \PhysicalState_0)$ be the latent trajectory starting at $\LatentState_0 \in \Zcal$ and driven by the physical observation sequence the initial condition of which $\PhysicalState_0$ is in $\X$ assumed to be forward invariant. The ideal estimate is defined as $\Estimate(t) = \Encoder^{-1}_\Gain(\LatentState(t))$. Since the physical trajectory $\PhysicalState(t)$ remains in $\X$ for all $t \ge 0$, we have the exact identity $\PhysicalState(t) = \Encoder^{-1}_\Gain(\Encoder_\Gain(\PhysicalState(t)))$. The estimation error is therefore:
\begin{equation*}
 |\Estimate(t) - \PhysicalState(t)| = |\Encoder^{-1}_\Gain(\LatentState(t)) - \Encoder^{-1}_\Gain(\Encoder_\Gain(\PhysicalState(t)))|.
\end{equation*}
Using the global Lipschitz property \eqref{eq:extended_lipschitz} of $\Encoder^{-1}_\Gain$:
\begin{equation*}
 |\Estimate(t) - \PhysicalState(t)| \le L_{inv} |\LatentState(t) - \Encoder_\Gain(\PhysicalState(t))|.
\end{equation*}
Applying the exponential attractivity bound from Theorem \ref{thm:embedding} yields:
\begin{equation*}
 |\Estimate(t) - \PhysicalState(t)| \le L_{inv} e^{-\Gain t} |\LatentState_0 - \Encoder_\Gain(\PhysicalState_0)| = \sqrt{n} c_T \Gain^\ell e^{-\Gain t} |\LatentState_0 - \Encoder_\Gain(\PhysicalState_0)|.
\end{equation*}
\startmodifVA Setting $c_T' = \sqrt{n} c_T$ gives the bound \eqref{eq:exponential_bound}. Since both $\LatentState_0$ and $\Encoder_\Gain(\PhysicalState_0)$ belong to the compact set $\Zcal$ (a ball of radius $R_z$), their distance is bounded by $2R_z$, so that\stopmodif
\begin{equation}
\startmodifVA  |\Estimate(t) - \PhysicalState(t)| \le  2R_z c_T' \Gain^\ell e^{-\Gain t}.\stopmodif
\end{equation}
For any finite $t_o > 0$ and any $\epsilon > 0$, the exponential term $e^{-\Gain t_o}$ dominates the polynomial term $\Gain^\ell$ for sufficiently large $\Gain$. Thus, we can tune $\Gain$ to ensure $|\Estimate(t) - \PhysicalState(t)| \le \epsilon$ for all $t \ge t_o$, proving tunability.

\subsection{Proof of Theorem \ref{thm:neural_complexity} (MLP Decoder Complexity)}
\label{sec_ProofMLPComplexity}
In practice, the theoretical decoder $\Encoder^{-1}_\Gain$ is replaced by an MLP $\Decoder_\Params(\LatentState)$. We decompose the overall estimation error using the triangle inequality:
\begin{equation*}
 |\Decoder_\Params(\LatentState(t)) - \PhysicalState(t)| \le \underbrace{|\Decoder_\Params(\LatentState(t)) - \Encoder^{-1}_\Gain(\LatentState(t))|}_{e_{approx}(t)} + \underbrace{|\Encoder^{-1}_\Gain(\LatentState(t)) - \PhysicalState(t)|}_{e_{obs}(t)}.
\end{equation*}
We must bound the sum by $\epsilon$ for $t \ge t_o$.
First, using Theorem \ref{thm:ideal_tunability}, we choose a structural gain $\Gain^*$ large enough so that the dynamic observer error satisfies:
\begin{equation*}
 e_{obs}(t) \le 2R_z c_T' (\Gain^*)^\ell e^{-\Gain^* t_o} \le \frac{\epsilon}{2} \qquad \forall t \ge t_o.
\end{equation*}

With $\Gain^*$ fixed, we must construct a neural network $\Decoder_\Params$ that approximates the Lipschitz function $\Encoder^{-1}_\Gain$ on the compact set $\Zcal$ with a uniform error bounded by $\epsilon/2$ in the Euclidean norm.

We invoke the quantitative approximation results from Yarotsky \cite[Theorem 1]{Yarotsky2017}, which states that any $1$-Lipschitz scalar function $f: [0,1]^m \to \RR$ can be approximated with precision $\delta \in (0,1)$ using a ReLU network containing $\mathcal{W}_{scalar}$ weights, satisfying:
\begin{equation*}
 \mathcal{W}_{scalar} \le c_m \left( \delta^{-m} \log(1/\delta) + 1 \right),
\end{equation*}
where $c_m$ is a constant depending only on the input dimension $m$.

Since our extended inverse map $\Encoder^{-1}_\Gain$ outputs an $n$-dimensional vector, bounding the Euclidean error by $\epsilon/2$ is guaranteed if each of the $n$ scalar coordinate functions is approximated with a  precision of $\epsilon/(2\sqrt{n})$. We approximate these $n$ coordinate functions by constructing $n$ parallel scalar networks (or equivalently, a single multi-head network), bounding the total weight complexity by $\mathcal{W} \le n \mathcal{W}_{scalar}$.

To apply Yarotsky's theorem to each coordinate function, we perform a spatial and an amplitude scaling to map their domains to the unit hypercube and normalize their Lipschitz constants to $1$:
\begin{itemize}
 \item \textbf{Spatial Scaling:} $\Zcal$ is contained in the hypercube $[-R_z, R_z]^m$. We map the standard hypercube $[0,1]^m$ to this domain via the affine transformation $x \mapsto 2R_z x - R_z \mathbf{1}_m$. Composing the coordinate functions with this mapping scales their effective Lipschitz constants with respect to $x$ by a factor of $2R_z$.
 \item \textbf{Amplitude Scaling:} Each resulting composed coordinate function has a Lipschitz constant bounded by $2 R_z L_{inv} = 2 R_z c_T' (\Gain^*)^\ell$. To normalize them into $1$-Lipschitz functions as required by the theorem, we must divide their outputs by this exact factor.
\end{itemize}

Consequently, achieving an absolute coordinate error of $\epsilon/(2\sqrt{n})$ for the original functions is mathematically equivalent to achieving a precision $\delta$ for the fully normalized functions, where $\delta$ is scaled by the same Lipschitz factor:
\begin{equation*}
 \delta = \frac{\epsilon / (2\sqrt{n})}{2 R_z c_T' (\Gain^*)^\ell} = \frac{\epsilon}{4\sqrt{n} R_z c_T' (\Gain^*)^\ell}.
\end{equation*}
Substituting this specific required precision $\delta$ into Yarotsky's complexity bound, and multiplying by $n$ for the total network, yields:
\begin{equation*}
 \mathcal{W} \le n c_m \left( \left( \frac{4\sqrt{n} R_z c_T' (\Gain^*)^\ell}{\epsilon} \right)^m \log\left(\frac{4\sqrt{n} R_z c_T' (\Gain^*)^\ell}{\epsilon}\right) + 1 \right).
\end{equation*}
By grouping all the constants that are independent of the tunable gain $\Gain^*$ and the precision $\epsilon$ (namely $n, 4, R_z, c_T', c_m$) into a single global constant $C_{net}$, we obtain the simplified asymptotic complexity bound:
\begin{equation*}
 \mathcal{W} \le C_{net} \left( \frac{(\Gain^*)^\ell}{\epsilon} \right)^m \log\left(\frac{(\Gain^*)^\ell}{\epsilon}\right).
\end{equation*}
Thus, there exists a network architecture $\Decoder_\Params$ with this bounded number of weights satisfying $e_{approx}(t) \le \epsilon/2$. The sum of the dynamic error and the approximation error guarantees the global precision $\epsilon$, concluding the proof.

\startmodifVA \subsection{Proof of Proposition \ref{prop:iss} (Robustness to Measurement Disturbances)}
\label{sec_ProofISS}

Throughout the proof, we fix $\Gain \ge \underline{\Gain}$, a physical initial condition $\PhysicalState_0 \in \X$ and a latent initial condition $\LatentState_0 \in \Zcal$. We denote by $\PhysicalState(t) = \flowf(\PhysicalState_0, t)$ the physical trajectory, which remains in $\X$ for all $t\ge 0$ by Assumption \ref{ass:regularity}, and by $\obsPhys(t) = \GenerativeObs(\PhysicalState(t)) \in \Ycal$ the nominal observation. The disturbance $\delta : [0,+\infty) \to \ObsSpace$ is Lebesgue measurable (it is unknown and need not be continuous), with $|\delta(t)| \le \|\delta\|_\infty \le \bar d$ for almost all $t \ge 0$: $\|\delta\|_\infty$ is the amplitude of the actual disturbance, whereas $\bar d$ is the admissible threshold defined in Proposition \ref{prop:iss}. The quantity $d(t)$ is its fading-memory version defined in Proposition \ref{prop:iss}. We compare two latent trajectories starting from the same initial condition $\LatentState_0 \in \Zcal$:
\begin{itemize}
 \item the \textit{nominal} latent trajectory $\LatentState(t)$, solution of $\frac{d\LatentState}{dt} = \Gain \BaseDynamics_\Params(\LatentState, \obsPhys(t))$, which is the one studied in Theorems \ref{thm:embedding} to \ref{thm:ideal_tunability};
 \item the \textit{perturbed} latent trajectory $\LatentState_d(t)$, solution of
 \begin{equation}\label{eq:perturbed_observer}
 \frac{d\LatentState_d}{dt} = \Gain \BaseDynamics_\Params(\LatentState_d, \obsPhys(t) + \delta(t)), \qquad \LatentState_d(0) = \LatentState_0 ,
 \end{equation}
 which is the one actually implemented, and from which the estimate $\Estimate(t) = \Decoder_\Params(\LatentState_d(t))$ is computed.
\end{itemize}

\paragraph{Preliminaries.}
Since $\delta$ is only measurable, the right-hand side of \eqref{eq:perturbed_observer} is measurable in $t$, and it is continuously differentiable in $\LatentState_d$ by Assumption \ref{ass:regularity}. Its solutions are therefore understood in the sense of Carath\'eodory, namely absolutely continuous functions satisfying \eqref{eq:perturbed_observer} for almost all $t$. As the considered input $\obsPhys(t) + \delta(t)$ takes its values in the compact set
\begin{equation*}
\lb{\Ycal^{d_0}} = \{\Observation \in \ObsSpace : \mathrm{dist}(\Observation, \Ycal) \le \lb{d_0}\},
\end{equation*}
the right-hand side of \eqref{eq:perturbed_observer} is bounded and Lipschitz in $\LatentState_d$ on any compact set, uniformly in $t$. By Carath\'eodory's theorem (see e.g. \cite[Chapter I.5]{Hale1980}), \eqref{eq:perturbed_observer} admits a unique solution defined on a maximal interval $[0, t_{\max})$, and $t_{\max} = +\infty$ if this solution remains in a compact set.

The set $\lb{\Ycal^{d_0}}$ is compact since $\Ycal$ is, so that the constant $\bar L_{go}$ defined in Proposition \ref{prop:iss} is finite as the maximum of a continuous function on the compact set $\Zcal \times \lb{\Ycal^{d_0}}$. Since $\Ycal \subset \lb{\Ycal^{d_0}}$, it is larger than or equal to the constant $L_{go}$ of \eqref{bound-for-g}. If $\bar L_{go} = 0$, we use the convention $\bar d = \lb{d_0}$.
The corrupted observation then takes its values in $\Ycal^{\bar d} = \{\Observation \in \ObsSpace : \mathrm{dist}(\Observation, \Ycal) \le \bar d\} \subset \lb{\Ycal^{d_0}}$. Recall also, from Appendix \ref{app:proof_invariance}, the notation $\bar{\BaseDynamics} = \max_{\Observation \in \Ycal} |\BaseDynamics_\Params(0, \Observation)|$, and that the radius of $\Zcal$ satisfies $R_z \ge 2 \bar{\BaseDynamics}$.

We then establish two useful inequalities: for all $\LatentState \in \Zcal$, all $\Observation \in \Ycal$ and all $\delta_0 \in \ObsSpace$ with $|\delta_0| \le \bar d$,
\begin{align}
 |\BaseDynamics_\Params(\LatentState, \Observation + \delta_0) - \BaseDynamics_\Params(\LatentState, \Observation)| &\le \bar L_{go} |\delta_0| , \label{eq:iss_ineq_a}\\
 \LatentState^\top \BaseDynamics_\Params(\LatentState, \Observation + \delta_0) &\le |\LatentState| \left( \bar{\BaseDynamics} + \bar L_{go} |\delta_0| - |\LatentState| \right) . \label{eq:iss_ineq_b}
\end{align}
To establish \eqref{eq:iss_ineq_a}, note that for all $s \in [0,1]$, $\mathrm{dist}(\Observation + s\delta_0, \Ycal) \le s|\delta_0| \le \bar d \le \lb{d_0}$, so that the segment $[\Observation, \Observation+\delta_0]$ is contained in $\lb{\Ycal^{d_0}}$. Hence, by the integral mean value theorem,
\begin{equation*}
 |\BaseDynamics_\Params(\LatentState, \Observation + \delta_0) - \BaseDynamics_\Params(\LatentState, \Observation)| = \left| \int_0^1 \frac{\partial \BaseDynamics_\Params}{\partial \Observation}(\LatentState, \Observation + s\delta_0)\, \delta_0 \, ds \right| \le \bar L_{go} |\delta_0| .
\end{equation*}
To establish \eqref{eq:iss_ineq_b}, we decompose
\begin{equation*}
 \LatentState^\top \BaseDynamics_\Params(\LatentState, \Observation + \delta_0) = \LatentState^\top \left[ \BaseDynamics_\Params(\LatentState, \Observation) - \BaseDynamics_\Params(0, \Observation) \right] + \LatentState^\top \BaseDynamics_\Params(0, \Observation) + \LatentState^\top \left[ \BaseDynamics_\Params(\LatentState, \Observation + \delta_0) - \BaseDynamics_\Params(\LatentState, \Observation) \right] .
\end{equation*}
Since $\Observation \in \Ycal$ and the segment $[0, \LatentState]$ is contained in the ball $\Zcal$, the first term is bounded by $-|\LatentState|^2$ exactly as in Appendix \ref{app:proof_invariance} (integral mean value theorem and Assumption \ref{ass:contraction}). The second term is bounded by $|\LatentState|\bar{\BaseDynamics}$ by the Cauchy-Schwarz inequality, and the third one by $|\LatentState| \bar L_{go}|\delta_0|$ by the Cauchy-Schwarz inequality and \eqref{eq:iss_ineq_a}. This gives \eqref{eq:iss_ineq_b}.

\paragraph{Step 1: forward invariance of $\Zcal$ for the perturbed latent dynamics.}
Recall that $\Ycal^{\bar d}$ contains the values of $\obsPhys(t) + \delta(t)$ for almost all $t$. Since $\Ycal$ is compact, any $\Observation' \in \Ycal^{\bar d}$ can be written $\Observation' = \Observation + \delta_0$ with $\Observation \in \Ycal$ and $|\delta_0| = \mathrm{dist}(\Observation', \Ycal) \le \bar d$. Applying \eqref{eq:iss_ineq_b} on the boundary $\partial \Zcal = \{|\LatentState| = R_z\}$, and using $\bar{\BaseDynamics} \le R_z/2$ and $\bar L_{go} \bar d \le R_z / 4$, we obtain
\begin{equation}\label{eq:iss_boundary}
 \LatentState^\top \BaseDynamics_\Params(\LatentState, \Observation') \le R_z \left( \frac{R_z}{2} + \frac{R_z}{4} - R_z \right) = -\frac{R_z^2}{4} , \qquad \forall \LatentState \in \partial \Zcal, \ \forall \Observation' \in \Ycal^{\bar d} .
\end{equation}
Since \eqref{eq:iss_boundary} holds for all $\Observation'$ in the compact set $\Ycal^{\bar d}$, in which the input $\obsPhys(t) + \delta(t)$ takes its values for almost all $t$, Lemma \ref{lemma-invariance-ball}, applied with $G = \Gain \BaseDynamics_\Params$, $\mathcal{U} = \Ycal^{\bar d}$, $R = R_z$ and $u(t) = \obsPhys(t) + \delta(t)$, shows that $\LatentState_d$ is defined on $[0, +\infty)$ and remains in $\Zcal$. This proves that $\Zcal$ is forward invariant for the perturbed latent dynamics \eqref{eq:perturbed_observer}, which is the first claim of Proposition \ref{prop:iss}.

\paragraph{Step 2: the perturbed and nominal latent trajectories remain close.}
By Proposition \ref{prop:invariance} and Step 1, $\LatentState(t)$ and $\LatentState_d(t)$ belong to $\Zcal$ for all $t \ge 0$. Let $\Delta(t) = \LatentState_d(t) - \LatentState(t)$, which is absolutely continuous and satisfies $\Delta(0) = 0$. For almost all $t \ge 0$, we have
\begin{equation*}
 \frac{d\Delta}{dt}(t) = \Gain \left[ \BaseDynamics_\Params(\LatentState_d(t), \obsPhys(t)) - \BaseDynamics_\Params(\LatentState(t), \obsPhys(t)) \right] + \Gain \left[ \BaseDynamics_\Params(\LatentState_d(t), \obsPhys(t) + \delta(t)) - \BaseDynamics_\Params(\LatentState_d(t), \obsPhys(t)) \right] .
\end{equation*}
As in Appendix \ref{app:proof_invariance}, the integral mean value theorem allows to write the first term as $A(t)\Delta(t)$, with
\begin{equation*}
 A(t) = \Gain \int_0^1 \frac{\partial \BaseDynamics_\Params}{\partial \LatentState}\left(\LatentState(t) + s\Delta(t), \obsPhys(t)\right) ds .
\end{equation*}
Since $\Zcal$ is convex, $\LatentState(t) + s\Delta(t) \in \Zcal$ for all $s \in [0,1]$, and since $\obsPhys(t) \in \Ycal$, Assumption \ref{ass:contraction} gives $A(t) + A(t)^\top \le -2\Gain I_m$.
We denote the second term by $B(t)$. Since $\LatentState_d(t) \in \Zcal$, $\obsPhys(t) \in \Ycal$ and $|\delta(t)| \le \bar d$, inequality \eqref{eq:iss_ineq_a} gives, for almost all $t \ge 0$,
\begin{equation*}
 |B(t)| \le \Gain \bar L_{go} |\delta(t)| .
\end{equation*}
Hence $\Delta$ is absolutely continuous and satisfies $\frac{d\Delta}{dt} = A(t)\Delta + B(t)$ for almost all $t$, with $A$ continuous, $B$ measurable and essentially bounded, $A(t) + A(t)^\top \le -2\Gain I_m$ and $\Delta(0) = 0$. Lemma \ref{lemma-dif-ineg}, with $a_{\!_+} = -\Gain$, gives
\begin{equation*}
 |\Delta(t)| \le \int_0^t e^{-\Gain(t-s)} |B(s)|\, ds \le \Gain \bar L_{go} \int_0^t e^{-\Gain(t-s)} |\delta(s)|\, ds .
\end{equation*}
By definition of $d(t)$, $e^{-\Gain(t-s)/2}|\delta(s)| \le d(t)$ for almost all $s \in [0, t]$, so that $e^{-\Gain(t-s)}|\delta(s)| \le e^{-\Gain(t-s)/2} d(t)$, and
\begin{equation}\label{eq:iss_latent}
 |\LatentState_d(t) - \LatentState(t)| \le \Gain \bar L_{go}\, d(t) \int_0^t e^{-\Gain(t-s)/2}\, ds \le 2 \bar L_{go}\, d(t) , \qquad \forall t \ge 0 .
\end{equation}
More generally, any rate $\lambda \in (0, \Gain)$ in place of $\Gain/2$ in the definition of $d(t)$ gives the same bound with the factor $2$ replaced by $\Gain/(\Gain - \lambda)$. Note that \eqref{eq:iss_latent} does not depend on the initial condition $\LatentState_0$, and that $d(t) \le \|\delta\|_\infty$.

\paragraph{Step 3: from the latent error to the estimation error.}
We first bound the distance between the perturbed latent state and the image of the physical state by $\Encoder_\Gain$. By the triangle inequality,
\begin{equation*}
 |\LatentState_d(t) - \Encoder_\Gain(\PhysicalState(t))| \le |\LatentState_d(t) - \LatentState(t)| + |\LatentState(t) - \Encoder_\Gain(\PhysicalState(t))| .
\end{equation*}
The first term is bounded by \eqref{eq:iss_latent}. Since $\LatentState(t)$ is the nominal latent trajectory starting from $\LatentState_0 \in \Zcal$, the attractivity property of Theorem \ref{thm:embedding}, which holds for $\Gain \ge \underline{\Gain}$ by Theorem \ref{thm:injectivity}, bounds the second term by $e^{-\Gain t} |\LatentState_0 - \Encoder_\Gain(\PhysicalState_0)|$. Hence
\begin{equation}\label{eq:iss_latent_total}
 |\LatentState_d(t) - \Encoder_\Gain(\PhysicalState(t))| \le e^{-\Gain t} |\LatentState_0 - \Encoder_\Gain(\PhysicalState_0)| + 2\bar L_{go}\, d(t) , \qquad \forall t \ge 0 .
\end{equation}
We now go back to the physical state. Since $\Gain \ge \underline{\Gain}$, Theorem \ref{thm:injectivity} applies, and the extended inverse $\Encoder^{-1}_\Gain : \LatentSpace \to \PhysicalSpace$ constructed in Appendix \ref{app:proof_decoder} is defined on the whole latent space and satisfies the global Lipschitz property \eqref{eq:extended_lipschitz} with constant $L_{inv} = \sqrt{n}\, c_T \Gain^\ell = c_T' \Gain^\ell$. Moreover, it is an exact left inverse on $\X$, namely $\Encoder^{-1}_\Gain(\Encoder_\Gain(\PhysicalState)) = \PhysicalState$ for all $\PhysicalState \in \X$. Since $\LatentState_d(t) \in \Zcal$ by Step 1, the decoder satisfies $|\Decoder_\Params(\LatentState_d(t)) - \Encoder^{-1}_\Gain(\LatentState_d(t))| \le \epsilon_{dec}$; this is where Step 1 is essential, the decoder being only guaranteed to be accurate on $\Zcal$. Since $\PhysicalState(t) \in \X$ for all $t \ge 0$, the triangle inequality gives
\begin{multline*}
 |\Estimate(t) - \PhysicalState(t)| \le |\Decoder_\Params(\LatentState_d(t)) - \Encoder^{-1}_\Gain(\LatentState_d(t))| + |\Encoder^{-1}_\Gain(\LatentState_d(t)) - \Encoder^{-1}_\Gain(\Encoder_\Gain(\PhysicalState(t)))| \\ \le \epsilon_{dec} + c_T' \Gain^\ell\, |\LatentState_d(t) - \Encoder_\Gain(\PhysicalState(t))| .
\end{multline*}
Combining with \eqref{eq:iss_latent_total} gives \eqref{eq:iss_bound}, which concludes the proof of Proposition \ref{prop:iss}.

Finally, the decoder constructed in the proof of Theorem \ref{thm:neural_complexity} satisfies $\epsilon_{dec} \le \epsilon/2$, with a gain such that $2R_z c_T' \Gain^\ell e^{-\Gain t_o} \le \epsilon/2$. Since $|\LatentState_0 - \Encoder_\Gain(\PhysicalState_0)| \le 2R_z$, \eqref{eq:iss_bound} then gives $|\Decoder_\Params(\LatentState_d(t)) - \PhysicalState(t)| \le \epsilon + 2c_T'\Gain^\ell \bar L_{go}\, d(t)$ for all $t \ge t_o$, as stated after Proposition \ref{prop:iss}.\stopmodif

\section{Discrete-Time Implementation and Euler Error (Theorem \ref{thm:discrete_tunability})}
\label{app:proof_euler}

In this section, we explicitly extend our continuous-time analysis to the discrete-time residual RNN implementation and prove Theorem \ref{thm:discrete_tunability}.

\subsection{RNN Structure as a Residual Network}
The theoretical continuous-time observer is defined by $\dot{\LatentState} = \Gain \BaseDynamics_\Params(\LatentState, \Observation)$.
Its discrete implementation via an explicit Euler scheme is given by:
\begin{equation}
 \LatentState_{k+1} = \LatentState_k + \TimeStep \Gain \BaseDynamics_\Params(\LatentState_k, \obsPhys_k) = \text{RNN}_{\Params}(\LatentState_k, \obsPhys_k).
 \label{eq:euler_scheme_app}
\end{equation}
This equation corresponds exactly to the structure of a Residual RNN, by setting $\LatentState_k$ as the hidden state traversing the sequence layers.

\paragraph{Example:} Let $\sigma(\cdot) = \tanh(\cdot)$ be the activation function. Since $|\tanh(v)| \le 1$ element-wise, the nonlinearity is bounded by $\sqrt{m}$ in Euclidean norm.
We explicitly define the compact operating set $\Zcal$ as the closed ball of radius $R_z > \sqrt{m}$:
\begin{equation*}
 \Zcal = \{ \LatentState \in \LatentSpace \mid |\LatentState| \le R_z \}.
\end{equation*}
Consider the base dynamics:
\begin{equation*}
 \BaseDynamics_\Params(\LatentState, \Observation) = -\LatentState + \tanh(W \LatentState + U \Observation + b).
\end{equation*}
For any $\LatentState$ on the boundary of $\Zcal$ (i.e., $|\LatentState| = R_z$), the linear dissipative term $-\LatentState$ strictly dominates the bounded term $\tanh(\dots)$, ensuring that the continuous vector field points inward. Thus, $\Zcal$ is forward invariant.
The discrete update becomes:
\begin{equation*}
 \LatentState_{k+1} = (1 - \TimeStep \Gain) \LatentState_k + \TimeStep \Gain \tanh(W \LatentState_k + U \obsPhys_k + b).
\end{equation*}
The weights $\Params = (W, U, b)$ are then chosen to ensure the strict monotonicity condition
(see (\ref{eq:contraction_condition})) for the function $\LatentState \in \Zcal \mapsto \tanh(W \LatentState + U \obsPhys + b)$.

The physical state estimate is finally given by the MLP Decoder:
\begin{equation}
 \Estimate_k = \Decoder_{\Params}(\LatentState_k) \approx \Encoder^{-1}_\Gain(\LatentState_k).
\end{equation}

\subsection{Proof of Tunability of the Residual RNN}

To prove Theorem \ref{thm:discrete_tunability}, we decompose the total discrete estimation error into two components: the continuous observer tracking error (which decreases with $\Gain$) and the numerical discretization error (which increases with $\Gain$ but decreases with $\TimeStep$).

Let $\LatentState(t)$ be the exact solution of the continuous observer $\dot{\LatentState} = \Gain \BaseDynamics_\Params(\LatentState, \obsPhys(t))$ initialized at $\LatentState_0$.
The overall estimation error using the exact left-inverse map at discrete time $t_k = k\TimeStep$ is:
\begin{align}
 |\Estimate_k - \PhysicalState(t_k)| &= |\Encoder^{-1}_\Gain(\LatentState_k) - \Encoder^{-1}_\Gain(\Encoder_\Gain(\PhysicalState(t_k)))| \nonumber \\
 &\le c_T \Gain^\ell |\LatentState_k - \Encoder_\Gain(\PhysicalState(t_k))|,
 \label{eq:total_error_lip}
\end{align}
where we used the global Lipschitz property of the extended inverse $\Encoder^{-1}_\Gain$ (established in Appendix \ref{app:proof_decoder}).
We introduce the exact continuous solution $\LatentState(t_k)$ to separate the error via the triangle inequality:
\begin{equation}
 |\LatentState_k - \Encoder_\Gain(\PhysicalState(t_k))| \le \underbrace{|\LatentState_k - \LatentState(t_k)|}_{e_{num, k} \text{ (Numerical Error)}} + \underbrace{|\LatentState(t_k) - \Encoder_\Gain(\PhysicalState(t_k))|}_{e_{obs}(t_k) \text{ (Observer Error)}}.
 \label{eq:triangle_ineq}
\end{equation}

\subsubsection*{Step 1: Continuous Observer Error Bound}
According to Theorem \ref{thm:embedding}, the exact continuous observer error decays exponentially:
\begin{equation}
 e_{obs}(t_k) \le e^{-\Gain t_k} |\LatentState_0 - \Encoder_\Gain(\PhysicalState_0)| \le C_0 e^{-\Gain t_k},
 \label{eq:continuous_bound_app}
\end{equation}
where $C_0 = 2 R_z$ is the maximum distance in $\Zcal$.

\subsubsection*{Step 2: Numerical Error Bound (Euler Discretization)}
We analyze the propagation of the global numerical error defined as $e_{num, k} = \LatentState_k - \LatentState(t_k)$.

\begin{enumerate}[label=\alph*)]
 \item \textbf{Stability (Discrete-time Contraction).}
 Let $\BaseDynamics_\Params$ be contracting (satisfying Assumption \ref{ass:contraction}) and $L_{\LatentState}$-Lipschitz with respect to the latent state. Consider two discrete trajectories initialized at $\LatentState_{a,k}$ and $\LatentState_{b,k}$, driven by the same observation $\obsPhys_k$. Let $\Delta \LatentState_k = \LatentState_{a,k} - \LatentState_{b,k}$ and $\Delta \BaseDynamics_k = \BaseDynamics_\Params(\LatentState_{a,k}, \obsPhys_k) - \BaseDynamics_\Params(\LatentState_{b,k}, \obsPhys_k)$. The squared norm of their difference evolves as:
 \begin{align*}
  |\LatentState_{a, k+1} - \LatentState_{b, k+1}|^2 &= |(\LatentState_{a,k} - \LatentState_{b,k}) + \TimeStep\Gain(\BaseDynamics_\Params(\LatentState_{a,k}, \obsPhys_k) - \BaseDynamics_\Params(\LatentState_{b,k}, \obsPhys_k))|^2 \\
  &= |\Delta \LatentState_k|^2 + 2\TimeStep\Gain \langle \Delta \LatentState_k, \Delta \BaseDynamics_k \rangle + \TimeStep^2 \Gain^2 |\Delta \BaseDynamics_k|^2.
 \end{align*}
 Using the strong monotonicity property $\langle \Delta \LatentState_k, \Delta \BaseDynamics_k \rangle \le -|\Delta \LatentState_k|^2$ and the Lipschitz property $|\Delta \BaseDynamics_k| \le L_{\LatentState} |\Delta \LatentState_k|$:
 \begin{equation*}
  |\LatentState_{a, k+1} - \LatentState_{b, k+1}|^2 \le (1 - 2\TimeStep\Gain + \TimeStep^2 \Gain^2 L_{\LatentState}^2) |\Delta \LatentState_k|^2.
 \end{equation*}
 To ensure discrete-time stability, we enforce the condition $\TimeStep \Gain L_{\LatentState}^2 \le 1$. Under this constraint, the coefficient satisfies $1 - 2\TimeStep\Gain + \TimeStep^2 \Gain^2 L_{\LatentState}^2 \le (1 - \frac{1}{2}\TimeStep\Gain)^2$. Taking the square root yields the discrete contraction inequality:
 \begin{equation}
  |\Delta \LatentState_{k+1}| \le \rho |\Delta \LatentState_k|, \quad \text{where } \rho = 1 - \frac{\TimeStep\Gain}{2}.
  \label{eq:discrete_contraction}
 \end{equation}

\item \textbf{Consistency (Local Truncation Error).}
 We define the local truncation error $\xi_k$ as the defect of the Euler step with respect to the exact continuous solution:
 \begin{equation*}
  \LatentState(t_{k+1}) = \LatentState(t_k) + \TimeStep \Gain \BaseDynamics_\Params(\LatentState(t_k), \obsPhys(t_k)) + \xi_k.
 \end{equation*}
 Using a second-order Taylor expansion of the exact solution $\LatentState(t)$ at $t=t_k$:
 \begin{equation*}
  \LatentState(t_{k+1}) = \LatentState(t_k) + \TimeStep \dot{\LatentState}(t_k) + \frac{\TimeStep^2}{2} \ddot{\LatentState}(\tau_k) \quad \text{for some } \tau_k \in [t_k, t_{k+1}].
 \end{equation*}
 Identifying terms with $\dot{\LatentState}(t_k) = \Gain \BaseDynamics_\Params(\LatentState(t_k), \obsPhys(t_k))$, we obtain $|\xi_k| = \frac{\TimeStep^2}{2} |\ddot{\LatentState}(\tau_k)|$.
 Differentiation of the dynamics yields:
 \begin{equation*}
  \ddot{\LatentState}(t) = \frac{d}{dt} (\Gain \BaseDynamics_\Params(\LatentState, \obsPhys(t))) = \Gain^2 \frac{\partial \BaseDynamics_\Params}{\partial \LatentState} \BaseDynamics_\Params(\LatentState, \obsPhys(t)) + \Gain \frac{\partial \BaseDynamics_\Params}{\partial \Observation} \dot{\obsPhys}(t).
 \end{equation*}
 By Assumption \ref{ass:regularity}, the physical state $\PhysicalState(t)$ remains in the forward-invariant compact set $\X$. Consequently, both the physical vector field $\GenerativeDrift$ and the observation Jacobian $\frac{\partial \GenerativeObs}{\partial \PhysicalState}$ are uniformly bounded, which implies that the time derivative $\dot{\obsPhys}(t) = \frac{\partial \GenerativeObs}{\partial \PhysicalState}(\PhysicalState(t)) \GenerativeDrift(\PhysicalState(t))$ is globally bounded for all $t \ge 0$. Since $\BaseDynamics_\Params$ and its partial derivatives are also bounded on the compact operating domain $\Zcal \times \Ycal$, there exists a global constant $M > 0$ (independent of $t$ and $\TimeStep$) such that for any gain $\Gain \ge 1$:
 \begin{equation*}
  |\ddot{\LatentState}(t)| \le M \Gain^2.
 \end{equation*}
 Consequently, the local error is uniformly bounded by:
 \begin{equation*}
  |\xi_k| \le \frac{M}{2} \Gain^2 \TimeStep^2.
 \end{equation*}

 \item \textbf{Global Bound (Finite Time).}
 The global numerical error $e_{num, k+1}$ satisfies the recurrence:
 \begin{align*}
  e_{num, k+1} &= \LatentState_{k+1} - \LatentState(t_{k+1}) \\
  &= \left[ \LatentState_k + \TimeStep \Gain \BaseDynamics_\Params(\LatentState_k, \obsPhys_k) \right] - \left[ \LatentState(t_k) + \TimeStep \Gain \BaseDynamics_\Params(\LatentState(t_k), \obsPhys_k) + \xi_k \right].
 \end{align*}
 Using the contraction property derived in step a):
 \begin{equation*}
  |e_{num, k+1}| \le \rho |e_{num, k}| + |\xi_k|.
 \end{equation*}
 Assuming an exact initial condition ($e_{num, 0} = 0$), expanding the recurrence gives:
 \begin{equation*}
  |e_{num, k}| \le \sum_{j=0}^{k-1} \rho^{k-1-j} |\xi_j| \le \frac{M}{2} \Gain^2 \TimeStep^2 \sum_{j=0}^{k-1} \rho^j.
 \end{equation*}
 Since $\rho < 1$, we can upper-bound the finite sum by the geometric series:
 \begin{equation*}
  \sum_{j=0}^{k-1} \rho^j < \sum_{j=0}^{\infty} \rho^j = \frac{1}{1-\rho} = \frac{1}{\TimeStep\Gain / 2}.
 \end{equation*}
 Substituting this back into the error bound yields a result valid for all $k \ge 0$:
 \begin{equation}
  |e_{num, k}| \le \left( \frac{M}{2} \Gain^2 \TimeStep^2 \right) \left( \frac{2}{\TimeStep\Gain} \right) = M \Gain \TimeStep.
  \label{eq:discrete_bound_app}
 \end{equation}
\end{enumerate}

\subsubsection*{Step 3: Synthesis of Tunability}
Combining the Lipschitz bound \eqref{eq:total_error_lip}, the continuous error \eqref{eq:continuous_bound_app}, and the discrete numerical error \eqref{eq:discrete_bound_app}, the total estimation error at step $k$ is strictly bounded by:
\begin{equation*}
 |\Estimate_k - \PhysicalState(t_k)| \le c_T \Gain^\ell \left( \underbrace{C_0 e^{-\Gain t_k}}_{\text{Transient Error}} + \underbrace{M \Gain \TimeStep}_{\text{Numerical Error}} \right).
\end{equation*}

Let us formally prove tunability for a fixed target precision $\epsilon > 0$ and observation time $t_o > 0$:
\begin{itemize}
 \item \textbf{Choice of the continuous gain $\Gain$:} We choose a structural gain $\Gain \ge \underline{\Gain}$ large enough such that, for all $t_k \ge t_o$, the continuous transient error is sufficiently damped:
 \begin{equation*}
  c_T C_0 \Gain^\ell e^{-\Gain t_o} \le \frac{\epsilon}{2}.
 \end{equation*}
 This is always achievable because the exponential decay $e^{-\Gain t_o}$ dominates the polynomial growth $\Gain^\ell$ as $\Gain \to \infty$. We now fix this chosen gain $\Gain$.

 \item \textbf{Choice of the step size $\TimeStep$:} With $\Gain$ fixed, we choose a maximum sampling step $\TimeStep^*$ small enough to simultaneously satisfy the numerical stability condition ($\TimeStep^* \le \frac{1}{\Gain L_{\LatentState}^2}$) and strictly suppress the numerical truncation error:
 \begin{equation*}
  c_T \Gain^\ell (M \Gain \TimeStep^*) \le \frac{\epsilon}{2} \implies \TimeStep^* \le \frac{\epsilon}{2 M c_T \Gain^{\ell+1}}.
 \end{equation*}
\end{itemize}
For any $\TimeStep \le \TimeStep^*$, the sum of both errors is strictly bounded by $\epsilon$ for all $t_k \ge t_o$. Thus, there exists a configuration pair $(\Gain, \TimeStep)$ enabling the achievement of arbitrary precision. The Residual RNN architecture is therefore formally tunable, concluding the proof of Theorem \ref{thm:discrete_tunability}. \qed

\section{More details on the Robot pose estimation experiments (Section \ref{sec:experiments})}
\label{app:rookie}

\subsection{Supervision and losses}
\label{app:rookie_losses}
The standard per-step loss,
\begin{equation}
  \mathcal{L}_{\mathrm{step}} = \frac{1}{T}\sum_{t=1}^{T} \bigl\|\hat{y}_t - y_t\bigr\|^2,
\end{equation}
treats each timestep independently.
AMCL noise at step $t$ contributes gradient signal only at step $t$.
Consistent drift—the primary cause of trajectory-level error—is not amplified relative to per-step noise.

We propose a \textbf{trajectory integration loss} that directly penalises accumulated drift.
The chunk is divided into non-overlapping windows of length $H$.
Within each window, both the corrected and target body-frame displacements are integrated kinematically, and position MSE is computed on the resulting trajectories:
\begin{equation}
  \mathcal{L}_{\mathrm{traj}}(H) = \frac{1}{\lfloor T/H \rfloor}
    \sum_{k=0}^{\lfloor T/H \rfloor - 1}
    \mathrm{MSE}\!\Bigl(
      \mathrm{Integrate}\bigl(\hat{d}_{kH:kH+H}\bigr),\;
      \mathrm{Integrate}\bigl(d^{\mathrm{amcl}}_{kH:kH+H}\bigr)
    \Bigr)
\end{equation}
The integration is implemented with \texttt{cumsum} over the heading increments, making it fully differentiable without a Python loop.
The gradient at correction $\hat{y}_t$ therefore flows from all position errors at times $t, t+1, \ldots, kH+H-1$: consistent drift that compounds forward is amplified, while zero-mean AMCL noise at individual steps tends to cancel across future loss terms.

The window length $H$ controls the trade-off between signal and noise: short windows ($H \lesssim 20$) carry too little drift signal (AMCL and odometry agree at sub-metre scales); long windows ($H \gtrsim 100$) accumulate enough AMCL noise in the integrated target to overwhelm the gradient signal.
Note that $H=1$ recovers $\mathcal{L}_{\mathrm{step}}$ exactly.

\textbf{Training.}
Training uses sliding-window chunks of length 300 timesteps (stride 150).
The GRU hidden state is reset to zero at each chunk boundary. Chunks containing AMCL relocalization jumps (consecutive AMCL displacement $>0.5$\,m) are excluded. We use Adam with initial learning rate $4\times10^{-4}$, cosine annealing with minimum learning rate $4\times10^{-5}$ over up to 2000 epochs, batch size 1024, and gradient clipping at norm 1.0.

\textbf{Checkpoint selection.}
Validation loss (per-step MSE on held-out chunks) and trajectory RMSE are anti-correlated after the training optimum: a model that continues to minimise per-step loss is fitting AMCL noise at individual timesteps, producing erratic corrections that accumulate into larger trajectory errors.
We therefore select checkpoints by trajectory RMSE evaluated every 10 epochs, not by validation loss.

\subsection{Network architectures: ``Contracting GRU''}
\label{app:rookie_arch}
\label{sec:contracting_gru}

The standard GRU provides no guarantee on the long-term behaviour of the hidden
state: two copies of the network initialised at different hidden states may not
converge under the same input sequence.
Motivated by the KKL observer framework of this paper, we introduce a \textbf{``Contracting GRU''}, which replaces the GRU cell with a
provably contracting residual update.

\paragraph{Architecture.}
Let $z_k \in \RR^m$ denote the hidden state at step $k$ and $o_k \in \RR^9$ the input.
The update rule is
\begin{align}
  r_k &= \sigma(W_r z_k + U_r o_k + b_r), \label{eq:cgru_reset} \\
  n_k &= \tanh(W_n z_k + U_n o_k + b_n), \label{eq:cgru_cand} \\
  g_k &= -\alpha\, z_k + r_k \odot n_k,   \label{eq:cgru_field} \\
  z_{k+1} &= z_k + \delta\kappa\, g_k,    \label{eq:cgru_step}
\end{align}
where $r_k$ is a GRU-style reset gate, $n_k$ is a candidate state, and $g_k$
is the monotonic velocity field.
The scalar $\alpha > 0$ is a fixed dissipation coefficient;
$\delta$ and $\kappa$ are the sampling period and gain inherited from the KKL
framework.
The reset gate allows $g$ to depend on the current input in a richer way than
the plain KKL cell, while keeping the contraction analysis tractable.

\paragraph{Contraction guarantee.}
To apply the tools of this paper, we define the associated continuous-time latent dynamics given by
\begin{equation}
\label{eq:gOmega_cGRU}
 g_\Omega(z,o) = -\alpha\, z + r_\Omega(z,o) \odot n_\Omega(z,o)   
\end{equation}
where $\Omega$ stores the weights $W_r,W_n,U_r,U_n,b_r,b_n$ and
$$
r_\Omega(z,o) = \sigma(W_r z + U_r o + b_r) \quad , \quad n_\Omega(z,o) = \tanh(W_n z + U_n o + b_n).
$$
The Jacobian of $g_\Omega$ with respect to $z$ is
\[
  \frac{\partial g_\Omega}{\partial z}
  = -\alpha I
    + \mathrm{diag}(r_\Omega)\, D_n\, W_n
    + \mathrm{diag}(n_\Omega)\, \mathrm{diag}(r_\Omega \odot (1-r_\Omega))\, W_r,
\]
where $D_n = \mathrm{diag}(\mathrm{sech}^2(\cdot))$.
Using $\|D_n\|_{\mathrm{op}} \le 1$, $r_i(1-r_i) \le 1/4$, and $|n_i| \le 1$,
the symmetric part satisfies
\begin{equation}
  \frac{\partial g_\Omega}{\partial z} + \left(\frac{\partial g_\Omega}{\partial z}\right)^\top
  \;\le\;
  \bigl(-2\alpha + 2\|W_n\|_2 + \tfrac{1}{2}\|W_r\|_2\bigr)\, I.
  \label{eq:cgru_bound}
\end{equation}
We enforce $\|W_n\|_2 \le 1$ and $\|W_r\|_2 \le 1$ throughout training via
spectral normalisation.
Setting $\alpha = 2.5$ gives an upper bound of $-2.5\,I$ on the symmetric part,
so Assumption~\ref{ass:contraction} of the main paper is globally satisfied by construction, namely for any values of the latent state and inputs $(z,o)$.

\startmodifVA \paragraph{Empirical contraction constant.}
The bound \eqref{eq:cgru_bound} is an a priori certificate; we also measured the quantity our theory actually relies on. To avoid any confusion between conventions, we express everything in terms of the largest eigenvalue $\lambda_{\max}$ of the symmetric part $\frac12\bigl(\frac{\partial g_\Omega}{\partial z} + (\frac{\partial g_\Omega}{\partial z})^\top\bigr)$ of the latent Jacobian. In this convention, Assumption~\ref{ass:contraction} requires $\lambda_{\max} \le -1$, and the a priori bound \eqref{eq:cgru_bound} with $\alpha = 2.5$ gives $\lambda_{\max} \le -1.25$. Computed by automatic differentiation along all validation trajectories ($8$ trajectories, $26{,}888$ time steps, latent dimension $128$, observation dimension $9$), $\lambda_{\max}$ for the first layer of the trained Contracting GRU averages $-2.18$ (standard deviation $0.015$, worst case $-2.16$ over the whole dataset); the second, stacked layer behaves similarly (mean $-2.19$). The monotonicity condition therefore holds strictly at $100\%$ of the time steps, with a margin of about $0.9$ with respect to the a priori bound, and of about $1.2$ with respect to the requirement of Assumption~\ref{ass:contraction}. The discrete residual RNN inherits the corresponding guarantees through the small-step analysis of Appendix~\ref{app:proof_euler}: the step size $\delta\kappa = 1/3$ used in our experiments (Section~\ref{app:rookie_hyperparameters}) keeps the discrete update in the regime where this continuous-time monotonicity is preserved.\stopmodif

\paragraph{Stacking layers.}
A second contracting cell can be stacked on top of the first by using
$z^{(0)}_k$ as the input to layer 1 in place of $o_k$:
\begin{align}
  z^{(0)}_{k+1} &= z^{(0)}_k +  \delta\kappa\, g^{(0)}_k, \\
  z^{(1)}_{k+1} &= z^{(1)}_k + \delta\kappa\, g^{(1)}_k,
\end{align}
where $g^{(0)}$ uses $(z^{(0)}_k, o_k)$ and $g^{(1)}$ uses $(z^{(1)}_k, z^{(0)}_k)$. In fact, denoting $g_\Omega$ the single-layer latent base map given by \eqref{eq:gOmega_cGRU} and globally verifying Assumption~{\startmodifVA \ref{ass:contraction}\stopmodif}, we claim that the two-layer architecture corresponding to a map $g_\Omega^{\rm tot}$ built as 
$$
g_\Omega^{\rm tot}(z,o) = (g_{\Omega_0}(z^0,o),g_{\Omega_1}(z^1,z^0))
$$ 
where $z=(z^0,z^1)$ stacks the latent state of each cell, $\Omega=(\Omega_0,\Omega_1)$ stores the weights of each layer, still falls into the scope of this paper. This is done by operating a change of coordinates on the latent states and a scaling in $\kappa$. Indeed, we have
$$
\frac{\partial g_\Omega^{\rm tot}}{\partial z}(z,o)
=
\left(
\renewcommand{\arraystretch}{1.5}\begin{array}{cc}
\frac{\partial g_{\Omega_0}}{\partial z}(z^0,o)& 0
\\
\frac{\partial g_{\Omega_1}}{\partial o}(z^1,z^0)
&
\frac{\partial g_{\Omega_1}}{\partial z}(z^1,z^0)
\end{array}\right)
$$
which implies
$$
\frac{\partial g_\Omega^{\rm tot}}{\partial z}(z,o)+\frac{\partial g_\Omega^{\rm tot}}{\partial z}(z,o)^\top 
=
\left(
\renewcommand{\arraystretch}{1.5}\begin{array}{cc}
\frac{\partial g_{\Omega_0}}{\partial z}(z^0,o)
+
\frac{\partial g_{\Omega_0}}{\partial z}(z^0,o)^\top
& \frac{\partial g_{\Omega_1}}{\partial o}(z^1,z^0)^\top
\\
\frac{\partial g_{\Omega_1}}{\partial o}(z^1,z^0)
&
\frac{\partial g_{\Omega_1}}{\partial z}(z^1,z^0)
+
\frac{\partial g_{\Omega_1}}{\partial z}(z^1,z^0)^\top
\end{array}\right)
$$
From the global monotonicity of $g_\Omega$,  
$$
\frac{\partial g_{\Omega_0}}{\partial z}(z^0,o)
+
\frac{\partial g_{\Omega_0}}{\partial z}(z^0,o)^\top\leq -2 I
\quad ,\qquad 
\frac{\partial g_{\Omega_1}}{\partial z}(z^1,z^0)
+
\frac{\partial g_{\Omega_1}}{\partial z}(z^1,z^0)^\top\leq -2I
$$
which is not enough to imply the monotonicity of $g_\Omega^{\rm tot}$.  However, notice that $\frac{\partial g_\Omega}{\partial o}$ in (\ref{eq:gOmega_cGRU}) is bounded and consider $\mu\geq 1$ such that
$$
\mu \geq \max_{(z,o)} \left|\frac{\partial g_{\Omega_1}}{\partial o} (z,o) \right|.
$$
Then, defining $P=\begin{pmatrix}
    \mu^2 I_m & 0 \\
    0&I_m
\end{pmatrix}$ (with $I_m$ the identity matrix of size $m\times m$), we have
\begin{multline*}
     P\frac{\partial g_\Omega^{\rm tot}}{\partial z}(z,o)+\frac{\partial g_\Omega^{\rm tot}}{\partial z}(z,o)^\top P \\=
\left(
\begin{array}{cc}
\mu^2 \left[\frac{\partial g_{\Omega_0}}{\partial z}(z^0,o)
+
\frac{\partial g_{\Omega_0}}{\partial z}(z^0,o)^\top\right]
& \frac{\partial g_{\Omega_1}}{\partial o}(z^1,z^0)^\top 
\\
\frac{\partial g_{\Omega_1}}{\partial o}(z^1,z^0)
&
\frac{\partial g_{\Omega_1}}{\partial z}(z^1,z^0)
+
\frac{\partial g_{\Omega_1}}{\partial z}(z^1,z^0)^\top
\end{array}\right)
\end{multline*}
so that, for all $\xi = (\xi_0, \xi_1) \in {\startmodifVA \RR^{2m}\stopmodif}$,
\begin{align*}
 \begin{bmatrix}
    \xi_0\\\xi_1
\end{bmatrix}^\top
\Bigg[P\frac{\partial g_\Omega^{\rm tot}}{\partial z}(z,o)&+\frac{\partial g_\Omega^{\rm tot}}{\partial z}(z,o)^\top P \Bigg]
\begin{bmatrix}
    \xi_0\\\xi_1
\end{bmatrix}
\\
&\leq -2\mu^2 \|\xi_0\|^2 + 2 \left| \frac{\partial g_{\Omega_1}}{\partial o}(z^1,z^0)\right|\|\xi_0\|\|\xi_1\| -2 \|\xi_1\|^2   \\
&\leq -2\mu^2 \|\xi_0\|^2 + 2 \mu\|\xi_0\|\|\xi_1\| -2 \|\xi_1\|^2\\
&\leq -\mu^2\|\xi_0\|^2-\|\xi_1\|^2 \startmodifVA = -\xi^\top P \xi\stopmodif,
\end{align*}
hence
$$
P\frac{\partial g_\Omega^{\rm tot}}{\partial z}(z,o)+\frac{\partial g_\Omega^{\rm tot}}{\partial z}(z,o)^\top P \leq {\startmodifVA -P\stopmodif} \qquad \forall (z,o).
$$
\startmodifVA (the last equality follows from $P=\mathrm{diag}(\mu^2 I_m, I_m)$).\stopmodif Defining the new latent coordinates $\tilde{z}=P^{1/2}z$ and the new scaling parameter $\tilde \kappa = \frac{\kappa}{2}$, the latent continuous-time dynamics can be written as $\dot{\tilde z} = \tilde \kappa \tilde g(\tilde z,o)$ where
$$
\tilde g_\Omega^{\rm tot}(\tilde z,o) = 2 P^{1/2}g_\Omega^{\rm tot}(P^{-1/2} \tilde z,o)
$$
which verifies\startmodifVA, since $\frac{\partial \tilde g_\Omega^{\rm tot}}{\partial \tilde z}+\frac{\partial \tilde g_\Omega^{\rm tot}}{\partial \tilde z}^\top = 2P^{-1/2}\bigl(P\frac{\partial g_\Omega^{\rm tot}}{\partial z}+\frac{\partial g_\Omega^{\rm tot}}{\partial z}^\top P\bigr)P^{-1/2}$,\stopmodif
$$
\frac{\partial \tilde g_\Omega^{\rm tot}}{\partial \tilde z}(\tilde z,o)+\frac{\partial \tilde g_\Omega^{\rm tot}}{\partial \tilde z}(\tilde z,o)^\top  \leq - 2 I_{2m} \qquad \forall (\tilde z,o).
$$
In other words, the two-layered contracting GRU is indeed globally monotonic, modulo a change of coordinates on $z$ and a rescaling of $\kappa$. This reasoning can be reproduced to handle multi-layered architecture where each cell is globally monotonic and with bounded derivatives with respect to the input $o$.

\subsection{Dataset details}
\label{app:rookie_datasets}

\begin{table}[h]
  \centering
  \caption{Dataset statistics per building: we collected data in 14 floors of 4 buildings distributed over 2 continents. Distance is path length along the odometry trajectory. Floors indicates distinct floor levels covered in each building.} 
  \label{tab:rookie_dataset}
  \setlength{\aboverulesep}{0pt}%
  \setlength{\belowrulesep}{0pt}%
  \begin{tabular}{llc|rrrrrr}
      \toprule                                                                               
      \rowcolor{lightgray}
      Split & Building & Continent & Floors & Runs & Steps & Duration & Distance \\
      \midrule           
      \rowcolor{softblue}
      Train & Building-17          & A & 7 & 51 &  83\,466 & 7h\,43m\,42s & 14.1\,km  \\      
      \rowcolor{softblue}
      Train & Building-GF          & A & 3 & 11 &  19\,105 & 1h\,46m\,08s &  4.5\,km  \\      
      \rowcolor{softblue}
      Train & Building-Q (1F, 2F)  & B & 2 & 16 &  43\,140 & 3h\,59m\,39s &  3.8\,km  \\      
      \rowcolor{softpurple}
      Val& Building-Q (3F) & B & 1 &  8 &  26\,888 & 2h\,29m\,22s &  2.0\,km \\      
      \rowcolor{softgreen}
      Test & Building-CC          & B & 1 & 14 &  24\,113 & 2h\,13m\,57s &  3.1\,km \\      
      \midrule
      \rowcolor{lightgray}
      Total &               &   & 14 & 100 & 196\,712 & 18h\,12m\,49s & 27.4\,km \\       
      \bottomrule
    \end{tabular}
\end{table}

The validation floor (Building-Q) is entirely unseen during training. The full test building is unseen during training and validation. 
The recordings from Buildings-17 and Building-GF are from a large multi-floor office building and introduce substantially more diverse motion patterns.%

In these data, $\delta=1/3s$. Also, we took $\kappa=1$.

\subsection{Training hyper-parameters}
\label{app:rookie_hyperparameters}

\begin{tabular}{l|l}
\hline
 Batch size          & 1024 \\
 Learning rate       & $10^{-4}$ \\
 Minimum LR          & $10^{-5}$ \\
 Scheduler           & Cosine annealing \\
 Optimizer           & Adam \\
 Gradient clipping   & max norm $1.0$ \\
 Epochs              & 2000 \\
\hline
 Chunk length        & 300 steps (100\,s) \\
 Chunk stride        & 150 steps (50\,s) \\
 Traj.\ loss horizon & $H = 50$ steps (17\,s, $\approx$5\,m) \\
 Min.\ speed filter  & $0.05$\,m/s \\
\hline
 Hidden dimension    & 128 \\
 GRU layers          & 2 \\
 Input dimension     & 9 \\
 \startmodifVA Gain / step (Contracting GRU)\stopmodif & \startmodifVA $\kappa = 1$, $\delta = 1/3$, $\alpha = 2.5$\stopmodif \\
\hline
\end{tabular}

\subsection{Empirical filter rank analysis}
\label{app:rookie_filter_rank}

We empirically check whether trained models satisfy the Filter Rank and Capacity Condition of Assumption~\ref{ass:filter_capacity} along realistic robot trajectories.

We consider the involved biases are visible from the input signal $\Observation$ and its first time derivative, i.e., the differential observability condition in Assumption \ref{ass:observability} holds at order $\ell=1$ (velocity biases).
However, notice that the physical system is not autonomous in this experiment due to trajectory control. As usual in control theory, inputs may be assumed to be generated by an autonomous system of sufficiently large dimension. Hence, we may need more time derivatives of $\Observation$ to recover all the missing information, i.e., $\ell >1$. We evaluate the matrix $C_\ell(\Observation)$ at sampled operating points and compute its column rank as a function of $\ell$.

\paragraph{Setup.}
Let $f_\Params(z, \Observation) = z + \delta\kappa\, g_\Params(z, \Observation)$ be the one-step recurrence of the trained encoder, with joint hidden state $z \in \RR^m$ obtained by concatenating the per-layer hidden vectors of the two-layer architecture (so $m = 256$, twice the per-layer dimension of $128$).
For each sampled operating point $\Observation^* \in \Ycal$ drawn from validation trajectories:
\begin{enumerate}
  \item we find the equilibrium $z^*$ such that $f_\Params(z^*, \Observation^*) = z^*$ by iterating the recurrence with constant input until $\|z_{k+1}-z_k\|_2 < 10^{-7}$. The contraction property of Assumption~\ref{ass:contraction} guarantees this for the Contracting GRU;
  \item we compute the local Jacobians by reverse-mode automatic differentiation,
  \begin{equation}
    A(\Observation^*) = \tfrac{\partial f_\Params}{\partial z}(z^*, \Observation^*) \in \RR^{m\times m},
    \qquad
    B(\Observation^*) = \tfrac{\partial f_\Params}{\partial \Observation}(z^*, \Observation^*) \in \RR^{m\times p};
  \end{equation}
  \item we build, for each $\ell \in \{1,\dots,15\}$,
  \begin{equation}
    C_\ell(\Observation^*) = \bigl[\, B(\Observation^*) \,\big|\, A(\Observation^*) B(\Observation^*) \,\big|\, \dots \,\big|\, A(\Observation^*)^{\ell} B(\Observation^*) \,\bigr] \in \RR^{m \times p(\ell+1)},
  \label{eq:controllability_matrix}
  \end{equation}
  and compute its numerical column rank as $r_\ell(\Observation^*) = |\{i : \sigma_i(C_\ell) > \tau\,\sigma_{\max}(C_\ell)\}|$ with $\tau = 10^{-6}$.
\end{enumerate}

\paragraph{Result.}
We discover that the Filter Rank and Capacity Condition is fulfilled up to $\ell=4$.

\section{Compute resources}
\label{app:rookie_compute}

The trained models are small and can even be trained on a laptop CPU in a couple of hours. We trained them on V100 GPUs and did not bother our A100 and H100 GPUs. Training one single model is typically done in a couple of minutes (under half an hour), but it took longer to generate the hyper-parameter sweeps of the paper.

\textbf{Real experiments (pose estimation).} Figure \ref{fig:comparison_gru_val} comparing 3 models on 5 training seeds each took 2.5h to calculate on a single V100. The Contracting GRU took a bit longer (3x or 4x longer) since the implementation is less efficient than the optimized PyTorch GRU.

\textbf{Simulated experiments.} Figures \ref{fig:gru_tuneability_capacity} and \ref{fig:tunability_largedecoder} required training 144 models (6 values of $\kappa$ times 8 decoder widths times 3 seeds for each model) which took around 30 hours.

\section{More details on the simulation experiments (Section \ref{sec:experiments_simulation})}
\label{app:experiments_simulation}

Let us define a compact forward invariant set $\X\subset \RR^3$ of trajectories of interest on which the differential observability Assumption \ref{ass:observability} is satisfied.
Let $\X_3$ be an arbitrary compact set in $(0, +\infty)$. Let $r_0 = \sqrt{\min \X_3}$ and $r_1 = \sqrt{\max \X_3}$.
Consider the function $V:\RR^2\to\RR$ defined by $V(x) = (x_1^2+x_2^2)$.
Let $\mathcal{D}_0 = \{(x_1, x_2)\in\RR^2\mid V(x_1, x_2) < 
r_0^2
\}$
and $\mathcal{D}_1 = \{(x_1, x_2)\in\RR^2\mid V(x_1, x_2) \leq 
r_1^2
\}$.
Define $\X = (\mathcal{D}_1 \setminus \mathcal{D}_0) \times \X_3$.
Then $\X$ is compact and forward invariant. Indeed, for all solutions to \eqref{eq:vdp_aug},
$\frac{d}{dt}V(x_1, x_2) = 2(x_3 - x_1^2 - x_2^2)x_2^2$ is non-negative if $(x_1, x_2)\in \mathcal{D}_0$ and non-positive if  $(x_1, x_2)\in \RR^2\setminus \mathcal{D}_1$.

Now, we prove that Assumption \ref{ass:observability} is satisfied on $\X$, with $\ell=3$.
Computing the successive Lie derivatives of $h(x)=x_1$ with respect to the vector field $f$ of \eqref{eq:vdp_aug}, we obtain for all $x\in\X$:
\begin{align*}
    \Lie h(x) =  x_2, \qquad \Lie^2 h(x) =  (x_3-x_1^2-x_2^2)x_2-x_1.
\end{align*}
Hence
\begin{align*}
    x_3x_2 = (\Lie^2 h + h + (h^2 + (\Lie h)^2)\Lie h)(x).
\end{align*}
Moreover
\begin{align*}
    \Lie (\Lie^2 h + h + (h^2 + (\Lie h)^2)\Lie h)(x)
    =
    x_3\Lie^2 h(x).
\end{align*}
Hence
\begin{align*}
    x_3(x_3x_2 - x_1)
    &= (\Lie+ h^2 + (\Lie h)^2) (\Lie^2 h + h + (h^2 + (\Lie h)^2)\Lie h)(x).
\end{align*}
The above equations define a polynomial map $P:\RR^4\to\RR^4$ such that, for all $x\in\RR^3$,
$(x_1, x_2, x_3x_2, x_3(x_3x_2 - x_1)) = P(h(x), \Lie h(x), \Lie^2 h(x), \Lie^3 h(x))$.

Moreover, \startmodifVA since $0 < x_3 \leq r_1^2$ and $2r_1^2|x_1x_2| \leq \frac{1}{2}x_1^2 + 2r_1^4x_2^2$,\stopmodif for all $x\in\X$,
\begin{align*}
    ({\startmodifVA 2r_1^4\stopmodif}+\frac{1}{2})x_2^2 + (x_3x_2 - x_1)^2
    \geq
    ({\startmodifVA 2r_1^4\stopmodif}+\frac{1}{2})x_2^2 + x_1^2 - {\startmodifVA 2r_1^2\stopmodif}|x_1x_2|
    \geq
    \frac{1}{2}(x_1^2+x_2^2)
    \geq \frac{1}{2}r_0^2.
\end{align*}
Hence,
\begin{align*}
    x_2^2 + (x_3x_2 - x_1)^2\geq \min\left\{1,\frac{1}{{\startmodifVA 2r_1^4\stopmodif}+\frac{1}{2}}\right\}\frac{1}{2}r_0^2.
\end{align*}
Thus
the polynomial map $I:\X\to \RR^4$ defined by
$I(x_1, x_2, x_3) = (x_1, x_2, x_3x_2, x_3(x_3x_2 - x_1))$
admits a globally Lipschitz left-inverse
$J:I(\X) \to\RR^3$ defined by
$J(x_1, x_2, a_1, a_2) = (x_1, x_2, \frac{x_2a_1+(a_1 - x_1)a_2}{x_2^2+(a_1 - x_1)^2})$.

To conclude, for all $x\in\X$,
$x = J(P(h(x), \Lie h(x), \Lie^2 h(x), \Lie^3 h(x)))$.
Since $J\circ P$ is globally Lipschitz over the compact set $h(\X)\times\Lie h(\X)\times\Lie^2h(\X)\times \Lie^3h(\X)$,
system \eqref{eq:vdp_aug} is differentially observable of order $\ell = 3$. In other words, \eqref{eq:observability} is verified, with $c_{obs}$ the Lipschitz constant of $J\circ P$.

\section{About Assumption \ref{ass:filter_capacity}}
\label{Sec_AssumpFilterCapacity}

This appendix provides a formal justification for the Filter Rank and Capacity Condition, demonstrating that the left-invertibility requirement on the matrix $C_\ell$ is a weak condition.

To illustrate how the Filter Rank and Capacity Condition can be structurally satisfied, let us consider a scenario where the physical observation is scalar ($p=1$), the required observability index is $\ell=2$, and the latent space dimension is $m=4$. Aligning with standard deep learning conventions where the input feeds into the first layer, we design the base latent dynamics as a \startmodifVA lower-triangular\stopmodif causal cascade:

\begin{align*}
\dot{\LatentState}_1 &= -\alpha_1 \LatentState_1 + \sigma_1(w_{11} \LatentState_1 + u_1 \Observation + b_1) \\
\dot{\LatentState}_2 &= -\alpha_2 \LatentState_2 + \sigma_2(u_2 \LatentState_1 + w_{22} \LatentState_2 + b_2) \\
\dot{\LatentState}_3 &= -\alpha_3 \LatentState_3 + \sigma_3(u_3 \LatentState_2 + w_{33} \LatentState_3 + b_3) \\
\dot{\LatentState}_4 &= -\alpha_4 \LatentState_4 + \sigma_4(u_4 \LatentState_3 + w_{44} \LatentState_4 + b_4)
\end{align*}

where the activation functions $\sigma_i$ are strictly monotonic (hence $\sigma'_i > 0$ everywhere) and the coupling weights $u_i$ are strictly non-zero. By evaluating the Jacobians at the equilibrium, we obtain the matrices $A \in \RR^{4 \times 4}$ and $B \in \RR^{4 \times 1}$:

$$
A = \begin{bmatrix}
-\alpha_1 + \sigma'_1 w_{11} & 0 & 0 & 0 \\
\sigma'_2 u_2 & -\alpha_2 + \sigma'_2 w_{22} & 0 & 0 \\
0 & \sigma'_3 u_3 & -\alpha_3 + \sigma'_3 w_{33} & 0 \\
0 & 0 & \sigma'_4 u_4 & -\alpha_4 + \sigma'_4 w_{44}
\end{bmatrix}, \quad 
B = \begin{bmatrix}
\sigma'_1 u_1 \\ 0 \\ 0 \\ 0
\end{bmatrix}
$$

To highlight the algebraic rank properties, let $0$ denote exactly zero, $\times$ denote a strictly non-zero element, and $\star$ represent an arbitrary scalar. The matrices exhibit the following structured sparsity pattern:

$$
A = \begin{bmatrix}
\star & 0 & 0 & 0 \\
\times & \star & 0 & 0 \\
0 & \times & \star & 0 \\
0 & 0 & \times & \star
\end{bmatrix}, \quad
B = \begin{bmatrix}
\times \\ 0 \\ 0 \\ 0
\end{bmatrix}
$$

This causal structure naturally maps to a lower-triangular rank progression. The required $(4, 3)$ matrix $C_2 = \begin{bmatrix} B & AB & A^2B \end{bmatrix}$ becomes:

$$
C_2 = \begin{bmatrix}
\sigma'_1 u_1 & \star & \star \\ 
0 & \sigma'_1 u_1 \sigma'_2 u_2 & \star \\ 
0 & 0 & \sigma'_1 u_1 \sigma'_2 u_2 \sigma'_3 u_3 \\ 
0 & 0 & 0
\end{bmatrix}
=
\begin{bmatrix}
\times & \star & \star \\
0 & \times & \star \\
0 & 0 & \times \\
0 & 0 & 0
\end{bmatrix}
$$

Because the upper $3 \times 3$ submatrix of $C_2$ is upper-triangular, if the $u_i$'s are non zero, $C_2$ maintains full column rank uniformly over the observation domain. This explicitly guarantees the left-invertibility required by Assumption \ref{ass:filter_capacity}. 

While this specific cascaded architecture provides a constructive proof of existence, we conjecture that this rank requirement is not a restrictive architectural constraint. In fact, left-invertibility likely constitutes a generic topological property for a broad class of over-parameterized latent dynamics. However, establishing such a general result involves complex differential topology arguments that fall beyond the scope of this paper. A first result that we can give is the following.

\begin{proposition}[Genericity for Single-Layer Analytic RNNs]
\label{prop:genericity_analytic}
Consider the latent dynamics of a single-layer RNN evaluated at a fixed observation point $o^* \in \mathbb{R}^p$. For simplicity of exposition, assume $p=1$:
\begin{equation}
    \dot{z} = -\alpha z + \sigma(W z + U o^* + b),
\end{equation}
where $z \in \mathbb{R}^m$, the parameter vector is $\Theta = (W, U, b) \in \mathbb{R}^{m \times m} \times \mathbb{R}^m \times \mathbb{R}^m$, and $\alpha > 0$. 
Assume that:
\begin{enumerate}
    \item \textbf{Analyticity:} The activation function $\sigma$ is real-analytic and  $\sigma'(x) > 0$ everywhere.
    \item \textbf{Dimensionality:} The latent dimension satisfies $m \ge \ell+1$.
\end{enumerate}
Then, the filter observability matrix $C_\ell(\Theta, o^*) \in \mathbb{R}^{m \times (\ell+1)}$ has full column rank for almost all parameter configurations $\Theta$ (i.e., outside a set of Lebesgue measure zero).
\end{proposition}

\begin{proof}
Let $v(\Theta) = W z_{eq} + U o^* + b$ denote the pre-activation state at the equilibrium $z_{eq}$. The local Jacobians evaluated at this equilibrium are:
\begin{align}
    A(\Theta) &= -\alpha I_m + \Sigma'(v(\Theta)) W, \label{eq:jac_A} \\
    B(\Theta) &= \Sigma'(v(\Theta)) U, \label{eq:jac_B}
\end{align}
where $\Sigma'(v) = \mathrm{diag}(\sigma'(v_1), \dots, \sigma'(v_m))$. Because $\sigma$ is strictly monotonic, $\sigma'(v_i) > 0$ for all $i$.

The matrix $C_\ell(\Theta)$ has full column rank $\ell+1$ if and only if at least one of its $(\ell+1) \times (\ell+1)$ minors has a non-zero determinant. 
Let $F(\Theta)$ be the sum of the squares of all such maximal minors. The matrix is rank-deficient if and only if $F(\Theta) = 0$.

Because $\sigma$ is a real-analytic function, its derivative $\sigma'$ is also real-analytic. Consequently, the entries of the matrices $A(\Theta)$ and $B(\Theta)$ in \eqref{eq:jac_A} and \eqref{eq:jac_B} are real-analytic functions of the parameters $\Theta$. It follows that the function $F(\Theta)$ is real-analytic on $\mathbb{R}^N$.

If an analytic function on a connected open set is not identically zero, its zero set has Lebesgue measure zero \cite[Proposition 0]{Mityagin2020}. Therefore, to prove that $C_\ell(\Theta)$ is generically full rank, it is sufficient to construct a single parameter configuration $\Theta^* = (W^*, U^*, b^*)$ for which $F(\Theta^*) \neq 0$.

We construct this configuration as follows:
\begin{enumerate}
    \item Choose $W^* = \mathrm{diag}(\lambda_1, \dots, \lambda_m)$ such that the eigenvalues $\lambda_i$ are pairwise distinct.
    \item Choose $U^* = [1, 1, \dots, 1]^\top$.
    \item Choose $b^*$ such that  $v_1 = v_2 = \dots = v_m$. This ensures that the derivative is a strictly positive constant $c = \sigma'(v_i) > 0$ for all $i$.
\end{enumerate}

Under the configuration $\Theta^*$, the matrix $\Sigma'(v)$ simplifies to $c I_m$. Thus, the Jacobian $A^*$ is diagonal:
\begin{equation}
    A^* = \mathrm{diag}(a_1, \dots, a_m), \quad \text{where } a_i = -\alpha + c \lambda_i.
\end{equation}
Because the $\lambda_i$ are pairwise distinct and $c > 0$, the eigenvalues $a_i$ are also pairwise distinct. The vector $B^*$ simplifies to $c [1, 1, \dots, 1]^\top$.

The observability matrix $C_\ell(\Theta^*)$ takes the explicit form:
\begin{equation}
    C_\ell(\Theta^*) = c \begin{bmatrix}
    1 & a_1 & a_1^2 & \dots & a_1^\ell \\
    1 & a_2 & a_2^2 & \dots & a_2^\ell \\
    \vdots & \vdots & \vdots & \ddots & \vdots \\
    1 & a_m & a_m^2 & \dots & a_m^\ell
    \end{bmatrix}
\end{equation}

Because $m \ge \ell+1$, we can extract the square submatrix $M \in \mathbb{R}^{(\ell+1) \times (\ell+1)}$ formed by the first $\ell+1$ rows of $C_\ell(\Theta^*)$. Up to the non-zero scaling factor $c$, $M$ is precisely a Vandermonde matrix generated by the elements $(a_1, \dots, a_{\ell+1})$.

Since the values $a_i$ are pairwise distinct, the determinant of the Vandermonde matrix is  non-zero. Consequently, the maximal minor corresponding to $M$ is non-zero, which  implies $F(\Theta^*) > 0$.

Because $F(\Theta)$ is not identically zero, the set $\{\Theta \in \mathbb{R}^N \mid F(\Theta) = 0\}$ has Lebesgue measure zero. We conclude that $C_\ell(\Theta, o^*)$ has full column rank for almost all parameters $\Theta$.
\end{proof}

\newpage
\section*{NeurIPS Paper Checklist}

\begin{enumerate}

\item {\bf Claims}
    \item[] Question: Do the main claims made in the abstract and introduction accurately reflect the paper's contributions and scope?
    \item[] Answer: \answerYes{}{} %
    \item[] Justification: We provided all proofs and added experimental corroboration.

\item {\bf Limitations}
    \item[] Question: Does the paper discuss the limitations of the work performed by the authors?
    \item[] Answer: \answerYes{}{} %
    \item[] Justification: Given in Section \ref{sec:limitations}
    
\item {\bf Theory assumptions and proofs}
    \item[] Question: For each theoretical result, does the paper provide the full set of assumptions and a complete (and correct) proof?
    \item[] Answer: \answerYes{} %
    \item[] Justification: All proofs are given in the supplementary material
    
    \item {\bf Experimental result reproducibility}
    \item[] Question: Does the paper fully disclose all the information needed to reproduce the main experimental results of the paper to the extent that it affects the main claims and/or conclusions of the paper (regardless of whether the code and data are provided or not)?
    \item[] Answer: \answerYes{} %
    \item[] Justification: The paper describes all details required to implement models and training algorithms and is thus reproducible in that sense. However, for the part of the experiments which deal with a real robot application, the models have been trained on an internal large-scale dataset, which we unfortunately have not yet made public. 
    
\item {\bf Open access to data and code}
    \item[] Question: Does the paper provide open access to the data and code, with sufficient instructions to faithfully reproduce the main experimental results, as described in supplemental material?
    \item[] Answer: \answerNo{} %
    \item[] Justification: This is mainly a theory paper. However, the part of the experiments which deal with a real robot application, the models have been trained on an internal large-scale dataset, which we unfortunately have not yet made public. Code will be released at some point but could unfortunately not do this before submission.

\item {\bf Experimental setting/details}
    \item[] Question: Does the paper specify all the training and test details (e.g., data splits, hyperparameters, how they were chosen, type of optimizer) necessary to understand the results?
    \item[] Answer: \answerYes{} %
    \item[] Justification: The training hyper-parameters have been given in Section \ref{app:rookie_hyperparameters}
    
\item {\bf Experiment statistical significance}
    \item[] Question: Does the paper report error bars suitably and correctly defined or other appropriate information about the statistical significance of the experiments?
    \item[] Answer: \answerYes{} %
    \item[] Justification: This is mainly a theoretical paper, but for the experimental part we trained the models with multiple seeds and we report plots with means $\pm$ standard-deviation.
    
\item {\bf Experiments compute resources}
    \item[] Question: For each experiment, does the paper provide sufficient information on the computer resources (type of compute workers, memory, time of execution) needed to reproduce the experiments?
    \item[] Answer: \answerYes{} %
    \item[] Justification: This is explained in Section \ref{app:rookie_compute}
        
\item {\bf Code of ethics}
    \item[] Question: Does the research conducted in the paper conform, in every respect, with the NeurIPS Code of Ethics \url{https://neurips.cc/public/EthicsGuidelines}?
    \item[] Answer: \answerYes{} %
    \item[] Justification: there is nothing to justify.

\item {\bf Broader impacts}
    \item[] Question: Does the paper discuss both potential positive societal impacts and negative societal impacts of the work performed?
    \item[] Answer: \answerNA{}{} %
    \item[] Justification: This is highly theoretical work and while we could invent societal impact, we are not sure that this is a worthwhile exercise.
    
\item {\bf Safeguards}
    \item[] Question: Does the paper describe safeguards that have been put in place for responsible release of data or models that have a high risk for misuse (e.g., pre-trained language models, image generators, or scraped datasets)?
    \item[] Answer: \answerNA{}{} %
    \item[] Justification: This is highly theoretical work and we are not sure how safeguards apply.

\item {\bf Licenses for existing assets}
    \item[] Question: Are the creators or original owners of assets (e.g., code, data, models), used in the paper, properly credited and are the license and terms of use explicitly mentioned and properly respected?
    \item[] Answer: \answerNA{}{} %
    \item[] Justification: The paper only uses internal datasets of the team.

\item {\bf New assets}
    \item[] Question: Are new assets introduced in the paper well documented and is the documentation provided alongside the assets?
    \item[] Answer: \answerNA{}{}{}{} %
    \item[] Justification: We do not introduce new datasets.
    
\item {\bf Crowdsourcing and research with human subjects}
    \item[] Question: For crowdsourcing experiments and research with human subjects, does the paper include the full text of instructions given to participants and screenshots, if applicable, as well as details about compensation (if any)? 
    \item[] Answer: \answerNA{}{} %
    \item[] Justification: We did not use crowd sourcing.
    
\item {\bf Institutional review board (IRB) approvals or equivalent for research with human subjects}
    \item[] Question: Does the paper describe potential risks incurred by study participants, whether such risks were disclosed to the subjects, and whether Institutional Review Board (IRB) approvals (or an equivalent approval/review based on the requirements of your country or institution) were obtained?
    \item[] Answer: \answerNA{}{} %
    \item[] Justification: We did not use research with or on human subjects.

\item {\bf Declaration of LLM usage}
    \item[] Question: Does the paper describe the usage of LLMs if it is an important, original, or non-standard component of the core methods in this research? Note that if the LLM is used only for writing, editing, or formatting purposes and does \emph{not} impact the core methodology, scientific rigor, or originality of the research, declaration is not required.
    \item[] Answer: \answerNA{}{} %
    \item[] Justification: LLMs were only used for coding assistance and for improving writing.

\end{enumerate}

\end{document}